\documentclass{article}
\extrafloats{100}
\usepackage{times}
\usepackage[utf8]{inputenc} %
\usepackage[T1]{fontenc}    %
\usepackage[hidelinks]{hyperref}       %
\usepackage{url}            %
\usepackage{booktabs}       %
\usepackage{amsfonts}       %
\usepackage{nicefrac}       %
\usepackage{microtype}      %
\usepackage{enumitem}

\usepackage{pifont}
\usepackage{xspace}
\newcommand{\circone}{\ding{172}\xspace}
\newcommand{\circtwo}{\ding{173}\xspace}
\newcommand{\circthree}{\ding{174}\xspace}

\usepackage{graphicx}
\usepackage{sidecap}
\usepackage[small]{caption}
\usepackage{subcaption}

\usepackage{wrapfig}
\usepackage{mathtools}

\usepackage{float}
\usepackage{algorithm}
\usepackage{algorithmic}

\usepackage[usenames,dvipsnames,svgnames,table]{xcolor}  %
\usepackage[disable]{todonotes}

\newcommand{\note}[4][]{{\todo[author=#2,color=#3,size=\scriptsize,fancyline,caption={},#1]{#4}}}

\newcommand{\dingpeng}[2][]{{\note[#1]{DP}{blue!20}{#2}}}

\newcommand{\Kangrui}[2][]{\Kangrui[inline,#1]{#2}\noindent}
\newcommand{\Jiading}[2][]{\Jiading[inline,#1]{#2}\noindent}

\newcommand{\cutforspace}[1]{}

\newcommand{\codefont}{\fontfamily{lmtt}\selectfont}
\usepackage{listings}

\usepackage{parcolumns}
\lstdefinestyle{datalogstyle}{
	basicstyle={\codefont\small},  %
	xleftmargin={6pt},
        xrightmargin={6pt},
        columns=flexible,
        breakindent=0pt,
        breaklines=true, 
	frame=tb,
	stepnumber=1,
	firstnumber=1,
	numberfirstline=true,
	tabsize=2,
	extendedchars=true,
	breaklines=true,
	columns=fullflexible,
	keepspaces=true,
	escapeinside={@}{@},
	firstnumber=last,
	captionpos=b, 
	commentstyle=\color{black!65},
	numberstyle=\tiny\color{black!65},
	stringstyle=\color{codepurple},
	breakatwhitespace=false, 
	keepspaces=true,              
        mathescape=true, 
	numbersep=5pt,                  
	showspaces=false,                
	showstringspaces=false,
	showtabs=false,
	aboveskip={0.8\baselineskip},
	belowskip={0.2\baselineskip},
}
\newcommand{\data}[1]{\texttt{\codefont#1}}
\newcommand{\code}[1]{\data{#1}}  %
\newcommand{\mvar}[1]{\code{\textit{#1}\hspace{-1pt}}} %

\makeatletter

\usepackage{tikz}
\usetikzlibrary{shapes,calc,positioning}

\global

\usepackage[most]{tcolorbox}
\tcbset{
  aibox/.style={
    top=10pt,
    colback=white,
    colframe=black,
    boxrule=0.5pt,
    colbacktitle=black,
    enhanced,
    center,
    attach boxed title to top left={yshift=-0.1in,xshift=0.15in},
    boxed title style={boxrule=0pt,colframe=white,},
  }
}
\newtcolorbox{AIbox}[2][]{aibox,title=#2,#1}

\usepackage{multirow}

\definecolor{aigold}{RGB}{244,210, 1} 
\definecolor{aigreen}{RGB}{213, 245, 227}

\definecolor{humanpurple}{RGB}{235, 222, 240}

\definecolor{commentgray}{RGB}{86, 101, 115}

\definecolor{aired}{RGB}{255,180,181}

\usepackage[noabbrev,capitalize]{cleveref} %
\crefname{equation}{equation}{equations}   %
\crefname{section}{section}{sections}      %
\crefname{footnote}{footnote}{footnotes}   
\crefname{listing}{Example}{Examples}
\crefname{assumption}{assumption}{assumptions}
\crefname{line}{line}{lines}   %

\newcommand{\defeq}{\mathrel{\stackrel{\textnormal{\tiny def}}{=}}}

\newcommand{\mangoicon}[0]{\!\includegraphics[height=.02\textwidth]{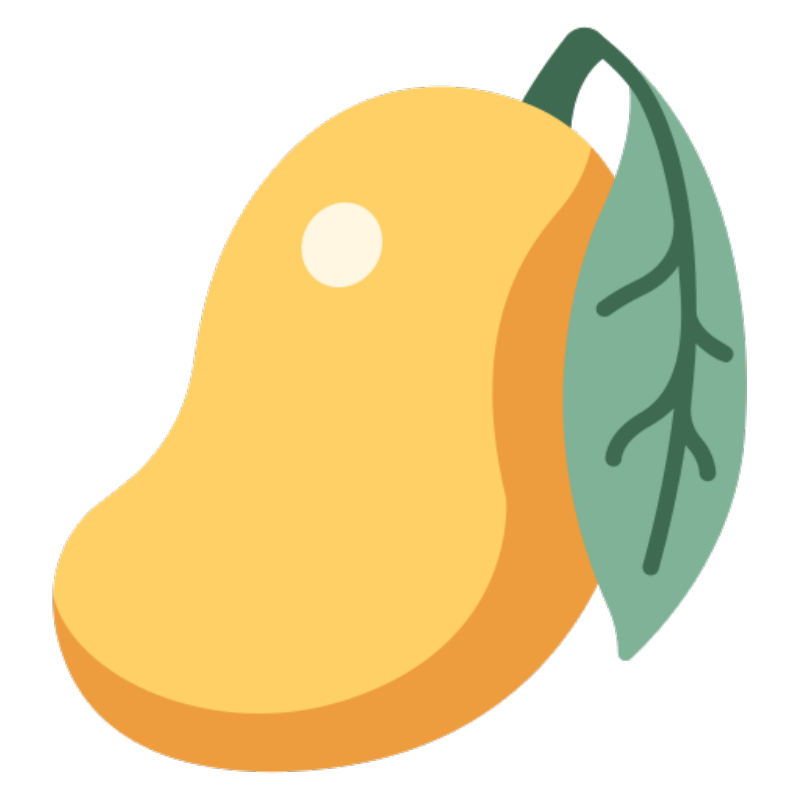}}

\newcommand{\weburl}{https://mango.ttic.edu}
\newcommand{\giturl}{https://github.com/oaklight/mango/}

\usepackage{colm2024_conference}
\colmfinalcopy %
\usepackage{natbib}

\usepackage{xpatch}
\usepackage[compact]{titlesec}
\titlespacing{\section}{0pt}{0.5ex}{0.5ex}
\titlespacing{\subsection}{0pt}{0.5ex}{0ex}
\usepackage{setspace}
\AtBeginDocument{%
  \addtolength\abovedisplayskip{-0.25\baselineskip}%
  \addtolength\belowdisplayskip{-0.25\baselineskip}%
  \addtolength\abovedisplayshortskip{-0.25\baselineskip}%
  \addtolength\belowdisplayshortskip{-0.25\baselineskip}%
}

\makeatletter
\renewcommand{\@fnsymbol}[1]{%
  \ifcase#1\or \mangoicon\else\@ctrerr\fi}
\makeatother

\title{MANGO: A Benchmark for Evaluating \underline{Ma}pping and \underline{N}avi\underline{g}ati\underline{o}n Abilities of Large Language Models}

\author{Peng Ding$^1$\thanks{Equal contribution. Work done while visiting TTIC.} \quad Jiading Fang$^{2\,\mangoicon}$ \quad Peng Li$^{2\,\mangoicon}$ \quad Kangrui Wang$^{2\,\mangoicon}$ \quad Xiaochen Zhou$^{3,5\,\mangoicon}$\\
    \textbf{Mo Yu}$^4$ \quad \textbf{Jing Li}$^5$ \quad
    \textbf{Matthew R.\@ Walter}$^2$ \quad \textbf{Hongyuan Mei}$^2$\\
    $^1$University of Chicago\enspace
    $^2$Toyota Technological Institute at Chicago\enspace
    $^3$Syracuse University\\
    $^4$WeChat AI\enspace
    $^5$New Jersey Institute of Technology\\
    \texttt{\{kangrui,mwalter,hongyuan\}@ttic.edu}
}

\begin{document}

\maketitle

\begin{abstract}
Large language models such as ChatGPT and GPT-4 have recently achieved astonishing performance on a variety of natural language processing tasks. 
In this paper, we propose MANGO, a benchmark to evaluate their ability to perform text-based mapping and navigation. 
Our benchmark includes $53$ mazes taken from a suite of textgames: each maze is paired with a walkthrough that visits every location but does \emph{not} cover all possible paths. 
The task is question-answering: for each maze, a large language model reads the walkthrough and answers hundreds of mapping and navigation questions such as ``How should you go to \code{Attic} from \code{West of House}?'' and ``Where are we if we go \code{north} and \code{east} from \code{Cellar}?''.
Although these questions are easy for humans, it turns out that even GPT-4, the best-to-date language model, performs poorly when answering them. 
Further, our experiments suggest that a strong mapping and navigation ability would benefit the performance of large language models on relevant downstream tasks, such as playing textgames. %
Our MANGO benchmark will facilitate future research on methods that improve the mapping and navigation capabilities of LLMs. 
We host our leaderboard, data, code, and evaluation program at 
{\small \url{\weburl}} and {\small \url{\giturl}}.

\end{abstract}

\section{Introduction}\label{sec:intro}
 Mapping and navigation are fundamental abilities of 
human intelligence~\citep{spiers06, epstein17}. 
Humans are able to construct maps---in their minds~\citep{epstein17} or on physical media like paper---as they explore unknown environments. 
Following these maps, humans can navigate through complex environments~\citep{spiers06,spiers15,javadi17}, making informed decisions, and interact with their surroundings.
Such abilities empower humans to explore, adapt, and thrive in diverse environments. 
An example is remote (e.g., deep-sea) exploration for which humans have drawn upon their intuition to develop algorithms that enable robots to autonomously navigate and map their surroundings based only on onboard sensing. %

\begin{figure}[t]
     \centering
    \includegraphics[width=0.85\linewidth]{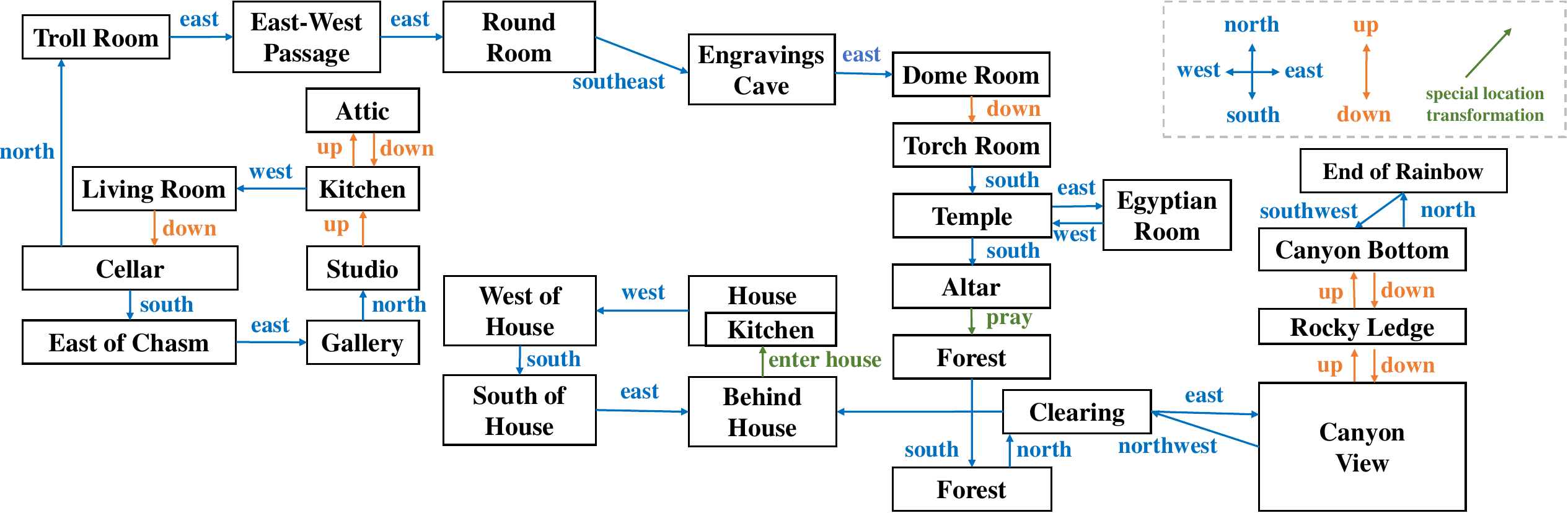}
    \vspace{-7pt}
    \caption{Map of Zork-I. Arrows denote the direction of travel during the walkthrough, while the reverse direction is unseen but may be possible. Note that it is a 3D map projected onto a 2D plane so \code{up} may not point upward in the 2D visualization (e.g., \code{Rocky Ledge} to \code{Canyon View}).}
    \label{fig:map}
\end{figure}

Do large language models (LLMs) possess such abilities? %
In this paper, we investigate this research question by creating a benchmark and evaluating several widely used LLMs. 
Our MANGO benchmark is the \emph{first} to measure the \underline{ma}pping and \underline{n}avi\underline{g}ati\underline{o}n abilities of LLMs. 
It includes $53$ complex mazes, such as the one visualized in \cref{fig:map}. %
It pairs each maze with hundreds of destination-finding questions (e.g., ``Where will you be if you go \code{north}, \code{north}, and then \code{up} from \code{Altar}?'') and route-finding questions (e.g., ``How do you reach \code{Dome Room} from \code{Altar}?'').
For each maze, the language model has to answer these questions after reading a walkthrough of the maze. 
Many questions involve possible routes that are not traced during the walkthrough, making the benchmark challenging. %
In our experiments, GPT-4 only correctly answered half of the route-finding questions, performing disastrously on the difficult questions (e.g., those involving long and unseen routes).
MANGO will facilitate future research in improving the mapping and navigation abilities of LLMs.

Another contribution of MANGO is to draw a novel connection between natural language processing and robotics. 
There has been significant interest in employing LLMs %
to endow intelligent agents (including robots) with complex reasoning~\citep{yang2023foundation}. 
Aligning with this interest, MANGO enables the investigation of the LLMs' capabilities in simultaneous localization and mapping (SLAM) within text-based worlds. 
Focusing on this aspect, our work stands out and complements previous SLAM-related research, 
which predominantly relies on richer sensory inputs (e.g., vision and LiDAR).

\section{MANGO: A Benchmark for Text-Based Mapping and Navigation}\label{sec:mango}
Our MANGO benchmark measures the mapping and navigation capabilities of LLMs.
It leverages a suite of text-based adventure games that offer expert-designed complex environments but only require simple actions. 
\cref{fig:map} is an example: 
it was drawn according to the first 70 steps of the walkthrough of Zork-I, which can be found in \cref{lst:walkthrough}. 
This map is imperfect: the annotator had to draw the only \code{Kitchen} twice to avoid a cluttered visualization; the \code{Living Room} was incorrectly placed outside the \code{House}. 
However, equipped with this map, one could correctly answer questions about any route in the maze such as ``How do you reach \code{Dome Room} from \code{Altar}?'' and ``Where will you be if you go \code{north}, \code{north}, and \code{up} from \code{Altar}?''. 
The walkthrough has not traced a route from \code{Altar} to \code{Dome Room}, but humans possess the remarkable capacity to %
plan a route by identifying the three individual steps---which the walkthrough has covered---from \code{Dome Room} to \code{Altar} and retracing those steps. 
MANGO tests whether a large language model can perform the same kind of reasoning. 
Particularly, when evaluating a language model, we first let it read a walkthrough like \cref{lst:walkthrough} and then ask it %
questions like those in \cref{lst:dfq,lst:rfq}. 
A question like \cref{lst:dfq} is a \emph{destination-finding} (DF) question, and a question like \cref{lst:rfq} is a \emph{route-finding} (RF) question. 
Users of MANGO have the flexibility to phrase the DF and RF questions in their own ways: as shown in \cref{lst:dfqraw,lst:rfqraw}, we provide the \emph{skeletons} of these questions, which users can plug into their own templates.  
\begin{figure}[t]
\begin{minipage}[t]{0.5\linewidth}
\begin{lstlisting}[caption={An example of Zork-I walkthrough.},label={lst:walkthrough}]
STEP NUM: 0
ACT: Init
OBSERVATION: West of House
You are standing in an open field west of a white house, with a boarded front door. There is a small mailbox here.

STEP NUM: 1
ACT: south
OBSERVATION: South of House
You are facing the south side of a white house.

STEP NUM: 2
ACT: east
OBSERVATION: Behind House
You are behind the white house. A path leads into the forest to the east. In one corner of the house there is a small window which is slightly ajar.@\progvdots@

STEP NUM: 70
ACT: east
OBSERVATION: Gallery
This is an art gallery. Most of the paintings have been stolen by vandals with exceptional taste. The vandals left through either the north or west exits. Fortunately, there is still one chance for you to be a vandal, for on the far wall is a painting of unparalleled beauty.
\end{lstlisting}
\end{minipage}
\begin{minipage}[t]{0.5\linewidth}

\begin{lstlisting}[caption={A destination-finding question.},label={lst:dfq}]
Starting from Altar, perform actions [north, north, up], where are you now? 
\end{lstlisting}

\vspace{4pt}
\begin{lstlisting}[caption={A route-finding question.},label={lst:rfq}]
How can you go from Altar to Dome Room? 
\end{lstlisting}

\vspace{6pt}
\begin{lstlisting}[caption={Skeleton of DF question in \cref{lst:dfq}.},label={lst:dfqraw}]
@\mvar{S}@: Altar            @\textcolor{commentgray}{\# starting location}@
@\mvar{A}@: north, north, up   @\textcolor{commentgray}{\# list of actions}@
\end{lstlisting}

\vspace{4pt}
\begin{lstlisting}[caption={Skeleton of RF question in \cref{lst:rfq}.},label={lst:rfqraw}]
@\mvar{S}@: Altar            @\textcolor{commentgray}{\# starting location}@
@\mvar{D}@: Dome Roomm             @\textcolor{commentgray}{\# destination}@
\end{lstlisting}

\vspace{6pt}
\begin{lstlisting}[caption={Full route of \cref{lst:dfq,lst:dfqraw}.},label={lst:route}]
@\mvar{S}@: Altar              
@\mvar{A}@: north
@\mvar{D}@: Temple

@\mvar{S}@: Temple              
@\mvar{A}@: north
@\mvar{D}@: Torch Room

@\mvar{S}@: Torch Room              
@\mvar{A}@: up
@\mvar{D}@: Dome Room
\end{lstlisting}

\end{minipage}
\vspace{-8pt}
\end{figure}

\dingpeng[]{in rebuttal, we promised a complete example of walkthrough, we may use partial zork1 as appendix or some other ones with less step}

\subsection{Maze Collection: From Game Walkthroughs to Mazes}\label{sec:data}
Our mazes are taken from the textgames in the Jericho game suite~\citep{hausknecht2020interactive}. 
The main release of Jericho includes $57$ popular textgames as well as a program that can generate walkthroughs for 56 of them.
The original walkthrough of a game is a list of actions (such as \code{east}, \code{north}, and \code{open door}) that one could execute to efficiently complete the game.
We enhanced %
each walkthrough by executing the sequence of actions and augmenting each step with the new observation (i.e., the text feedback that the game engine provides after the action is executed).
Unless explicitly specified, the word ``walkthrough'' refers to the enhanced, but not original, walkthroughs (such as \cref{lst:walkthrough}) throughout the paper. 
More details about walkthroughs can be found in \cref{app:walkthrough}. 

In a walkthrough, not every action triggers a location change: it may update the inventory (such as \code{take lamp} and \code{drop pen}) or time (such as \code{wait}). 
For each game, we read the walkthrough, labeled the actions (such as \code{east} and \code{up}) that change the locations, and made note of the names of the locations (such as \code{Temple} and \code{Altar}). 
This annotation is nontrivial and can not be automated. 
We had to pay extra attention to appropriately handle the tricky cases including: 
\circone the name of a location may be mentioned in a rich, but distracting context (e.g., the context may have ten paragraphs and hundreds of words with the name briefly mentioned in the middle); 
\circtwo a location may be visited multiple times, so we need to assign the same name to all its mentions; 
\circthree different locations may be referred to with the same name in the textual feedback, so we need to rename them in a sensible way.

The location name resolution (see \cref{app:location} for a full procedure) results in a maze for each game. 
Three of the games have no location change, and so we left them out, resulting in $53$ mazes. %
We store each maze as a directed graph: each node is a named location (e.g., \code{Altar}); each directed edge is a movement (e.g., \code{north}); and each node-edge-node combination is a location-changing step that was followed in the walkthrough.
Note that a graph may be cyclic since the walkthrough may trace back-and-forth between locations (e.g., \code{Temple} and \code{Egyptian Room} in \cref{fig:map}). 

\subsection{Generation of Question Skeletons: Traversing Mazes and Imputing Edges}\label{sec:qg}
To generate DF and RF skeletons for a maze, a naive approach is to perform brute-force traversal. 
First, we collect all the possible \mvar{S}-\mvar{P}-\mvar{D} tuples, where \mvar{S} and \mvar{D} are locations and \mvar{P} is a simple path from \mvar{S} to \mvar{D}. 
A simple path %
is a directed path that does not visit any location more than once. 
This ``simple'' restriction ensures that we will have a finite number of \mvar{S}-\mvar{P}-\mvar{D} tuples. 
\cref{lst:route} is a simple path of 3 \mvar{S}-\mvar{A}-\mvar{D} edges from \code{Altar} to \code{Dome Room}. 
Each unique \mvar{S}-\mvar{P}-\mvar{D} tuple gives a unique DF skeleton: e.g., \cref{lst:dfqraw} is obtained from \cref{lst:route}. 
Each unique \mvar{S}-\mvar{P}-\mvar{D} tuple gives an RF skeleton as well, such as \cref{lst:rfqraw} obtained from \cref{lst:route}. 
However, the same RF skeleton may be obtained from other tuples since there may be multiple possible simple paths between the same pair of locations \mvar{S} and \mvar{D}. 
As a consequence, we may end up with fewer RF questions than DF questions for a given maze. 

The particular DF and RF questions in \cref{lst:dfq,lst:rfq} are challenging to large language models, since they involve actions---such as going \code{north} from \code{Altar} to \code{Temple}---that are not covered in the walkthrough. 
Answering such hard questions requires a deeper understanding of the spatial relationships between locations. 
However, also because these steps are not in the walkthrough, the skeletons in \cref{lst:dfqraw,lst:rfqraw} can not be obtained through a naive traversal of the directed graph in \cref{fig:map}. 
That is, we have to traverse an extended graph that includes \emph{imputed} edges. 
An imputed edge denotes a valid step that is not explicitly mentioned in the walkthrough, such as going \code{north} from \code{Altar} to \code{Temple} (i.e., \code{Altar}-\code{north}-\code{Temple}). 
Most mentioned edges involve directional moves (e.g., \code{up}, \code{east}), so reversing them is a straightforward way to impute new edges. 
We manually examined other edges: for some of them, we proposed intuitive reverses (such as \code{exit} for \code{enter}); for the others (e.g., \code{pray}), no reverse could be found. 
We then examined the imputed edges through real game play and discarded those failing to cause the expected location changes. 
\cref{app:qg} documents the full procedure of edge imputation and examination.

After extending all the mazes in our benchmark, we collected 21046 DF skeletons and 14698 RF skeletons by traversing the extended graphs. 
Being evaluated on a maze, the LLM may not be able to consume the entire walkthrough in its context window. 
That is, we may only feed it an appropriate prefix of the walkthrough (e.g., the first $70$ steps for Zork-I as shown in \cref{lst:walkthrough}), leaving some of the DF and RF skeletons \emph{unanswerable} given that prefix. 
Therefore, our benchmark provides the ANSWERABLE label (an integer) for each skeleton such that this skeleton is only answerable if the maximum STEP NUM in that prefix (e.g., $70$ in \cref{lst:walkthrough}) is greater than or equal to its ANSWERABLE label. 
Furthermore, given a walkthrough prefix, an answerable skeleton may be easy or hard, depending on whether it involves edges that are not covered in the prefix. 
Precisely, a DF skeleton is considered to be easy if all the \mvar{S}-\mvar{A}-\mvar{D} edges in its corresponding simple path are covered in the walkthrough prefix; an RF skeleton is easy if the shortest simple path from its starting location to its destination only involves the \mvar{S}-\mvar{A}-\mvar{D} steps covered in the prefix.
When a longer walkthrough prefix is used, more answerable questions tend to become easy. 
Our benchmark provides the EASY label (also an integer) for each skeleton: a skeleton is easy if the maximum STEP NUM in the walkthrough prefix is no smaller than its EASY label; otherwise, it is a hard skeleton. 
\cref{tab:fulldatastat} in \cref{app:data} documents the statistics of the full dataset, such as the number of locations and the number of skeletons. 
\cref{tab:expdatastatgpt,tab:expdatastatllama-2,tab:expdatastatother,tab:expdatastatllama-1} in \cref{app:exp} shows the information about the data on which each LLM was evaluated in our experiments. 

\subsection{Evaluation Program}\label{sec:eval}
The evaluation program in our benchmark implements a range of evaluation and analysis methods.
Reading the model-generated answers, it can return a set of evaluation scores together with rich analysis. 
In this section, we introduce the most important scores used in our main experiments. 
Other scores are discussed in \cref{app:eval}, with their related experiments presented in \cref{app:results}.

For DF questions, the most straightforward evaluation is the success rate: 
i.e., the fraction of questions that the language model answers correctly. 
What answers will be considered to be correct? 
A strict criteria is that the model answer is correct if and only if it exactly matches the ground-truth location name. 
However, due to the variability of natural language, a correct answer may not exactly match the ground-truth. 
For example, the model may yield \code{The House} or \code{That House} when the ground-truth location name is just \code{House}. 
To account for such cases, we generalize the success rate to allow partial matches. 
Given a model answer $\hat{A}$ and the ground-truth answer $A$, we compute their (character-level) edit-distance $d$ and define a correctness score $c \defeq 1-d/\ell$ where $\ell$ is the length of the longer answer. 
The score is $\in [0,1]$: when the answer exactly matches the ground-truth, we have $c=1$; if they have no character overlap at all, then $c=0$. 
We then define the success rate to be the sum of the correctness scores over all the questions, divided by the number of questions. 

For RF questions, the main metric is still the success rate, but the definition of ``success'' is different from that for DF questions. 
Note that an answer to an RF question is a sequence of moves. 
We consider an answer to be correct if and only if it can reach the destination after our evaluation program executes it in the maze. 
A correct answer to an RF question may not be a good path: it doesn't have to be the shortest; it doesn't even have to be a simple path. 
It is possible that an LLM-generated move is meaningful but doesn't exactly match any valid move in the graph: e.g., the LLM may give \code{walk south}, which means the same as \code{south}.   
Therefore, when executing a model-generated move, our evaluation program will select the closest (i.e., smallest edit-distance) valid move.

\section{Experiments}\label{sec:exp}
In this section, we present the results our evaluation of several widely used LLMs. 

\subsection{Experiment Setup}\label{sec:setup}
The evaluated models are: GPT-3.5-turbo~\citep{brown-2020-gpt,stiennon2020learning,gao2022scaling}, GPT-4~\citep{gpt4}, Claude-instant-1~\citep{claude1}, Claude-2~\citep{claude2}, Llama-2 with 13B parameters~\citep{touvron2023llama2}, %
and RWKV with 14B parameters~\citep{peng2023rwkv}. %
For GPTs and Claudes, we used the prompt templates in \cref{lst:maptemplate,lst:navtemplate}, converting the DF and RF skeletons like \cref{lst:dfqraw,lst:rfqraw} into LLM-friendly questions like \cref{lst:dfq,lst:rfq}. 
The templates were carefully designed and examined through pilot experiments, in order to ensure that we do not underestimate the models on our benchmark. 
In our templates, each question starts with a list of legal actions, followed by a list of reachable locations; these lists help mitigate the hallucination of language models. 
The templates ask the model to spell out the entire trajectory including all the intermediate locations. 
This design is inspired by Chain-of-Thought prompting~\citep{wei2022chain}: eliciting an LLM to give its entire reasoning process tends to improve its overall performance on downstream tasks. 
In addition, it allows us to conduct a deeper evaluation and analysis, such as the reasoning accuracies of the models (see \cref{app:eval,app:results}). 
Note that our templates request the model to form its answer as a list of Python dictionaries with specific key names. 
We found that this restriction encourages the model to generate structured answers---which are easy to parse and analyze---as well as improves its performance. %
For Llama-2 and RWKV, we made moderate revisions to the prompts in order to generate well-structured answers as well as optimize for their performance. %
\begin{figure}[t]%
\begin{minipage}[t]{0.5\linewidth}
\begin{lstlisting}[caption={Our DF template.},label={lst:maptemplate}]
The allowed actions are: @\ldots@
The list of places are: @\ldots@
Starting from @\mvar{S}@, perform a list of 
actions [@\mvar{A}@], where are you now? 
Describe the trajectory in a Python list of Python dictionaries with keys 'prev_node', 'node' and 'action'. 
Start your response with '['.
\end{lstlisting}
\end{minipage}
\begin{minipage}[t]{0.5\linewidth}
\begin{lstlisting}[caption={Our RF template.},label={lst:navtemplate}]
The allowed actions are: @\ldots@
The list of places are: @\ldots@
How can you go from @\mvar{S}@ to @\mvar{D}@? 

Describe the trajectory in a Python list of Python dictionaries with keys 'prev_node', 'node' and 'action'. 
Start your response with '['.
\end{lstlisting}
\end{minipage}
\vspace{-8pt}
\end{figure}

For GPT-3.5, we experimented with the 4K version, which can consume 4096 tokens in its context window. 
This context limit restricts the length of the walkthrough that it can read, and the number of DF and RF questions that it can answer. 
\cref{tab:expdatastatgpt} shows the statistics about the walkthrough prefix and questions that GPT-3.5 used for each maze. 
For GPT-4, Claude-1 and Claude-2, we used the same walkthrough prefixes and questions as GPT-3.5 for a fair comparison. we used the same walkthrough prefixes and questions as GPT-3.5 for a fair comparison. 
Llama-2 has a $4096$ context window as well. But its tokenizer is different from GPTs' so we evaluated it on a slightly different set of questions. 
RWKV is capable of handling infinite context. 
For each maze, we experimented it with the $70$-step prefix of the walkthrough so that its set of answerable questions includes all the questions answered by all the other models. 
We also evaluated Llama-2 and RWKV in a simplified setting, where the observation at each step of the walkthrough only includes the location name but nothing else. 
For example, at STEP 1 of the simplified \cref{lst:walkthrough}, OBSERVATION only has \code{South of House} and everything else (i.e., \code{Your are...}) is omitted. 
We refer to Llama-2 and RWKV with the simplified walkthroughs as Llama-2-S and RWKV-S, respectively. %
More details about the experiment setup are in \cref{app:exp}. 

\subsection{Main Results}\label{sec:mainresult}
\begin{figure*}[t]
	\begin{center}
	    \begin{subfigure}[t]{0.48\linewidth}
                \includegraphics[width=0.49\linewidth]{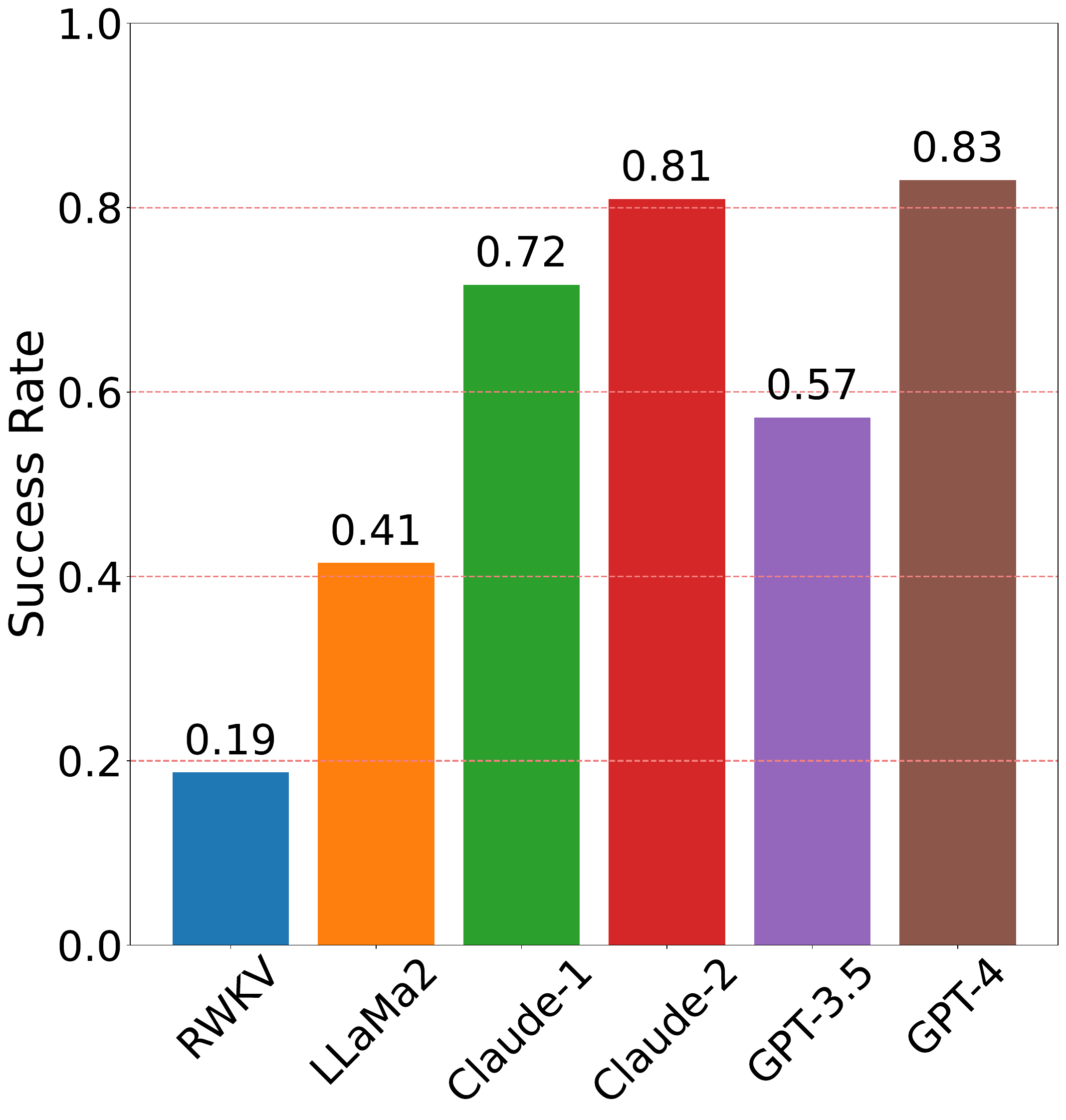}
                \hfill
                \includegraphics[width=0.49\linewidth]{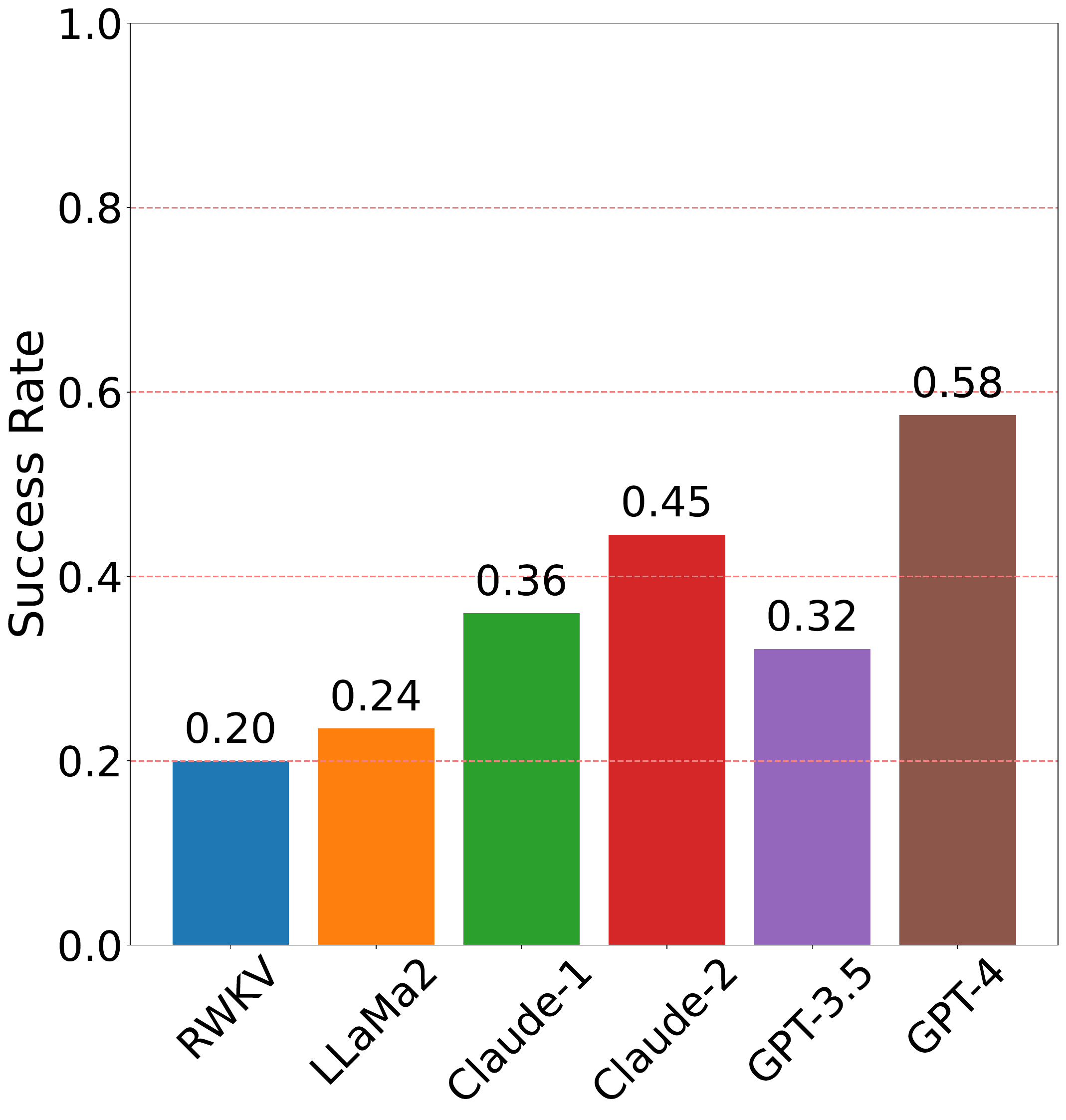}
		    \vspace{-16pt}
			\caption{On easy (left) and hard (right) DF questions.}\label{fig:main_df}
		\end{subfigure}
		\hfill
            \begin{subfigure}[t]{0.48\linewidth}
			\includegraphics[width=0.49\linewidth]{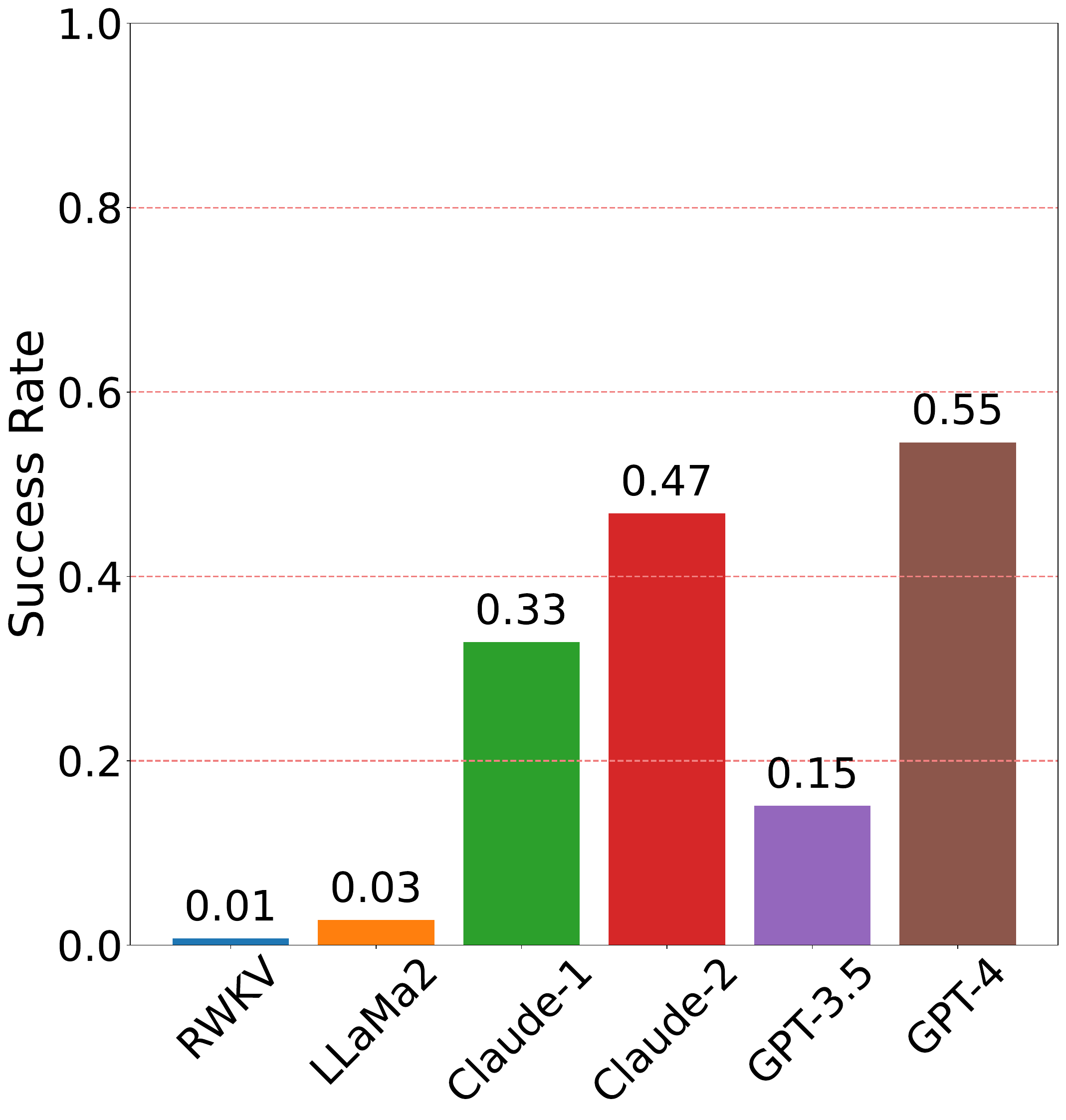}
                \hfill
                \includegraphics[width=0.49\linewidth]{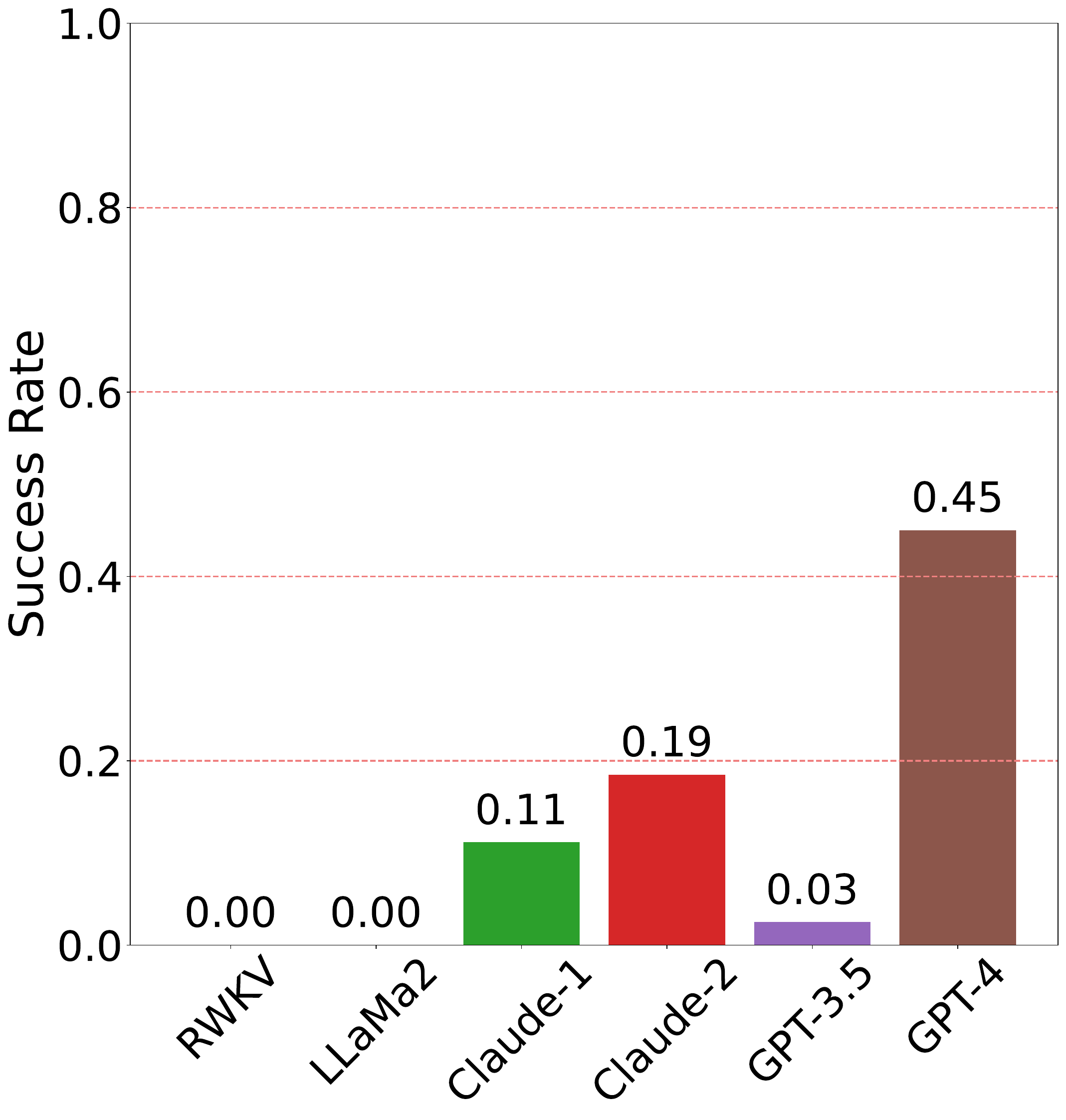}
			\vspace{-16pt}
			\caption{On easy (left) and hard (right) RF questions.}\label{fig:main_rf}
		\end{subfigure}
		\vspace{-8pt}
		\caption{Success rates of the examined models on (\subref{fig:main_df}) DF and (\subref{fig:main_rf}) RF questions, averaged over all $53$ mazes. 
            \cref{app:results} provides similar graphs (e.g., \cref{fig:main_rea_acc}) for other evaluation metrics.}\label{fig:main}
	\end{center}
	\vspace{-4pt}
\end{figure*}

\cref{fig:main} presents the success rates of all models. 
For each kind of question (i.e., DF or RF), we show the results on easy and hard questions separately. 
As we can see, GPT-4 significantly outperforms all the other models on all kinds of questions. 
However, it only correctly answers half of the RF questions, far worse than what a human could do: in our experiments, humans perfectly answered a randomly sampled set of questions.
\begin{table}[t]
\small
\setlength{\tabcolsep}{4pt} %
\begin{sc}
\begin{subtable}{1.0\linewidth}
\begin{center}
\begin{tabular}{lccccccc}
\toprule
Method & RWKV & Llama-2 & Claude-1 & Claude-2 & GPT-3.5 & GPT-4 & $\overline{{\rm HARD}}|$ \\
\midrule
RWKV & * & 0.20 | 0.24 & 0.19 | 0.41 & 0.19 | 0.51 & 0.19 | 0.33 & 0.19 | 0.62 & * \\
Llama-2 & 0.43 | 0.20 & * & 0.24 | 0.41 & 0.24 | 0.45 & 0.24 | 0.31 & 0.24 | 0.66 & * \\
Claude-1 & 0.74 | 0.19 & 0.78 | 0.41 & * & 0.36 | 0.44 & 0.38 | 0.32 & 0.36 | 0.57 & * \\
Claude-2 & 0.82 | 0.19 & 0.85 | 0.41 & 0.81 | 0.72 & * & 0.44 | 0.32 & 0.44 | 0.58 & * \\
GPT-3.5 & 0.59 | 0.19 & 0.61 | 0.42 & 0.57 | 0.74 & 0.57 | 0.83 & * & 0.32 | 0.59 & * \\
GPT-4 & 0.86 | 0.19 & 0.90 | 0.42 & 0.84 | 0.72 & 0.83 | 0.81 & 0.86 | 0.57 & * & * \\
|$\underline{{\rm EASY}}$ & * & * & * & * & * & * & *\\
\bottomrule
\end{tabular}
\vspace{-4pt}
\caption{Pairwise comparison on easy (lower left) and hard (higher right) DF questions.}
\vspace{4pt}
\begin{tabular}{lccccccc}
\toprule
Method & RWKV & LlaMa-2 & Claude-1 & Claude-2 & GPT-3.5 & GPT-4 & $\overline{{\rm HARD}}|$ \\
\midrule
RWKV & * & 0.00 | 0.00 & 0.00 | 0.13 & 0.00 | 0.20 & 0.00 | 0.03 & 0.00 | 0.54 & * \\
Llama-2 & 0.02 | 0.02 & * & 0.00 | 0.16 & 0.00 | 0.21 & 0.00 | 0.05 & 0.00 | 0.46 & * \\
Claude-1 & 0.36 | 0.01 & 0.34 | 0.03 & * & 0.11 | 0.19 & 0.13 | 0.03 & 0.11 | 0.45 & * \\
Claude-2 & 0.49 | 0.01 & 0.46 | 0.03 & 0.47 | 0.33 & * & 0.20 | 0.03 & 0.19 | 0.46 & * \\
GPT-3.5 & 0.16 | 0.01 & 0.17 | 0.03 & 0.15 | 0.36 & 0.15 | 0.50 & * & 0.03 | 0.48 & * \\
GPT-4 & 0.57 | 0.01 & 0.56 | 0.03 & 0.55 | 0.33 & 0.55 | 0.47 & 0.58 | 0.15 & * & * \\
|$\underline{{\rm EASY}}$ & * & * & * & * & * & * & *\\
\bottomrule
\end{tabular}
\vspace{-4pt}
\caption{Pairwise comparison on easy (lower left) and hard (higher right) RF questions.}
\end{center}
\end{subtable}%
\end{sc}
\vspace{-4pt}
\caption{Success rates on DF and RF questions broken down into pairwise comparisons. In each table, the cell of row-A and col-B contains the success rates of the models---in the format of A | B---on the intersection of the questions that A and B answered individually. The lower left triangle displays the results on easy questions, while the upper right triangle shows the results on hard questions.}
\label{tab:successratepartialmatch}
\end{table}
Note that each model was evaluated on its specific set of questions determined by the length and format of the walkthrough it read. 
To be fair, we also compared each pair of models on the intersection of the questions that they answered. 
The results are presented in \cref{tab:successratepartialmatch}: as we can see, GPT-4 and GPT-3.5 consistently outperform the other models and GPT-4 significantly outperforms GPT-3.5. 

More results are in \cref{app:results}, including results on other evaluation metrics (e.g., weighted success rates) and comparison between Llama-2 with Llama-1 and Llama-2-chat.
We also explored an alternative approach that first maps a walkthrough to a symbolic graph and then uses a search algorithm to answer the given (DF or RF) question.
Results and analysis of this approach can be found in \cref{app:search}. 
Overall, we found it to be very challenging to translate natural language walkthroughs into searchable symbolic graphs in the first place, making this approach not promising.

\subsection{Analysis of GPTs}\label{sec:analysis}
Now, we focus our analysis on the best model, namely GPT-4.
Particularly, we would like to understand the improvements of GPT-4 over GPT-3.5 as well as its current bottlenecks, shedding light on opportunities for future improvements. 

\begin{figure*}[t]
	\begin{center}
	    \begin{subfigure}[t]{0.24\linewidth}
                \begin{center}
                \includegraphics[width=1.00\linewidth]{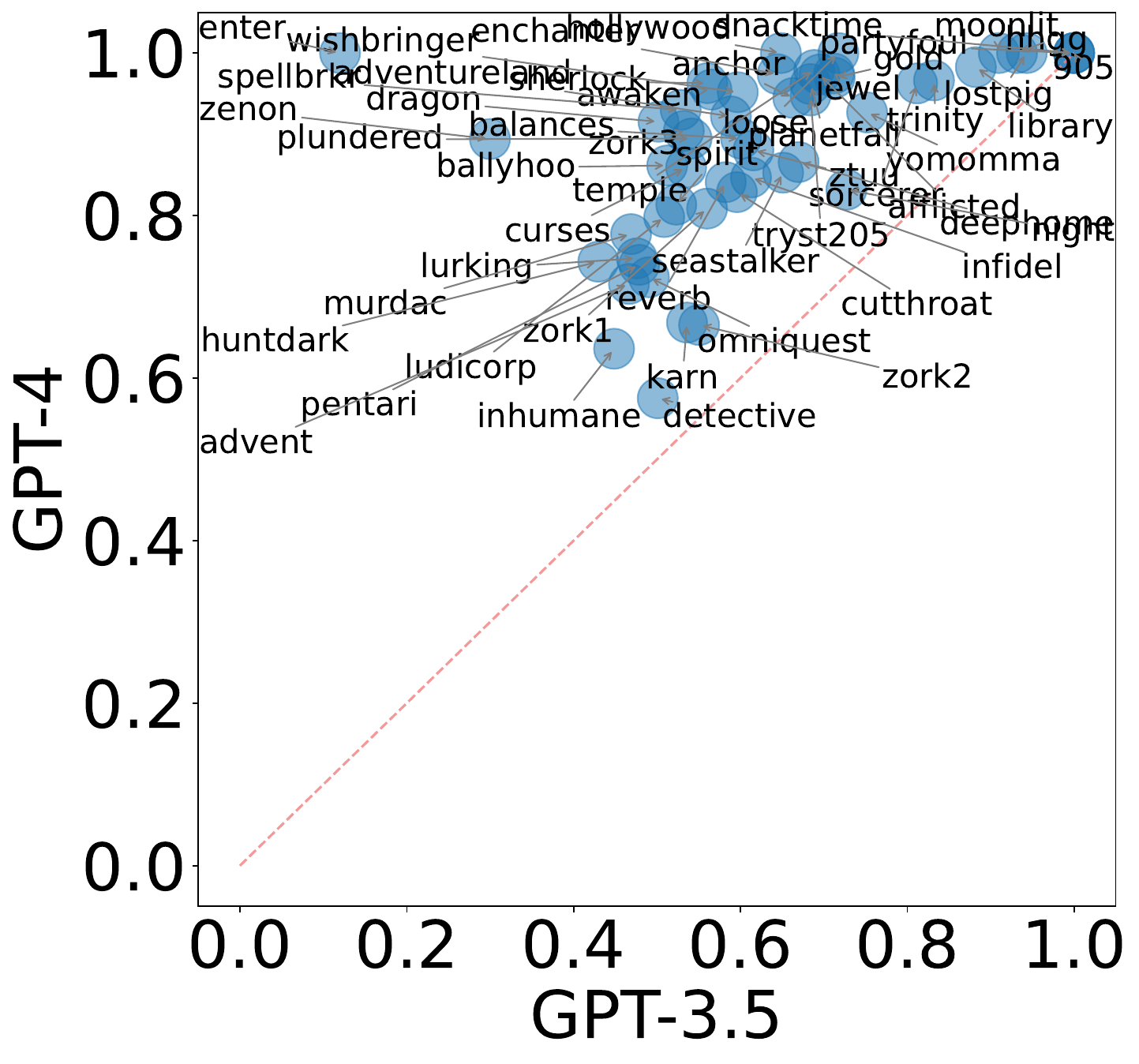}
                \end{center}
                \vspace{-8pt}
               \caption{Easy DF.}\label{fig:gpt3vs4_df_easy}
		\end{subfigure}
            \hfill
            \begin{subfigure}[t]{0.24\linewidth}
                \begin{center}
                \includegraphics[width=1.00\linewidth]{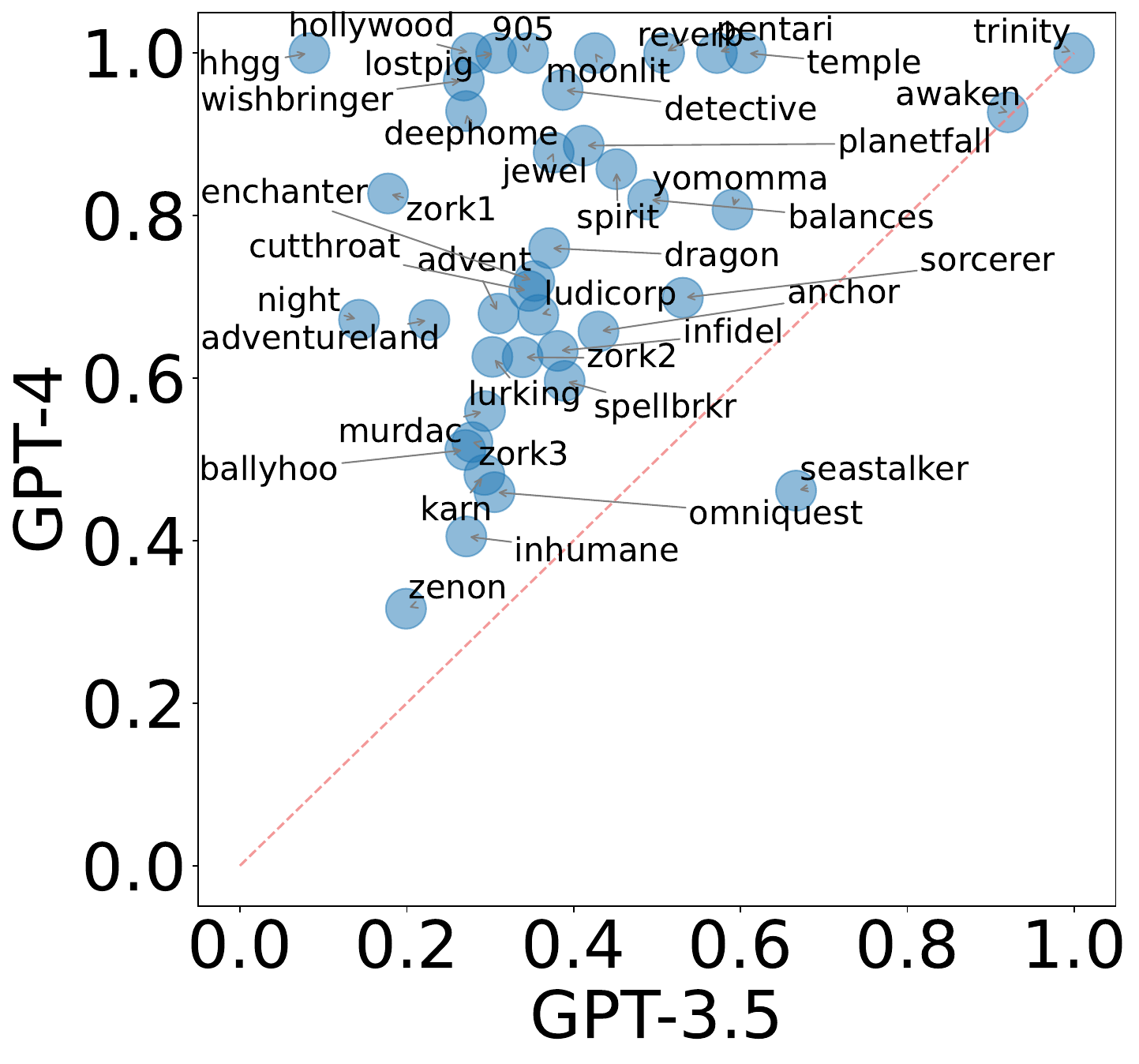}
                \end{center}
                \vspace{-8pt}
               \caption{Hard DF.}\label{fig:gpt3vs4_df_hard}
		\end{subfigure}
		\hfill
            \begin{subfigure}[t]{0.24\linewidth}
                \begin{center}
                \includegraphics[width=1.00\linewidth]{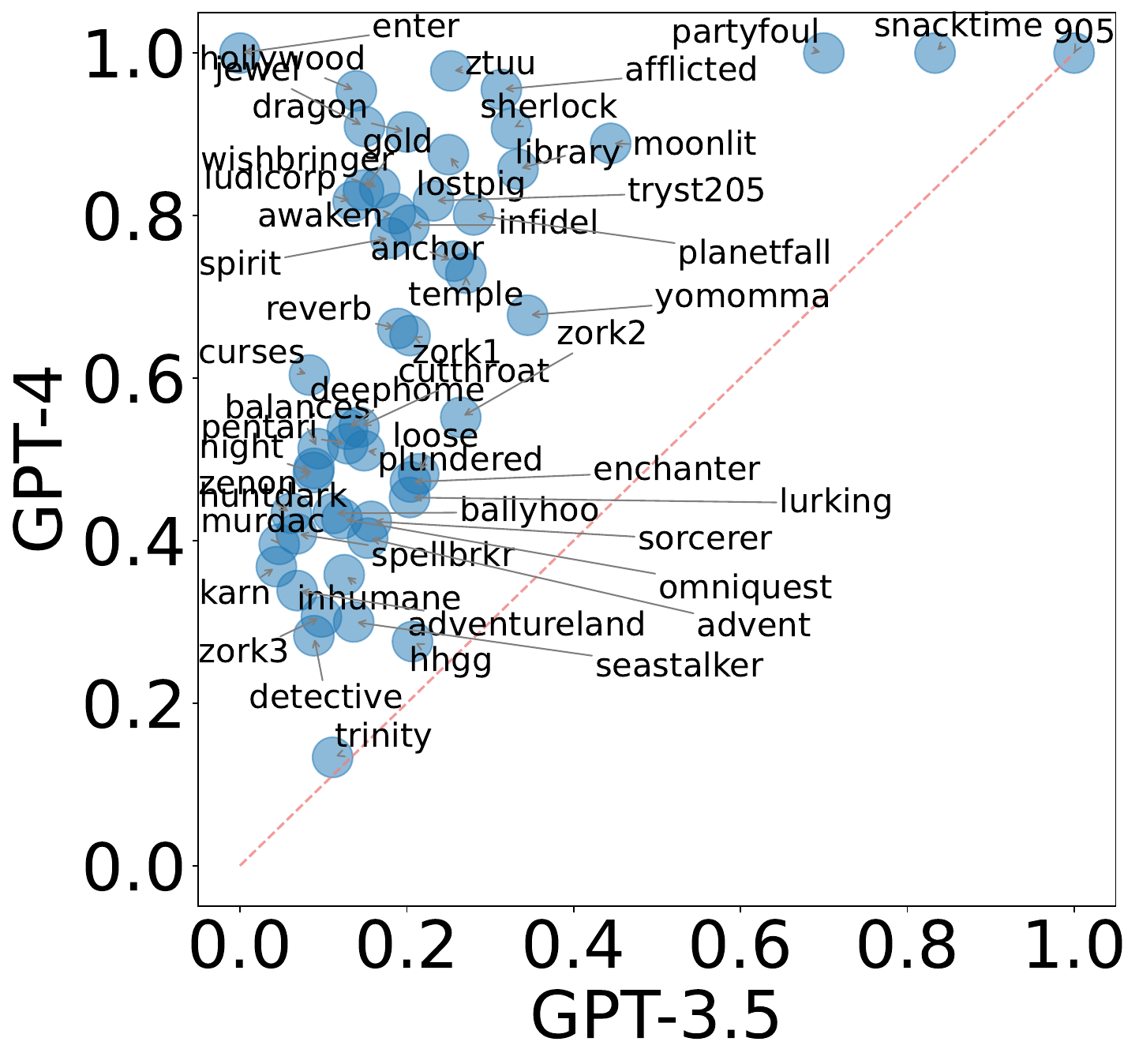}
                \end{center}
                \vspace{-8pt}
			\caption{Easy RF.}\label{fig:gpt3vs4_rf_easy}
		\end{subfigure}
            \hfill
            \begin{subfigure}[t]{0.24\linewidth}
                \begin{center}
                \includegraphics[width=1.00\linewidth]{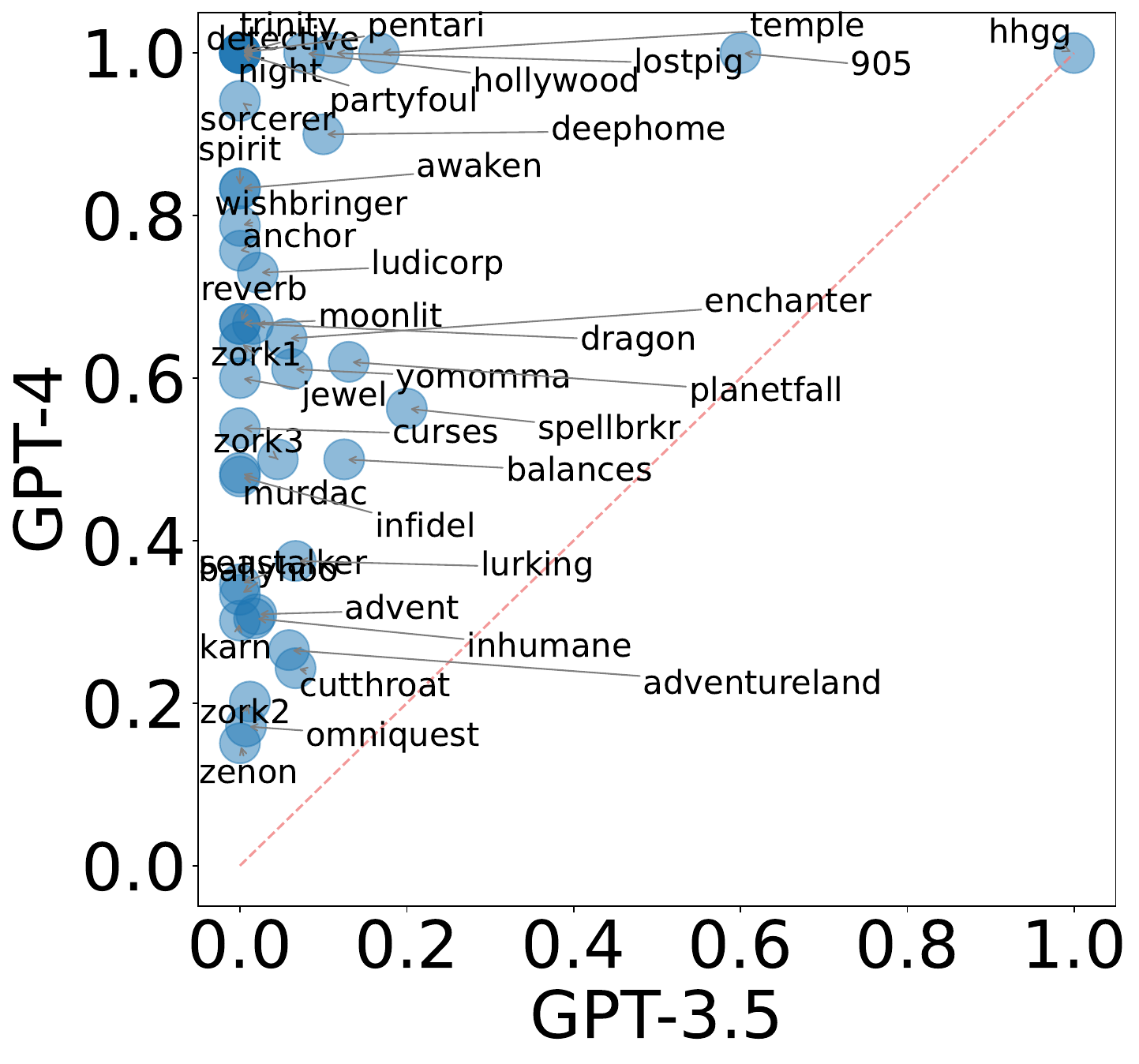}
                \end{center}
                \vspace{-8pt}
			\caption{Hard RF.}\label{fig:gpt3vs4_rf_hard}
		\end{subfigure}
		\vspace{-4pt}
		\caption{
            Success rates of GPT-3.5 and GPT-4 broken down into individual games. 
            \cref{fig:gpt3vs4_reasoning} in \cref{app:exp} provides a similar visualization of reasoning accuracy.
            }\label{fig:gpt3vs4}
	\end{center}
	\vspace{-8pt}
\end{figure*}

By analyzing the errors of GPT-3.5 and GPT-4, we discovered that these models occasionally hallucinate nonexistent locations or edges. 
Once they made such a mistake at any step of reasoning, they would be misled and deviate from the correct path towards the correct answer. 
Furthermore, we found that the mazes are not equally difficult for the models. 
\cref{fig:gpt3vs4} displays the success rates of the GPT models broken down into their per-game results.
In \cref{fig:gpt3vs4}, each dot is a maze: the $x$-axis coefficient is the performance of GPT-3.5 on this maze while the $y$-axis is that of GPT-4. 
As we can see, the success rates of the models vary across different mazes as well as across different kinds of questions. 
GPT-4 consistently outperforms GPT-3.5 across nearly all the mazes. 
The only exception is Seastalker: there are too few hard DF questions for this maze, and thus it is a noisy outlier. 
Apparently, both GPTs tend to work better on easy questions than on hard questions. 
However, some mazes seem to be particularly challenging to GPT-4, such as Zenon and OMNIQuest.%

\paragraph{What makes those mazes challenging?} %
We collected some important statistics about the mazes and analyzed their correlation with the success rates of the models. 
To understand the success rates on the easy questions, it is interesting to investigate: 
\begin{itemize}[leftmargin=*,nosep,topsep=0pt]
    \item number of locations (\# locations) and number of explicit edges (\# exp edges). They directly measure the size of a maze, which may be a key indicator of its difficulty.
    \item number of potentially confusingly named locations (\# conf locations). Recall from \cref{sec:data} that different locations may have similar or related names, which may confuse a language model. 
    To quantify the number of confusingly named locations, we compute a confusion score for each location, and then sum the scores across all the locations. 
    For a location name $A$, the confusion score is defined to be the maximum word-level edit distance between $A$ and any other location name in the maze, divided by the maximum word-level length of the pair of location names being compared. Technically, it is $\max_{B} \left( \text{edit-distance}(A,B) / \max(\text{len}(A),\text{len}(B)) \right)$, and it is $\in [0,1]$. 
    \item average length of the easy simple paths (avg len easy), i.e., the simple paths that do not include any imputed edges. A longer path may tend to be more difficult for models. 
    \item average number of words in the scene descriptions (avg len scene). The walkthroughs exhibit very diverse styles: for some of them, the text description for each scene is very concise and the name of each location is appropriately highlighted; for others, each description may be verbose (e.g., ten paragraphs and hundreds of words) and the location names are often not obvious from the contexts. 
    It is useful to analyze whether a long scene description poses a challenge for the models. 
\end{itemize}
In order to understand the models' performance on hard questions, we analyze the effects of the variables above (except avg len easy) as well as the following: 
\begin{itemize}[leftmargin=*,nosep,topsep=0pt]
    \item number of imputed edges (\# imp edges); 
    \item average length of hard---i.e., involving imputed edges---simple routes (avg len hard); 
    \item average number of imputed edges in the hard simple routes (avg \# imp in hard). 
\end{itemize}

We use regression analysis to understand the effects of these variables on model performance. 
In particular, for each model on each type of question (DF or RF, easy or hard), we ran single-variable linear regression to understand how the success rate varies with each variable. 
Detailed results (e.g., coefficients and $p$-values) are \cref{app:analysis}. Overall, 
\begin{itemize}[leftmargin=*,nosep,topsep=0pt]
    \item on easy questions, GPTs are significantly influenced by the size of the maze, the confusion level of location name, and the path length. The $p$-values are extremely small. 
    \item on easy questions, the average length of the scene descriptions does not have a significant effect on the performance of GPT-3.5, but interestingly has a significant positive effect on GPT-4's performance. 
    It is perhaps because GPT-4 possesses a strong capability to understand texts and can leverage the rich contexts in each description. 
    This allows it to better distinguish confusingly named locations and establish a better internal representation of the map. %
    However, this richness seems to confuse GPT-3.5 and impede its ability to create a good internal representation of the maze, possibly due to GPT-3.5's weaker overall language understanding capabilities.    
    \item on hard questions, the variables do not significantly affect the performance of GPT-3.5. 
    Note that GPT-3.5 yields very low success rates when answering the hard DF and RF questions. 
    GPT-3.5 seems to struggle when it has to reason about a path with any number of imputed edges, making the effect of other factors less important to its performance.
    \item on hard questions, GPT-4 exhibits a stronger ability to handle paths with imputed edges, compared to GPT-3.5. 
    However, it will experience difficulties when the challenge of inferring imputed edges is amplified by other factors such as the size of the maze or the length of the path. 
    As a result, nearly all the variables have significant effects on GPT-4. 
\end{itemize}

The results of our regression analysis are consistent with the plots in \cref{fig:gpt3vs4}. 
For example, both Zenon and OMNIQuest stay at the lower-left corners of the hard-question plots in \cref{fig:gpt3vs4} since their mazes are particularly challenging to both GPT-3.5 and GPT-4: they both have substantially larger numbers of imputed edges than the other mazes; OMNIQuest also has more locations. 
Wishbringer and Lost Pig have several imputed edges, but their paths are short, so they fall in the upper-left corners of the hard-question plots in \cref{fig:gpt3vs4}.

\subsection{Human Performance}\label{sec:human}
We measured human performance on a subset of our data. 
This subset includes 30 DF questions (21 easy and 9 hard) and 31 RF questions (20 easy, 11 hard). 
The student authors participated in two rounds of evaluation: 
in the first round, each author answered a random split of the questions, and all the questions were answered; 
in the second round, we randomly sampled 10 DF and 10 RF questions and let each of them be re-evaluated by a different author. 
The second round allows us to analyze human agreement; this analysis can be found in \cref{app:human}. 
\begin{table}[h]
\small
\begin{sc}
\begin{center}
\begin{tabular}{llccc}
\hline
Task & Difficulty & Success Rate & Reasoning Accuracy \\
\hline
\multirow{3}{*}{Route Finding} & All (31) & 0.8211 & 0.6129 \\
 & Easy (20) & 0.7727 & 0.5500 \\
 & Hard (11) & 0.9091 & 0.7273 \\
\hline
\multirow{3}{*}{Destination Finding} & All (30) & 1.0000 & 0.5667 \\
 & Easy (21) & 1.0000 & 0.6667 \\
 & Hard (9) & 1.0000 & 0.3333 \\
\hline
\end{tabular}
\end{center}
\end{sc}
\vspace{-4pt}
\caption{Human Evaluation Results}
\label{tab:human_eval}
\end{table}

As shown in \cref{tab:human_eval}, humans generally achieve high success rates on both RF and DF questions, and exhibit small to no difference across easy and hard questions. 
This is interesting since LLMs tend to struggle on hard questions. 
Reasoning accuracies are lower than success rates; this is because human answers exhibit a relatively high variability in writing (e.g., location names with special symbols). 

Despite the high scores, answering the questions is actually not trivial for human raters. 
Answering each DF question takes an average of 15 minutes, while answering each RF question takes an average of 30 minutes. 
All the human raters needed to take notes (e.g., drawing maps) while completing these tasks.

\subsection{Does Mapping and Navigation Ability Matter in Downstream Tasks?}\label{sec:downstream}
Now we present a case study showing that a strong mapping and navigation ability of an LLM would benefit it in downstream tasks. 
In particular, we selected 284 \emph{minigames} in the Jericho game suite, and investigated how the map knowledge may improve the performance of an LLM in playing these minigames. 
Each minigame is a selected prefix of a walkthrough from one of 53 textgames; the selection criterion is that the best action to take at this step is a movement. 
In other words, each minigame is a scenario where the LLM has to figure out the best action to take given its previous actions and observations (i.e., the prefix of walkthrough up to the current step). 
This task is different and more challenging than answering the DF and RF questions: the LLM is not explicitly given a route (as in DF questions) or a destination (as in RF questions), but has to spontaneously figure out which action may contribute to its long-term goal.

\begin{figure}[t]%
\begin{minipage}[t]{0.62\linewidth}
\vspace{0pt}
\begin{lstlisting}[caption={A prompt of the playing game experiments},label={lst:playgameprompt}]
@\ldots@           @\textcolor{commentgray}{\# previous actions and observations}@
Small local map info: if you want to go to North of House, you should go south; if you want to go to Up a Tree, you should go up; if you want to go to Altar, you should go west. 
Consider what you should do next, and choose one appropriate action from the valid actions list: [up, take on egg, put down egg, go around forest, throw egg at tree, open egg with all, north, south, west, east]
Please just tell me the selected action without any extra words.
\end{lstlisting}
\end{minipage}
\begin{minipage}[t]{0.37 \linewidth}
\vspace{0pt}
\includegraphics[width=0.9\linewidth]{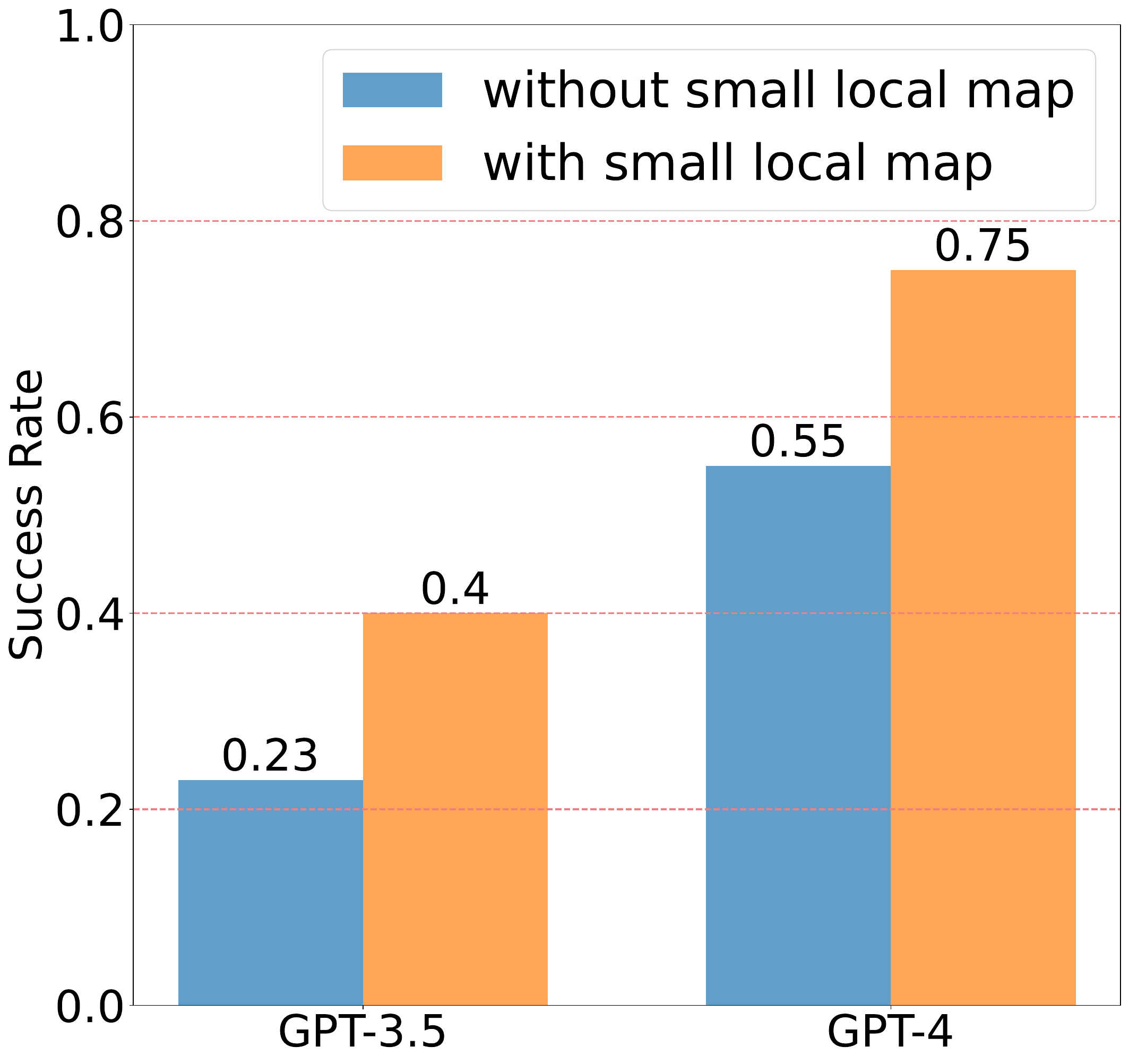}
\vspace{-2pt}
\caption{Playing minigames.}\label{fig:game_results}
\end{minipage}
\vspace{-8pt}
\end{figure}

For this task, we evaluated GPT-3.5 and GPT-4. 
For each model, we tried two settings: the first is to condition the LLM on the walkthrough like \cref{lst:walkthrough}; the second is to include in the prompt the information about the nearby locations, and an example of the full prompt is given in \cref{lst:playgameprompt}. 
The information about nearby locations is the ground-truth information that the LLM, in principle, should have learned from the walkthrough prefix. If the LLM had a perfect mapping and navigation ability, it would be able to perfectly spell it out and use that information to guide its decision making. \cref{fig:game_results} presents the results of this experiment. GPT-4 significantly outperforms GPT-3.5 in playing these minigames, consistent with their relative performance when answering the DF and RF questions of our MANGO benchmark. 
For each of the GPT models, having access to nearby location information significantly improves its performance, demonstrating that a strong mapping and navigation ability is essential to succeeding at relevant downstream tasks.

\section{Related Work}\label{sec:related}
Over the past few years, the field of natural language processing has experienced remarkable advancements with the emergence of large language models. 
This progress has spurred a multitude of research endeavors that propose benchmarks challenging the limits of these models. 
Those benchmarks assess the capacities of LLMs in linguistics~\citep{wang2018glue,superglue}, 
reading comprehension~\citep{richardson2013mctest,lai2017race}, 
commonsense reasoning~\citep{zellers2019hellaswag,bisk2020piqa,huang2019cosmos,talmor2018commonsenseqa}, 
arithmetic reasoning~\citep{miao2021diverse,cobbe2021training,patel2021nlp}, 
and knowledge memorization and understanding~\citep{clark2018think,mihaylov2018can,khot2020qasc,clark2020f,hendrycks2020measuring,srivastava2022beyond}. 
Recent models have achieved remarkable performance not only on these benchmarks, but also across a diversity of human-oriented academic and professional exams~\citep{gpt4} as well as general tasks~\citep{bubeck2023sparks}. %
Our benchmark presents a unique challenge to large language models, evaluating their capacity to acquire spatial 
knowledge about new environments and answering complex navigation questions; it is a dimension orthogonal to the aforementioned reasoning abilities. 

The advances of LLMs have sparked a recent wave of endeavors that integrate these models into embodied agents~\citep{huang2022inner,yang2023foundation,vemprala2023chatgpt,Wang2023VoyagerAO}. 
Generally, they utilize language models as a means to understand human instructions and plan executable actions~\citep{driess2023palm,liang2022code,huang2022language,pmlr-v205-ichter23a}. 
This includes instructions related to object manipulation and tool operation~\citep{wang2023programmatically,ren2023leveraging} as well as localization and navigation~\citep{majumdar2020improving,gadre2023cows,shah2023lm,Huang2022VisualLM}. 
Our MANGO benchmark aligns with the growing trend to deploy LLMs in embodied agents and provides a comprehensive investigation of their capacities in mapping and navigation. 
Our benchmark operates in text-based environments, distinguishing itself from previous benchmarks~\citep{puig2018virtualhome,shridhar2020alfred,fan2022minedojo} that allow agents to utilize visual signals. 
This ``text-only'' design enables us to conduct controlled experiments that investigate the capacity of language models to acquire knowledge about environments solely from textual inputs and answer navigation questions based on that knowledge. 
It complements the existing benchmark and methodological research in vision-language navigation~\citep{duvallet14, mei2016listen, Anderson2017VisionandLanguageNI, fried18, zhu20, Min2021FILMFI}. Our work is related to recent studies that demonstrate the emergence of maps with learned neural representations as a consequence of navigation~\citep{huynh20, wijmans2023emergence} with the key distinction that our agents are provided with textual descriptions of their environments.

Given our focus on mapping and navigation, it is worth noting the work on simultaneous localization and mapping (SLAM), a classic problem in which a mobile agent (e.g., a robot or hand-held camera) is tasked with mapping an a priori unknown environment while concurrently using its estimated map to localize itself in the environment~\citep{MurArtal2015ORBSlamaV, cadena16}. Particularly relevant are the methods that maintain spatial-semantic maps of the environments based on natural language descriptions~\citep{Walter2013LearningSM, hemachandra15a}, however they rely on non-linguistic observations (e.g., vision) to ground these descriptions.

\section{Conclusion}\label{sec:conclusion}
We present MANGO, a benchmark that evaluates the mapping and navigation abilities of large language models.
Our benchmark covers a diversity of mazes as well as a variety of evaluation and analysis programs, offering a comprehensive testbed in a great breadth and depth.
In our experiments, the current best model still performs poorly on the benchmark, with a sharp degradation on the more difficult questions. 
We release our benchmark---along with the source code for data generation and evaluation---to track the advances of the mapping and navigation capabilities of future LLMs as well as to facilitate future research in related areas. 
Several interesting future directions are discussed in \cref{app:future}.

\subsubsection*{Acknowledgments}
This work was supported by a research gift to the last author by Adobe Research. 
It was also supported by the TRI University 2.0 Project, the U.S.\ National Science Foundation (Grant Number CNS-1948457 and CNS-2340171), and Office of Advanced Scientific Computing Research at Department of Energy (Award Number DE-SC0024424). 
The views and opinions of authors expressed herein do not necessarily state or reflect those of the United States Government or any agency thereof. 
We thank the anonymous COLM reviewers and meta-reviewer for their constructive feedback. We also thank our colleagues at TTIC and UChicago for helpful discussion.

\bibliography{gamegpt}

\begin{thebibliography}{65}
\providecommand{\natexlab}[1]{#1}
\providecommand{\url}[1]{\texttt{#1}}
\expandafter\ifx\csname urlstyle\endcsname\relax
  \providecommand{\doi}[1]{doi: #1}\else
  \providecommand{\doi}{doi: \begingroup \urlstyle{rm}\Url}\fi

\bibitem[Anderson et~al.(2017)Anderson, Wu, Teney, Bruce, Johnson, S{\"u}nderhauf, Reid, Gould, and van~den Hengel]{Anderson2017VisionandLanguageNI}
Anderson, P., Wu, Q., Teney, D., Bruce, J., Johnson, M., S{\"u}nderhauf, N., Reid, I.~D., Gould, S., and van~den Hengel, A.
\newblock \href {https://arxiv.org/abs/1711.07280} {Vision-and-language navigation: {I}nterpreting visually-grounded navigation instructions in real environments}.
\newblock \emph{Proceedings of the IEEE Conference on Computer Vision and Pattern Recognition (CVPR)}, pp.\  3674--3683, 2017.

\bibitem[Anderson et~al.(2018)Anderson, Chang, Chaplot, Dosovitskiy, Gupta, Koltun, Kosecka, Malik, Mottaghi, Savva, et~al.]{anderson2018evaluation}
Anderson, P., Chang, A., Chaplot, D.~S., Dosovitskiy, A., Gupta, S., Koltun, V., Kosecka, J., Malik, J., Mottaghi, R., Savva, M., et~al.
\newblock \href {https://arxiv.org/abs/1807.06757} {On evaluation of embodied navigation agents}.
\newblock \emph{arXiv preprint arXiv:1807.06757}, 2018.

\bibitem[Anthopic(2023{\natexlab{a}})]{claude1}
Anthopic.
\newblock \href {https://www.anthropic.com/news/introducing-claude} {Introducing {C}laude}.
\newblock \url{https://www.anthropic.com/news/introducing-claude}, 2023{\natexlab{a}}.
\newblock Accessed: March 1, 2024.

\bibitem[Anthopic(2023{\natexlab{b}})]{claude2}
Anthopic.
\newblock \href {https://www-cdn.anthropic.com/bd2a28d2535bfb0494cc8e2a3bf135d2e7523226/Model-Card-Claude-2.pdf} {Model card and evaluations for {C}laude models}.
\newblock \url{https://www-cdn.anthropic.com/bd2a28d2535bfb0494cc8e2a3bf135d2e7523226/Model-Card-Claude-2.pdf}, 2023{\natexlab{b}}.
\newblock Accessed: March 1, 2024.

\bibitem[Bisk et~al.(2020)Bisk, Zellers, Gao, and Choi]{bisk2020piqa}
Bisk, Y., Zellers, R., Gao, J., and Choi, Y.
\newblock \href {https://arxiv.org/abs/1911.11641} {{PIQA}: {R}easoning about physical commonsense in natural language}.
\newblock In \emph{Proceedings of the AAAI Conference on Artificial Intelligence (AAAI)}, 2020.

\bibitem[Brown et~al.(2020)Brown, Mann, Ryder, Subbiah, Kaplan, Dhariwal, Neelakantan, Shyam, Sastry, Askell, Agarwal, Herbert-Voss, Krueger, Henighan, Child, Ramesh, Ziegler, Wu, Winter, Hesse, Chen, Sigler, Litwin, Gray, Chess, Clark, Berner, McCandlish, Radford, Sutskever, and Amodei]{brown-2020-gpt}
Brown, T., Mann, B., Ryder, N., Subbiah, M., Kaplan, J.~D., Dhariwal, P., Neelakantan, A., Shyam, P., Sastry, G., Askell, A., Agarwal, S., Herbert-Voss, A., Krueger, G., Henighan, T., Child, R., Ramesh, A., Ziegler, D., Wu, J., Winter, C., Hesse, C., Chen, M., Sigler, E., Litwin, M., Gray, S., Chess, B., Clark, J., Berner, C., McCandlish, S., Radford, A., Sutskever, I., and Amodei, D.
\newblock \href {https://papers.nips.cc/paper/2020/file/1457c0d6bfcb4967418bfb8ac142f64a-Paper.pdf} {Language models are few-shot learners}.
\newblock In \emph{Advances in Neural Information Processing Systems (NeurIPS)}, 2020.

\bibitem[Bubeck et~al.(2023)Bubeck, Chandrasekaran, Eldan, Gehrke, Horvitz, Kamar, Lee, Lee, Li, Lundberg, et~al.]{bubeck2023sparks}
Bubeck, S., Chandrasekaran, V., Eldan, R., Gehrke, J., Horvitz, E., Kamar, E., Lee, P., Lee, Y.~T., Li, Y., Lundberg, S., et~al.
\newblock \href {https://arxiv.org/abs/2303.12712.pdf} {Sparks of artificial general intelligence: {E}arly experiments with {GPT}-4}.
\newblock \emph{arXiv preprint arXiv:2303.12712}, 2023.

\bibitem[Cadena et~al.(2016)Cadena, Carlone, Carrillo, Latif, Scaramuzza, Neira, Reid, and Leonard]{cadena16}
Cadena, C., Carlone, L., Carrillo, H., Latif, Y., Scaramuzza, D., Neira, J., Reid, I., and Leonard, J.~J.
\newblock \href {https://arxiv.org/abs/1606.05830} {Past, present, and future of simultaneous localization and mapping: {T}oward the robust-perception age}.
\newblock \emph{IEEE Transactions on Robotics}, 2016.

\bibitem[Clark et~al.(2018)Clark, Cowhey, Etzioni, Khot, Sabharwal, Schoenick, and Tafjord]{clark2018think}
Clark, P., Cowhey, I., Etzioni, O., Khot, T., Sabharwal, A., Schoenick, C., and Tafjord, O.
\newblock \href {https://arxiv.org/abs/1803.05457} {Think you have solved question answering? {T}ry {ARC}, the {AI2} {R}easoning {C}hallenge}.
\newblock \emph{arXiv preprint arXiv:1803.05457}, 2018.

\bibitem[Clark et~al.(2020)Clark, Etzioni, Khot, Khashabi, Mishra, Richardson, Sabharwal, Schoenick, Tafjord, Tandon, et~al.]{clark2020f}
Clark, P., Etzioni, O., Khot, T., Khashabi, D., Mishra, B., Richardson, K., Sabharwal, A., Schoenick, C., Tafjord, O., Tandon, N., et~al.
\newblock \href {https://ojs.aaai.org/aimagazine/index.php/aimagazine/article/view/5304} {From \emph{F} to \emph{A} on the {NY} {R}egents {S}cience {E}xams: {A}n overview of the {Aristo} project}.
\newblock \emph{AI Magazine}, 2020.

\bibitem[Cobbe et~al.(2021)Cobbe, Kosaraju, Bavarian, Chen, Jun, Kaiser, Plappert, Tworek, Hilton, Nakano, et~al.]{cobbe2021training}
Cobbe, K., Kosaraju, V., Bavarian, M., Chen, M., Jun, H., Kaiser, L., Plappert, M., Tworek, J., Hilton, J., Nakano, R., et~al.
\newblock \href {https://arxiv.org/abs/2110.14168} {Training verifiers to solve math word problems}.
\newblock \emph{arXiv preprint arXiv:2110.14168}, 2021.

\bibitem[Driess et~al.(2023)Driess, Xia, Sajjadi, Lynch, Chowdhery, Ichter, Wahid, Tompson, Vuong, Yu, Huang, Chebotar, Sermanet, Duckworth, Levine, Vanhoucke, Hausman, Toussaint, Greff, Zeng, Mordatch, and Florence]{driess2023palm}
Driess, D., Xia, F., Sajjadi, M. S.~M., Lynch, C., Chowdhery, A., Ichter, B., Wahid, A., Tompson, J., Vuong, Q., Yu, T., Huang, W., Chebotar, Y., Sermanet, P., Duckworth, D., Levine, S., Vanhoucke, V., Hausman, K., Toussaint, M., Greff, K., Zeng, A., Mordatch, I., and Florence, P.
\newblock \href {https://arxiv.org/abs/2303.03378} {{PaLM}-{E}: {A}n embodied multimodal language model}.
\newblock \emph{arXiv preprint arXiv:2303.03378}, 2023.

\bibitem[Duvallet et~al.(2014)Duvallet, Walter, Howard, Hemachandra, Oh, Teller, Roy, and Stentz]{duvallet14}
Duvallet, F., Walter, M.~R., Howard, T., Hemachandra, S., Oh, J., Teller, S., Roy, N., and Stentz, A.
\newblock \href {https://link.springer.com/chapter/10.1007/978-3-319-23778-7_25} {Inferring maps and behaviors from natural language instructions}.
\newblock In \emph{Proceedings of the International Symposium on Experimental Robotics (ISER)}, 2014.

\bibitem[Epstein et~al.(2017)Epstein, Patai, Julian, and Spiers]{epstein17}
Epstein, R.~A., Patai, E.~Z., Julian, J.~B., and Spiers, H.~J.
\newblock \href {https://www.nature.com/articles/nn.4656} {The cognitive map in humans: spatial navigation and beyond}.
\newblock \emph{Nature Neuroscience}, \penalty0 (11):\penalty0 1504--1513, 2017.

\bibitem[Fan et~al.(2022)Fan, Wang, Jiang, Mandlekar, Yang, Zhu, Tang, Huang, Zhu, and Anandkumar]{fan2022minedojo}
Fan, L., Wang, G., Jiang, Y., Mandlekar, A., Yang, Y., Zhu, H., Tang, A., Huang, D.-A., Zhu, Y., and Anandkumar, A.
\newblock \href {https://arxiv.org/abs/2206.08853} {{MineDojo}: {B}uilding open-ended embodied agents with internet-scale knowledge}.
\newblock In \emph{Advances in Neural Information Processing Systems (NeurIPS)}, 2022.

\bibitem[Fried et~al.(2018)Fried, Hu, Cirik, Rohrbach, Andreas, Morency, Berg-Kirkpatrick, Saenko, Klein, and Darrell]{fried18}
Fried, D., Hu, R., Cirik, V., Rohrbach, A., Andreas, J., Morency, L.-P., Berg-Kirkpatrick, T., Saenko, K., Klein, D., and Darrell, T.
\newblock \href {https://arxiv.org/abs/1806.02724} {Speaker-follower models for vision-and-language navigation}.
\newblock In \emph{Advances in Neural Information Processing Systems (NeurIPS)}, December 2018.

\bibitem[Gadre et~al.(2023)Gadre, Wortsman, Ilharco, Schmidt, and Song]{gadre2023cows}
Gadre, S.~Y., Wortsman, M., Ilharco, G., Schmidt, L., and Song, S.
\newblock \href {https://arxiv.org/abs/2203.10421} {Cows on pasture: Baselines and benchmarks for language-driven zero-shot object navigation}.
\newblock In \emph{Proceedings of the IEEE Conference on Computer Vision and Pattern Recognition (CVPR)}, 2023.

\bibitem[Gao et~al.(2022)Gao, Schulman, and Hilton]{gao2022scaling}
Gao, L., Schulman, J., and Hilton, J.
\newblock \href {https://arxiv.org/abs/2210.10760} {Scaling laws for reward model overoptimization}.
\newblock \emph{arXiv preprint arXiv:2210.10760}, 2022.

\bibitem[Hausknecht et~al.(2020)Hausknecht, Ammanabrolu, C{\^o}t{\'e}, and Yuan]{hausknecht2020interactive}
Hausknecht, M., Ammanabrolu, P., C{\^o}t{\'e}, M.-A., and Yuan, X.
\newblock \href {https://ojs.aaai.org/index.php/AAAI/article/view/6297/6153} {Interactive fiction games: A colossal adventure}.
\newblock In \emph{Proceedings of the AAAI Conference on Artificial Intelligence (AAAI)}, 2020.

\bibitem[Hemachandra \& Walter(2015)Hemachandra and Walter]{hemachandra15a}
Hemachandra, S. and Walter, M.~R.
\newblock \href {https://ieeexplore.ieee.org/document/7354097} {Information-theoretic dialog to improve spatial-semantic representations}.
\newblock In \emph{Proceedings of the IEEE/RSJ International Conference on Intelligent Robots and Systems (IROS)}, 2015.

\bibitem[Hendrycks et~al.(2021)Hendrycks, Burns, Basart, Zou, Mazeika, Song, and Steinhardt]{hendrycks2020measuring}
Hendrycks, D., Burns, C., Basart, S., Zou, A., Mazeika, M., Song, D., and Steinhardt, J.
\newblock \href {https://arxiv.org/abs/2009.03300} {Measuring massive multitask language understanding}.
\newblock In \emph{Proceedings of the International Conference on Learning Representations (ICLR)}, 2021.

\bibitem[Huang et~al.(2022{\natexlab{a}})Huang, Mees, Zeng, and Burgard]{Huang2022VisualLM}
Huang, C., Mees, O., Zeng, A., and Burgard, W.
\newblock \href {https://arxiv.org/abs/2210.05714} {Visual language maps for robot navigation}.
\newblock \emph{arXiv preprint arXiv:2210.05714}, 2022{\natexlab{a}}.

\bibitem[Huang et~al.(2019)Huang, Bras, Bhagavatula, and Choi]{huang2019cosmos}
Huang, L., Bras, R.~L., Bhagavatula, C., and Choi, Y.
\newblock \href {https://arxiv.org/abs/1909.00277} {Cosmos {QA}: {M}achine reading comprehension with contextual commonsense reasoning}.
\newblock In \emph{Proceedings of the Conference on Empirical Methods in Natural Language Processing (EMNLP)}, 2019.

\bibitem[Huang et~al.(2022{\natexlab{b}})Huang, Abbeel, Pathak, and Mordatch]{huang2022language}
Huang, W., Abbeel, P., Pathak, D., and Mordatch, I.
\newblock \href {https://arxiv.org/abs/2201.07207} {Language models as zero-shot planners: Extracting actionable knowledge for embodied agents}.
\newblock In \emph{Proceedings of the International Conference on Machine Learning (ICML)}, 2022{\natexlab{b}}.

\bibitem[Huang et~al.(2022{\natexlab{c}})Huang, Xia, Xiao, Chan, Liang, Florence, Zeng, Tompson, Mordatch, Chebotar, et~al.]{huang2022inner}
Huang, W., Xia, F., Xiao, T., Chan, H., Liang, J., Florence, P., Zeng, A., Tompson, J., Mordatch, I., Chebotar, Y., et~al.
\newblock \href {https://arxiv.org/abs/2207.05608} {Inner monologue: Embodied reasoning through planning with language models}.
\newblock \emph{arXiv preprint arXiv:2207.05608}, 2022{\natexlab{c}}.

\bibitem[Huynh et~al.(2020)Huynh, Maire, and Walter]{huynh20}
Huynh, T., Maire, M., and Walter, M.~R.
\newblock \href {https://arxiv.org/abs/1906.05948} {Multigrid neural memory}.
\newblock In \emph{Proceedings of the International Conference on Machine Learning (ICML)}, 2020.

\bibitem[Ichter et~al.(2023)Ichter, Brohan, Chebotar, Finn, Hausman, Herzog, Ho, Ibarz, Irpan, Jang, Julian, Kalashnikov, Levine, Lu, Parada, Rao, Sermanet, Toshev, Vanhoucke, Xia, Xiao, Xu, Yan, Brown, Ahn, Cortes, Sievers, Tan, Xu, Reyes, Rettinghouse, Quiambao, Pastor, Luu, Lee, Kuang, Jesmonth, Joshi, Jeffrey, Ruano, Hsu, Gopalakrishnan, David, Zeng, and Fu]{pmlr-v205-ichter23a}
Ichter, b., Brohan, A., Chebotar, Y., Finn, C., Hausman, K., Herzog, A., Ho, D., Ibarz, J., Irpan, A., Jang, E., Julian, R., Kalashnikov, D., Levine, S., Lu, Y., Parada, C., Rao, K., Sermanet, P., Toshev, A.~T., Vanhoucke, V., Xia, F., Xiao, T., Xu, P., Yan, M., Brown, N., Ahn, M., Cortes, O., Sievers, N., Tan, C., Xu, S., Reyes, D., Rettinghouse, J., Quiambao, J., Pastor, P., Luu, L., Lee, K.-H., Kuang, Y., Jesmonth, S., Joshi, N.~J., Jeffrey, K., Ruano, R.~J., Hsu, J., Gopalakrishnan, K., David, B., Zeng, A., and Fu, C.~K.
\newblock \href {https://proceedings.mlr.press/v205/ichter23a.html} {Do as {I} can, not as {I} say: {G}rounding language in robotic affordances}.
\newblock In \emph{Proceedings of the Conference on Robot Learning (CoRL)}, 2023.

\bibitem[Javadi et~al.(2017)Javadi, Emo, Howard, Zisch, Yu, Knight, Pinelo~Silva, and Spiers]{javadi17}
Javadi, A.-H., Emo, B., Howard, L.~R., Zisch, F.~E., Yu, Y., Knight, R., Pinelo~Silva, J., and Spiers, H.~J.
\newblock \href {https://www.nature.com/articles/ncomms14652} {Hippocampal and prefrontal processing of network topology to simulate the future}.
\newblock \emph{Nature Communications}, 2017.

\bibitem[Khot et~al.(2020)Khot, Clark, Guerquin, Jansen, and Sabharwal]{khot2020qasc}
Khot, T., Clark, P., Guerquin, M., Jansen, P., and Sabharwal, A.
\newblock \href {https://arxiv.org/abs/1910.11473} {Qasc: A dataset for question answering via sentence composition}.
\newblock In \emph{Proceedings of the AAAI Conference on Artificial Intelligence (AAAI)}, 2020.

\bibitem[Lai et~al.(2017)Lai, Xie, Liu, Yang, and Hovy]{lai2017race}
Lai, G., Xie, Q., Liu, H., Yang, Y., and Hovy, E.
\newblock \href {https://arxiv.org/abs/1704.04683} {Race: Large-scale reading comprehension dataset from examinations}.
\newblock In \emph{Proceedings of the Conference on Empirical Methods in Natural Language Processing (EMNLP)}, 2017.

\bibitem[Li et~al.(2023)Li, Hopkins, Bau, Vi{\'e}gas, Pfister, and Wattenberg]{li2022emergent}
Li, K., Hopkins, A.~K., Bau, D., Vi{\'e}gas, F., Pfister, H., and Wattenberg, M.
\newblock \href {https://arxiv.org/abs/2210.13382} {Emergent world representations: Exploring a sequence model trained on a synthetic task}.
\newblock In \emph{Proceedings of the International Conference on Learning Representations (ICLR)}, 2023.

\bibitem[Liang et~al.(2022)Liang, Huang, Xia, Xu, Hausman, Ichter, Florence, and Zeng]{liang2022code}
Liang, J., Huang, W., Xia, F., Xu, P., Hausman, K., Ichter, B., Florence, P., and Zeng, A.
\newblock \href {https://arxiv.org/abs/2209.07753} {Code as policies: Language model programs for embodied control}.
\newblock \emph{arXiv preprint arXiv:2209.07753}, 2022.

\bibitem[Majumdar et~al.(2020)Majumdar, Shrivastava, Lee, Anderson, Parikh, and Batra]{majumdar2020improving}
Majumdar, A., Shrivastava, A., Lee, S., Anderson, P., Parikh, D., and Batra, D.
\newblock \href {https://arxiv.org/abs/2004.14973} {Improving vision-and-language navigation with image-text pairs from the {W}eb}.
\newblock In \emph{Proceedings of the European Conference on Computer Vision (ECCV)}, 2020.

\bibitem[Mei et~al.(2016)Mei, Bansal, and Walter]{mei2016listen}
Mei, H., Bansal, M., and Walter, M.
\newblock \href {https://arxiv.org/abs/1506.04089} {Listen, attend, and walk: Neural mapping of navigational instructions to action sequences}.
\newblock In \emph{Proceedings of the AAAI Conference on Artificial Intelligence (AAAI)}, 2016.

\bibitem[Miao et~al.(2020)Miao, Liang, and Su]{miao2021diverse}
Miao, S.-Y., Liang, C.-C., and Su, K.-Y.
\newblock \href {https://arxiv.org/abs/2106.15772} {A diverse corpus for evaluating and developing english math word problem solvers}.
\newblock In \emph{Proceedings of the Annual Meeting of the Association for Computational Linguistics (ACL)}, 2020.

\bibitem[Mihaylov et~al.(2018)Mihaylov, Clark, Khot, and Sabharwal]{mihaylov2018can}
Mihaylov, T., Clark, P., Khot, T., and Sabharwal, A.
\newblock \href {https://arxiv.org/abs/1809.02789} {Can a suit of armor conduct electricity? a new dataset for open book question answering}.
\newblock In \emph{Proceedings of the Conference on Empirical Methods in Natural Language Processing (EMNLP)}, 2018.

\bibitem[Min et~al.(2021)Min, Chaplot, Ravikumar, Bisk, and Salakhutdinov]{Min2021FILMFI}
Min, S.~Y., Chaplot, D.~S., Ravikumar, P., Bisk, Y., and Salakhutdinov, R.
\newblock \href {https://arxiv.org/abs/2110.07342} {{FILM}: {F}ollowing instructions in language with modular methods}.
\newblock \emph{arXiv preprint arXiv:2110.07342}, 2021.

\bibitem[Mur-Artal et~al.(2015)Mur-Artal, Montiel, and Tard{\'o}s]{MurArtal2015ORBSlamaV}
Mur-Artal, R., Montiel, J. M.~M., and Tard{\'o}s, J.~D.
\newblock \href {https://ieeexplore.ieee.org/document/7219438} {{ORB-SLAM}: {A} versatile and accurate monocular {SLAM} system}.
\newblock \emph{IEEE Transactions on Robotics}, 31:\penalty0 1147--1163, 2015.

\bibitem[OpenAI(2023)]{gpt4}
OpenAI.
\newblock \href {https://arxiv.org/abs/2303.08774.pdf} {{GPT}-4 technical report}.
\newblock \emph{arXiv preprint arXiv:2303.08774}, 2023.

\bibitem[Patel et~al.(2021)Patel, Bhattamishra, and Goyal]{patel2021nlp}
Patel, A., Bhattamishra, S., and Goyal, N.
\newblock \href {https://arxiv.org/abs/2103.07191} {Are nlp models really able to solve simple math word problems?}
\newblock In \emph{Proceedings of the Conference of the North American Chapter of the Association for Computational Linguistics (NAACL)}, 2021.

\bibitem[Peng et~al.(2023)Peng, Alcaide, Anthony, Albalak, Arcadinho, Cao, Cheng, Chung, Grella, GV, He, Hou, Kazienko, Kocon, Kong, Koptyra, Lau, Mantri, Mom, Saito, Tang, Wang, Wind, Wozniak, Zhang, Zhang, Zhao, Zhou, Zhu, and Zhu]{peng2023rwkv}
Peng, B., Alcaide, E., Anthony, Q., Albalak, A., Arcadinho, S., Cao, H., Cheng, X., Chung, M., Grella, M., GV, K.~K., He, X., Hou, H., Kazienko, P., Kocon, J., Kong, J., Koptyra, B., Lau, H., Mantri, K. S.~I., Mom, F., Saito, A., Tang, X., Wang, B., Wind, J.~S., Wozniak, S., Zhang, R., Zhang, Z., Zhao, Q., Zhou, P., Zhu, J., and Zhu, R.-J.
\newblock \href {https://arxiv.org/abs/2305.13048} {{RWKV}: {R}einventing {RNNs} for the transformer era}.
\newblock \emph{arXiv preprint arXiv:2305.13048}, 2023.

\bibitem[Puig et~al.(2018)Puig, Ra, Boben, Li, Wang, Fidler, and Torralba]{puig2018virtualhome}
Puig, X., Ra, K., Boben, M., Li, J., Wang, T., Fidler, S., and Torralba, A.
\newblock \href {https://arxiv.org/abs/1806.07011} {Virtualhome: Simulating household activities via programs}.
\newblock In \emph{Proceedings of the IEEE Conference on Computer Vision and Pattern Recognition (CVPR)}, 2018.

\bibitem[Ren et~al.(2023)Ren, Govil, Yang, Narasimhan, and Majumdar]{ren2023leveraging}
Ren, A.~Z., Govil, B., Yang, T.-Y., Narasimhan, K.~R., and Majumdar, A.
\newblock \href {https://arxiv.org/abs/2206.13074} {Leveraging language for accelerated learning of tool manipulation}.
\newblock In \emph{Proceedings of the Conference on Robot Learning (CoRL)}, 2023.

\bibitem[Richardson et~al.(2013)Richardson, Burges, and Renshaw]{richardson2013mctest}
Richardson, M., Burges, C.~J., and Renshaw, E.
\newblock \href {https://aclanthology.org/D13-1020/} {{MCTest}: {A} challenge dataset for the open-domain machine comprehension of text}.
\newblock In \emph{Proceedings of the Conference on Empirical Methods in Natural Language Processing (EMNLP)}, 2013.

\bibitem[Rozière et~al.(2024)Rozière, Gehring, Gloeckle, Sootla, Gat, Tan, Adi, Liu, Sauvestre, Remez, Rapin, Kozhevnikov, Evtimov, Bitton, Bhatt, Ferrer, Grattafiori, Xiong, Défossez, Copet, Azhar, Touvron, Martin, Usunier, Scialom, and Synnaeve]{codellama}
Rozière, B., Gehring, J., Gloeckle, F., Sootla, S., Gat, I., Tan, X.~E., Adi, Y., Liu, J., Sauvestre, R., Remez, T., Rapin, J., Kozhevnikov, A., Evtimov, I., Bitton, J., Bhatt, M., Ferrer, C.~C., Grattafiori, A., Xiong, W., Défossez, A., Copet, J., Azhar, F., Touvron, H., Martin, L., Usunier, N., Scialom, T., and Synnaeve, G.
\newblock \href {https://arxiv.org/abs/2308.12950} {Code llama: Open foundation models for code}.
\newblock \emph{arXiv preprint arXiv:2308.12950}, 2024.

\bibitem[Shah et~al.(2023)Shah, Osi{\'n}ski, Levine, et~al.]{shah2023lm}
Shah, D., Osi{\'n}ski, B., Levine, S., et~al.
\newblock \href {https://arxiv.org/abs/2207.04429} {{LM-Nav}: {R}obotic navigation with large pre-trained models of language, vision, and action}.
\newblock In \emph{Proceedings of the Conference on Robot Learning (CoRL)}, 2023.

\bibitem[Shridhar et~al.(2020)Shridhar, Thomason, Gordon, Bisk, Han, Mottaghi, Zettlemoyer, and Fox]{shridhar2020alfred}
Shridhar, M., Thomason, J., Gordon, D., Bisk, Y., Han, W., Mottaghi, R., Zettlemoyer, L., and Fox, D.
\newblock \href {https://arxiv.org/abs/1912.01734} {Alfred: A benchmark for interpreting grounded instructions for everyday tasks}.
\newblock In \emph{Proceedings of the IEEE Conference on Computer Vision and Pattern Recognition (CVPR)}, 2020.

\bibitem[Spiers \& Gilbert(2015)Spiers and Gilbert]{spiers15}
Spiers, H.~J. and Gilbert, S.~J.
\newblock \href {https://pubmed.ncbi.nlm.nih.gov/25852515/} {Solving the detour problem in navigation: {A} model of prefrontal and hippocampal interactions}.
\newblock \emph{Frontiers in Human Neuroscience}, 2015.

\bibitem[Spiers \& Maguire(2006)Spiers and Maguire]{spiers06}
Spiers, H.~J. and Maguire, E.~A.
\newblock \href {https://pubmed.ncbi.nlm.nih.gov/16584892/} {Thoughts, behaviour, and brain dynamics during navigation in the real world}.
\newblock \emph{Neuroimage}, 31\penalty0 (4):\penalty0 1826--1840, 2006.

\bibitem[Srivastava et~al.(2022)Srivastava, Rastogi, Rao, Shoeb, Abid, Fisch, Brown, Santoro, Gupta, Garriga-Alonso, et~al.]{srivastava2022beyond}
Srivastava, A., Rastogi, A., Rao, A., Shoeb, A. A.~M., Abid, A., Fisch, A., Brown, A.~R., Santoro, A., Gupta, A., Garriga-Alonso, A., et~al.
\newblock \href {https://arxiv.org/abs/2206.04615} {Beyond the imitation game: Quantifying and extrapolating the capabilities of language models}.
\newblock \emph{Transactions of Machine Learning Research}, 2022.

\bibitem[Stiennon et~al.(2020)Stiennon, Ouyang, Wu, Ziegler, Lowe, Voss, Radford, Amodei, and Christiano]{stiennon2020learning}
Stiennon, N., Ouyang, L., Wu, J., Ziegler, D., Lowe, R., Voss, C., Radford, A., Amodei, D., and Christiano, P.~F.
\newblock \href {https://arxiv.org/abs/2009.01325} {Learning to summarize with human feedback}.
\newblock In \emph{Advances in Neural Information Processing Systems (NeurIPS)}, 2020.

\bibitem[Talmor et~al.(2019)Talmor, Herzig, Lourie, and Berant]{talmor2018commonsenseqa}
Talmor, A., Herzig, J., Lourie, N., and Berant, J.
\newblock \href {https://arxiv.org/abs/1811.00937} {{CommonsenseQA}: {A} question answering challenge targeting commonsense knowledge}.
\newblock In \emph{Proceedings of the Conference of the North American Chapter of the Association for Computational Linguistics (NAACL)}, 2019.

\bibitem[Touvron et~al.(2023{\natexlab{a}})Touvron, Lavril, Izacard, Martinet, Lachaux, Lacroix, Rozi{\`e}re, Goyal, Hambro, Azhar, Rodriguez, Joulin, Grave, and Lample]{touvron2023llama}
Touvron, H., Lavril, T., Izacard, G., Martinet, X., Lachaux, M.-A., Lacroix, T., Rozi{\`e}re, B., Goyal, N., Hambro, E., Azhar, F., Rodriguez, A., Joulin, A., Grave, E., and Lample, G.
\newblock \href {https://arxiv.org/abs/2302.13971} {{LLaMA}: {O}pen and efficient foundation language models}.
\newblock \emph{arXiv preprint arXiv:2302.13971}, 2023{\natexlab{a}}.

\bibitem[Touvron et~al.(2023{\natexlab{b}})Touvron, Martin, Stone, Albert, Almahairi, Babaei, Bashlykov, Batra, Bhargava, Bhosale, et~al.]{touvron2023llama2}
Touvron, H., Martin, L., Stone, K., Albert, P., Almahairi, A., Babaei, Y., Bashlykov, N., Batra, S., Bhargava, P., Bhosale, S., et~al.
\newblock \href {https://arxiv.org/abs/2307.09288} {Llama 2: Open foundation and fine-tuned chat models}.
\newblock \emph{arXiv preprint arXiv:2307.09288}, 2023{\natexlab{b}}.

\bibitem[Vemprala et~al.(2023)Vemprala, Bonatti, Bucker, and Kapoor]{vemprala2023chatgpt}
Vemprala, S., Bonatti, R., Bucker, A., and Kapoor, A.
\newblock \href {https://arxiv.org/abs/2306.17582} {{ChatGPT} for robotics: {D}esign principles and model abilities}.
\newblock Technical Report MSR-TR-2023-8, Microsoft, February 2023.

\bibitem[Walter et~al.(2013)Walter, Hemachandra, Homberg, Tellex, and Teller]{Walter2013LearningSM}
Walter, M.~R., Hemachandra, S., Homberg, B., Tellex, S., and Teller, S.~J.
\newblock \href {https://www.roboticsproceedings.org/rss09/p04.pdf} {Learning semantic maps from natural language descriptions}.
\newblock \emph{The International Journal of Robotics Research}, 2013.

\bibitem[Wang et~al.(2018)Wang, Singh, Michael, Hill, Levy, and Bowman]{wang2018glue}
Wang, A., Singh, A., Michael, J., Hill, F., Levy, O., and Bowman, S.~R.
\newblock \href {https://arxiv.org/pdf/1804.07461.pdf} {{GLUE}: {A} multi-task benchmark and analysis platform for natural language understanding}.
\newblock \emph{arXiv preprint arXiv:1804.07461}, 2018.

\bibitem[Wang et~al.(2019)Wang, Pruksachatkun, Nangia, Singh, Michael, Hill, Levy, and Bowman]{superglue}
Wang, A., Pruksachatkun, Y., Nangia, N., Singh, A., Michael, J., Hill, F., Levy, O., and Bowman, S.~R.
\newblock \href {https://arxiv.org/pdf/1905.00537.pdf} {Super{GLUE}: A stickier benchmark for general-purpose language understanding systems}.
\newblock \emph{arXiv preprint 1905.00537}, 2019.

\bibitem[Wang et~al.(2023{\natexlab{a}})Wang, Xie, Jiang, Mandlekar, Xiao, Zhu, Fan, and Anandkumar]{Wang2023VoyagerAO}
Wang, G., Xie, Y., Jiang, Y., Mandlekar, A., Xiao, C., Zhu, Y., Fan, L., and Anandkumar, A.
\newblock \href {https://arxiv.org/abs/2305.16291} {Voyager: {A}n open-ended embodied agent with large language models}.
\newblock \emph{arXiv preprint arXiv:2305.16291}, 2023{\natexlab{a}}.

\bibitem[Wang et~al.(2023{\natexlab{b}})Wang, Mao, Hsu, Zhao, Wu, and Gao]{wang2023programmatically}
Wang, R., Mao, J., Hsu, J., Zhao, H., Wu, J., and Gao, Y.
\newblock \href {https://arxiv.org/abs/2304.13826} {Programmatically grounded, compositionally generalizable robotic manipulation}.
\newblock In \emph{Proceedings of the International Conference on Learning Representations (ICLR)}, 2023{\natexlab{b}}.

\bibitem[Wei et~al.(2022)Wei, Wang, Schuurmans, Bosma, Chi, Le, and Zhou]{wei2022chain}
Wei, J., Wang, X., Schuurmans, D., Bosma, M., Chi, E., Le, Q., and Zhou, D.
\newblock \href {https://arxiv.org/abs/2201.11903.pdf} {Chain of thought prompting elicits reasoning in large language models}.
\newblock In \emph{Advances in Neural Information Processing Systems (NeurIPS)}, 2022.

\bibitem[Wijmans et~al.(2023)Wijmans, Savva, Essa, Lee, Morcos, and Batra]{wijmans2023emergence}
Wijmans, E., Savva, M., Essa, I., Lee, S., Morcos, A.~S., and Batra, D.
\newblock \href {https://arxiv.org/abs/2301.13261} {Emergence of maps in the memories of blind navigation agents}.
\newblock In \emph{Proceedings of the International Conference on Learning Representations (ICLR)}, 2023.

\bibitem[Yang et~al.(2023)Yang, Nachum, Du, Wei, Abbeel, and Schuurmans]{yang2023foundation}
Yang, S., Nachum, O., Du, Y., Wei, J., Abbeel, P., and Schuurmans, D.
\newblock \href {https://arxiv.org/abs/2303.04129} {Foundation models for decision making: Problems, methods, and opportunities}.
\newblock \emph{arXiv preprint arXiv:2303.04129}, 2023.

\bibitem[Zellers et~al.(2019)Zellers, Holtzman, Bisk, Farhadi, and Choi]{zellers2019hellaswag}
Zellers, R., Holtzman, A., Bisk, Y., Farhadi, A., and Choi, Y.
\newblock \href {https://arxiv.org/abs/1905.07830} {{HellaSwag}: {C}an a machine really finish your sentence?}
\newblock In \emph{Proceedings of the Annual Meeting of the Association for Computational Linguistics (ACL)}, 2019.

\bibitem[Zhu et~al.(2020)Zhu, Zhu, Chang, and Liang]{zhu20}
Zhu, F., Zhu, Y., Chang, X., and Liang, X.
\newblock \href {https://arxiv.org/abs/1911.07883} {Vision-language navigation with self-supervised auxiliary reasoning tasks}.
\newblock In \emph{Proceedings of the IEEE Conference on Computer Vision and Pattern Recognition (CVPR)}, June 2020.

\end{thebibliography}
\bibliographystyle{icml2020_url}

\newpage
\appendix

\section{Benchmark Details}\label{app:data}\label{app:benchmark}
Our data and program is released at 
{\small \url{\weburl}} and {\small \url{\giturl}}. 
In the \code{data} folder, each game has a folder that contains multiple JSON files. 
The most important files are the DF and RF skeletons (where \code{X} is the game name such as \code{zork} or \code{detective}): 
\begin{itemize}
    \item \code{X.df}: it contains the DF skeletons like \cref{lst:dfqraw}. 
    \item \code{X.rf}: it contains the RF skeletons like \cref{lst:rfqraw}. 
\end{itemize}
As introduced in \cref{sec:qg}, each skeleton is paired with an ANSWERABLE label and an EASY label: given a prefix of the walkthrough, the ANSWERABLE label indicates whether this skeleton is answerable, and the EASY label will decide whether this skeleton is easy or hard given this walkthrough prefix. 
\cref{tab:fulldatastat} displays some important statistics of the full dataset broken down into each individual maze, including the number of DF and RF skeletons. 
In total, our full dataset has about 3M DF questions and 200K RF questions. 
On average, each maze has around 60K DF questions and 4K RF questions. 
Noticeably, the 53 games in our dataset are very diverse: they cover a range of topics and genres; they are situated in different eras; they cover a wide variety of maps (small vs.\ big houses, long vs.\ short halls, towns vs.\ forests, verbose vs.\ concise scene descriptions, etc). 
In addition, we also provide the following data files for easy reference: 
\begin{itemize}[leftmargin=*]
    \item \code{X.walkthrough}: it contains the full walkthrough of the game. See \cref{app:walkthrough} for details. 
    \item \code{X.locations}: it lists all the locations. %
    Details about annotating these locations can be found in \cref{app:location}. 
    \item \code{X.moves}: it lists all the moves that may change the location. 
    \item \code{X.all\_pairs}: it contains all the pairs of distinct locations.  
    \item \code{X.all2all}: it contains all the simple paths between any pair of distinct locations. 
\end{itemize}
\begin{table}[htbp]
\small
\setlength{\tabcolsep}{4pt}
\begin{sc}
\begin{subtable}{1.0\linewidth}
\begin{center}
\begin{tabular}{lrrrrrrrr}
\toprule
\multirow{2}{*}{Maps} & \multirow{2}{*}{\# Locs} & \multirow{2}{*}{\# Edges} & \multirow{2}{*}{Avg Len Path} & \multirow{2}{*}{\# Steps} & \multicolumn{2}{c}{DF} & \multicolumn{2}{c}{RF} \\
\cmidrule(lr){6-7} \cmidrule(lr){8-9}
& & & & &Easy & Hard & Easy & Hard \\
\midrule
905 & 5 & 7 & 1.88 & 22 & 11 & 5 & 11 & 5 \\
advent & 70 & 137 & 19.38 & 277 & 24788 & 4220 & 4694 & 0 \\
adventureland & 23 & 48 & 7.46 & 170 & 1117 & 208 & 484 & 22 \\
afflicted & 13 & 24 & 3.18 & 99 & 156 & 0 & 156 & 0 \\
anchor & 118 & 228 & 27.46 & 531 & 66133 & 23564 & 13573 & 0 \\
awaken & 15 & 28 & 5.02 & 57 & 365 & 45 & 171 & 25 \\
balances & 12 & 20 & 3.22 & 122 & 136 & 1 & 94 & 1 \\
ballyhoo & 39 & 90 & 9.26 & 416 & 18148 & 6298 & 1444 & 0 \\
curses & 150 & 304 & 33.83 & 816 & 1951931 & 549433 & 22201 & 0 \\
cutthroat & 52 & 108 & 10.76 & 336 & 5969 & 2688 & 1832 & 59 \\
deephome & 66 & 127 & 8.64 & 327 & 4161 & 235 & 4161 & 1 \\
detective & 32 & 41 & 8.76 & 51 & 505 & 6 & 505 & 6 \\
dragon & 24 & 52 & 8.9 & 101 & 1100 & 2262 & 529 & 23 \\
enchanter & 57 & 110 & 10.93 & 265 & 5118 & 2602 & 2773 & 99 \\
enter & 18 & 34 & 3.88 & 102 & 306 & 0 & 306 & 0 \\
gold & 22 & 44 & 4.99 & 345 & 682 & 0 & 462 & 0 \\
hhgg & 42 & 65 & 7.02 & 361 & 3794 & 3 & 1602 & 1 \\
hollywood & 109 & 219 & 27.33 & 397 & 21641 & 15557 & 11451 & 213 \\
huntdark & 12 & 13 & 4.24 & 67 & 66 & 2 & 66 & 2 \\
infidel & 58 & 122 & 9.27 & 250 & 2786 & 1729 & 2158 & 453 \\
inhumane & 43 & 91 & 6.49 & 122 & 1623 & 3020 & 1474 & 332 \\
jewel & 43 & 74 & 8.66 & 223 & 1157 & 61 & 1157 & 61 \\
karn & 56 & 124 & 15.36 & 362 & 24479 & 108983 & 3025 & 0 \\
library & 7 & 12 & 2.48 & 52 & 42 & 0 & 42 & 0 \\
loose & 12 & 21 & 4.18 & 50 & 94 & 27 & 94 & 27 \\
lostpig & 13 & 26 & 4.5 & 146 & 492 & 105 & 114 & 0 \\
ludicorp & 86 & 176 & 14.96 & 364 & 7099 & 7281 & 6002 & 1308 \\
lurking & 60 & 116 & 13.04 & 294 & 5187 & 2316 & 2946 & 105 \\
moonlit & 6 & 9 & 2.2 & 59 & 18 & 7 & 18 & 7 \\
murdac & 84 & 157 & 11.05 & 304 & 6914 & 2016 & 5967 & 1005 \\
night & 20 & 41 & 6.93 & 90 & 633 & 59 & 380 & 0 \\
omniquest & 32 & 65 & 7.55 & 78 & 648 & 1431 & 642 & 90 \\
partyfoul & 4 & 8 & 1.92 & 56 & 24 & 0 & 12 & 0 \\
pentari & 18 & 31 & 3.76 & 49 & 208 & 20 & 208 & 4 \\
planetfall & 69 & 138 & 12.41 & 399 & 7100 & 2246 & 3887 & 247 \\
plundered & 45 & 87 & 13.46 & 189 & 3705 & 2394 & 1393 & 26 \\
reverb & 17 & 31 & 5.26 & 74 & 321 & 20 & 256 & 16 \\
seastalker & 20 & 43 & 7.55 & 204 & 1123 & 1143 & 260 & 3 \\
sherlock & 71 & 140 & 12.39 & 339 & 16417 & 11758 & 4214 & 27 \\
snacktime & 4 & 6 & 1.5 & 34 & 12 & 0 & 12 & 0 \\
sorcerer & 64 & 120 & 11.72 & 254 & 4042 & 2285 & 2503 & 264 \\
spellbrkr & 73 & 111 & 13.62 & 412 & 8713 & 1762 & 3462 & 1 \\
spirit & 229 & 466 & 26.39 & 1264 & 111102 & 47410 & 51531 & 226 \\
temple & 24 & 46 & 6.07 & 181 & 563 & 106 & 486 & 21 \\
theatre & 20 & 38 & 5.33 & 296 & 329 & 130 & 329 & 51 \\
trinity & 94 & 189 & 20.39 & 610 & 65049 & 131324 & 8742 & 0 \\
tryst205 & 73 & 139 & 10.04 & 518 & 5089 & 392 & 3983 & 181 \\
wishbringer & 46 & 94 & 13.32 & 184 & 4988 & 5646 & 2070 & 0 \\
yomomma & 9 & 25 & 3.55 & 98 & 189 & 263 & 64 & 0 \\
zenon & 17 & 32 & 4.85 & 83 & 144 & 128 & 144 & 128 \\
zork1 & 84 & 166 & 20.79 & 396 & 20984 & 26857 & 6889 & 0 \\
zork2 & 66 & 136 & 11.79 & 296 & 15174 & 13081 & 3798 & 178 \\
zork3 & 55 & 105 & 15.37 & 273 & 6202 & 2754 & 2294 & 133 \\
ztuu & 16 & 30 & 3.27 & 84 & 225 & 15 & 225 & 15 \\
\bottomrule
\end{tabular}
\end{center}
\end{subtable}%
\end{sc}
\caption{Statistics of full data. 
Here, \textsc{\# Locs} represents the number of locations in each walkthrough; \textsc{\# Edges} represents the number of edges; \textsc{AVG LEN PATH} denotes the average length of all paths; \textsc{\# STEPS} indicates the number of steps in each walkthrough; \textsc{EASY} and \textsc{HARD} of the \textsc{DF} and \textsc{RF} respectively represent the number of easy and hard skeletons of the DF and RF tasks.
When counting the easy and hard skeletons, we assume that the full walkthrough will be used. For the statistics of the data used in our experiments, please see 
\cref{tab:expdatastatgpt,tab:expdatastatllama-2,tab:expdatastatllama-1,tab:expdatastatother}.
}
\label{tab:fulldatastat}
\end{table}

In the following subsections, we will explain the details about collecting this data.

\subsection{Walkthrough Details}\label{app:walkthrough}
In this section, we document the technical details about walkthroughs. 

The only game that doesn't have a walkthrough is Leather Goddesses of Phobos (LGoP).

There may be multiple correct ways to complete a game with the same level of efficiency. So the program has some randomness. In our experiments, we fixed the random seed for better reproducibility.

Each action in the program-generated walkthroughs is a highly abbreviated symbol such as E for East and NW for Northwest. 
For a better readability, we use the full words in our enhanced walkthrough such as \cref{lst:walkthrough}. 

\subsection{Location Resolution Details}\label{app:location}
For each maze, our human annotator read the %
walkthrough and annotated all the locations.
In most cases, the surrounding description given by the game engine includes the name of the location, and the annotator needs to manually extract it from the text; this process is difficult to automate because the text is often unstructured and an automatic extractor is hard to build. 
What makes this annotation tricky is
\begin{itemize}[leftmargin=*]
    \item a location may be visited more than once in the walkthrough but we should avoid assigning multiple names to it. 
    \item distinct locations may be referred to in the same way by the game engine but we should distinguish them. The game of Night is an example: hallways on different floors are all referred to as \code{Hall}; we renamed each of them, e.g., with \code{Hall (1st floor, north end)}.
\end{itemize}

We solved the problems above by hacking into the source code of the game engines and conducting multiple rounds of human verification. 
First, for each maze, we checked the source code of the game provided by the Jericho game suite~\citep{hausknecht2020interactive} and found the unique ID for each location. 
Matching IDs with human annotations allows us to perform the following post processing: 
\begin{itemize}[leftmargin=*]
    \item when a location is given multiple names, our human annotators work together to select the most proper unique name that they all agree on. 
    The selection principles are: it is descriptive; whenever this location is visited in the walkthrough, the name has an intuitive match with the surrounding description given by the game engine. 
    \item when multiple locations share a name, our human annotators work together to distinguish them by adding descriptive marks. 
    An example is the \code{Hall}s mentioned above: we renamed them to be \code{Hall (1st floor, north end of north/south hall)}, \code{Hall (1st floor, middle of north/south hall)}, and \code{Hall (2nd floor, middle of north/south hall)}. 
    There are rare cases in which all the annotators agreed that no marks could be added and the location names had to be kept fuzzy (i.e., a name corresponds to multiple different locations). 
    The rationale is: if a human may confuse with these locations, then it is reasonable for a model to have the same confusion. Then allowing them to share the name is essentially to apply a looser evaluation to the models: e.g., if the name \code{Forest} is overloaded, then when the model answers ``how to reach \code{Forest} from \code{House}'', any path that ends at any of the \code{Forest}s will be considered to be correct. 
    This treatment is equivalent to merging the locations with the same human-annotated name. %
\end{itemize}
In our repository, there is a \code{data-intermediate} folder that tracks such intermediate annotations. 
In the folder of each game, the JSON files \code{anno2code} and \code{code2anno} track the mapping between machine IDs and human-annotated location names. %

Why don't we just use the unique IDs as the location names? 
Because the IDs are often not intuitive or descriptive and they are often just strings of digits. 
Such IDs may not match any content in the walkthrough so a human or model may be confused when asked about the path between ``loc12'' and ``loc5'' after reading the very descriptive walkthrough. 

In addition, we also provide an alternative version of our data in which location names include machine IDs. %
Precisely, while resolving each location name, we %
\begin{itemize}[leftmargin=*]
    \item add the machine ID (e.g., ``(obj59)'') to the location name as a mark (e.g., ``in debris room'' to ``in debris room (obj59)''); 
    \item if the location name already has a mark (e.g., \code{1st floor, north end of north/south hall}, thanks to the name resolution process), we replace the mark with the machine ID (e.g., ``hall (1st floor, middle of north/south hall)'' to ``hall (obj66)''). 
\end{itemize}

\subsection{Move Resolution Details}\label{app:move}\label{app:qg}
During human annotation, we labeled all the moves that change the locations. 
Like in \cref{app:location}, we used the source code of the game engine to verify the human annotations. 
This annotation led to a map for each game---like what's shown in \cref{fig:map}---but this map is incomplete since there exist implicit moves between locations. 
For example, \code{south} is a movement that can end up at \code{Altar} if we start from \code{Temple}; see \cref{fig:map}. 
It means that \code{north} is also a possible move from \code{Altar} and it leads to \code{Temple}; but this edge has never explicitly shown up in the walkthrough. 
We would like our questions to cover such implicit edges, so we examined every possible implicit edge and inserted the valid ones into our map (though \cref{fig:map} only displays the explicit edges).

The examination was carried out through real game playing by our human annotators. 
For each directional move (e.g., \code{south}), we tested if its reverse directional move (e.g., \code{north} for \code{south}, \code{down} for \code{up}) could lead to the previous location. 
Not every move is directional: e.g., in Zork-I, you may \code{enter} and \code{exit} the \code{House}; \code{pray} moves the player from \code{Altar} to \code{Forest}. 
We had such edges examined by human annotators: for some (e.g., \code{enter}), we could find intuitive reverse moves and verify them; for the others (e.g., \code{pray}), we didn't propose any reverse. 
Usually, the list of moves we ended up for a game includes eight possible directional moves as well as a few special moves.

\subsection{Path Details}\label{app:path}
Once we have figured out all the locations and moves for a game, we will end up with a map. 
We collected all the unique pairs of distinct locations in the map and stored them in the \code{all\_pairs} file: for each pair of locations \mvar{A} and \mvar{B}, we could ask a route finding question that aims to reach \mvar{B} from \mvar{A}. 
Note that \mvar{A},\mvar{B} and \mvar{B},\mvar{A} are different pairs. 

For each pair of the starting point \mvar{A} and destination \mvar{B}, we collected all the simple paths {\mvar{P}} that connects from \mvar{A} to \mvar{B}. 
Each (\mvar{A},\mvar{B},\mvar{P}) tuple defines a destination finding question about where one will be if they go through path \mvar{P} from \mvar{A}.

\subsection{Program Details}\label{app:other}
Our graph and path operations (e.g., finding simple paths) are handled by the networkx package.
Its documentation is at {\small \url{https://networkx.org/}}. Particularly, the program that finds all the simple paths for a pair of graph nodes is at {\small \url{https://networkx.org/documentation/stable/reference/algorithms/generated/networkx.algorithms.simple_paths.all_simple_paths.html}}.

\subsection{Evaluation Details}\label{app:eval}

Another important---yet more strict---evaluation is the reasoning accuracy. 
This evaluation requires the language model to spell out its planned trajectories of moves when answering questions (like requested in our \cref{lst:maptemplate,lst:navtemplate}). 
For an RF question, the reasoning process is correct if and only if the model-generated trajectory is a valid path from the starting position \mvar{S} to the destination \mvar{D}: the first step starts from \mvar{S}; each step starts from where it ended up in the last step; the final step reaches \mvar{D}. 
For a DF question, the model-generated trajectory has to be a valid path from \mvar{S} to the model-generated destination \mvar{D}; in addition, the sequence of moves in the trajectory has to match the given list of actions \mvar{A}.
Like we explained in \cref{sec:eval}, the ``match'' here is not an exact match: if the closest valid move is the correct move, then it is counted as a ``match''.

\section{Experiment Details}\label{app:exp}

In this section, we present our experiment details for reproducing the results.

\subsection{Model Configuration Details}\label{app:modelconfig}

The specific GPTs used in our experiments are GPT-4-0314 and for GPT-3.5-turbo-0301. 
For Claude-1 and Claude-2, the specific model versions we are using are Claude-instant-1.2 and Claude-2.0.
For 14B RWKV model, the specific model version we are using is ``RWKV-4-Pile-14B-20230313-ctx8192-test1050''.
For Llama-2, we used the ckeckpoint officially released by Meta.

For all the models, we set the temperature of the LLMs to be 0 for reproducibility.

\subsection{Prompt Details}\label{app:promptdetail}

For GPTs, our prompts are the concatenation of walkthrough like in \cref{lst:walkthrough} and the questions like in \cref{lst:dfq,lst:rfq}; the questions are obtained by filling the templates in \cref{lst:maptemplate,lst:navtemplate}. 
When calling OpenAI API, we set the ``role'' to be ``user'' and the ``content'' to be the prompt. 
After receiving the response from the API call, we processed the output string by fetching the content from the structured output (recall that we request the models to return Python lists of Python dictionaries).

What prompts are to LLMs are like what hyperparameters are to classifical deep neural nets.
Our experiments show that the structure of LLM output is sensitive to the prompts, and thus we had to carefully tune the prompts such that LLMs could return easy-to-parse output. 
For example, we found it helpful to ask LLMs to format their answers as a Python list of Python dictionaries; we also found it helpful to end our prompt with the symbol `['---i.e., the start symbol of a Python list---to elicit the LLM to actually output in the desired format. 
Fortunately, as long as the output follows the desired structure, its content is relatively robust to the prompts. 
In our pilot experiments, we found that different prompts (created by different authors) would generate the same or similar content (regardless of its format) and achieve the same level of success rates (after parsing). 

To ensure that we could obtain the optimal results of RWKV and Llamas, we tuned the prompts---more precisely, experimented with variants of the prompts of GPTs---and ended up with a set of new prompts that are mostly the same but exhibit some prose differences. 
For example, at the beginning of the prompts, we added ``Here is a walkthrough of a text game.'' %
It is worth noting that, even though we carefully tuned the prompts, the non-GPT models still suffer a high chance of failing to return well-structured answers. 
\cref{tab:valideval} shows the average number of answered questions as well as how many of them received ill-formatted answers. 
As we could see, GPT-3.5 and GPT-4 could generate well-structured answers for a large portion of the DF and RF questions, but the other models often gave ill-structured answers. 
As a result, we could only evaluate the non-GPT models on the questions to which the answers were well-structured and thus could be parsed. 
We also experimented with the function-calling interface of GPTs, but it didn't lead to an increased amount of well-structured answers compared to our prompt design. %
\begin{table}[htbp]%
\small
\setlength{\tabcolsep}{4pt}
\begin{sc}
\begin{center}
\begin{tabular}{l|cc|cc}
\hline
\multirow{2}{*}{Model} & \multicolumn{2}{c|}{RF Question} & \multicolumn{2}{c}{DF Question} \\ \cline{2-5}
 & \# answerable & \# ill-structured & \# answerable & \# ill-structured \\ \hline
RWKV & 277.32 & 240.04 & 397.08 & 325.98 \\
Llama-2 & 138.40 & 112.25 & 157.98 & 98.0 \\
Claude-1 & 184.34 & 3.88 & 244.92 & 4.66 \\
Claude-2 & 184.34 & 6.28 & 244.92 & 1.86 \\
GPT-3.5 & 184.34 & 30.81 & 244.92 & 65.60 \\
GPT-4 & 184.34 & 2.49 & 244.92 & 5.74 \\
\hline
\end{tabular}
\end{center}
\end{sc}
\caption{Average (per-maze) numbers of answerable questions and ill-structured answers for each model.}
\label{tab:valideval}
\end{table}

\subsection{Walkthrough and Question Details}\label{app:walkthroughdetails}\label{app:qdetails}
Due to the context window size of an LLM, it is often the case that we have to only use a prefix of the walkthrough when evaluating an LLM on a maze. 
Recall from \cref{sec:qg} that not every question is answerable given a walkthrough prefix. 
Therefore, each model was evaluated on a different set of questions. 
\cref{tab:expdatastatgpt,tab:expdatastatllama-2,tab:expdatastatllama-1,tab:expdatastatother} display the statistics about the data that each LLM was actually evaluated on in our experiments. 
\begin{table}[htbp]
\small
\setlength{\tabcolsep}{4pt}
\begin{sc}
\begin{subtable}{1.0\linewidth}
\begin{center}
\begin{tabular}{lrrrrrrrr}
\toprule
\multirow{2}{*}{Maps} & \multirow{2}{*}{\# Locs} & \multirow{2}{*}{\# Edges} & \multirow{2}{*}{Avg Len Path} & \multirow{2}{*}{\# Steps} & \multicolumn{2}{c}{DF} & \multicolumn{2}{c}{RF} \\
\cmidrule(lr){6-7} \cmidrule(lr){8-9}
& & & & &Easy & Hard & Easy & Hard \\
\midrule
905 & 5 & 7 & 1.88 & 21 & 11 & 5 & 11 & 5 \\
advent & 31 & 57 & 7.79 & 70 & 692 & 100 & 532 & 100 \\
adventureland & 18 & 35 & 6.13 & 70 & 579 & 80 & 260 & 46 \\
afflicted & 10 & 16 & 2.78 & 40 & 67 & 0 & 67 & 0 \\
anchor & 13 & 24 & 3.94 & 24 & 95 & 91 & 82 & 74 \\
awaken & 14 & 24 & 4.8 & 44 & 262 & 12 & 157 & 12 \\
balances & 11 & 18 & 3.09 & 67 & 96 & 8 & 76 & 8 \\
ballyhoo & 14 & 28 & 4.8 & 56 & 225 & 99 & 156 & 13 \\
curses & 14 & 24 & 3.3 & 53 & 122 & 13 & 122 & 0 \\
cutthroat & 22 & 40 & 5.81 & 62 & 303 & 158 & 202 & 107 \\
deephome & 17 & 28 & 4.35 & 49 & 175 & 10 & 175 & 10 \\
detective & 26 & 34 & 7.17 & 43 & 334 & 4 & 334 & 4 \\
dragon & 14 & 25 & 3.59 & 29 & 105 & 64 & 111 & 58 \\
enchanter & 21 & 38 & 5.69 & 53 & 216 & 165 & 216 & 165 \\
enter & 2 & 1 & 1.0 & 20 & 1 & 0 & 1 & 0 \\
gold & 11 & 17 & 2.82 & 47 & 83 & 0 & 83 & 0 \\
hhgg & 8 & 9 & 2.6 & 51 & 29 & 1 & 29 & 1 \\
hollywood & 8 & 14 & 2.71 & 50 & 43 & 13 & 43 & 13 \\
huntdark & 10 & 9 & 3.67 & 55 & 45 & 0 & 45 & 0 \\
infidel & 13 & 26 & 3.6 & 55 & 114 & 138 & 88 & 68 \\
inhumane & 21 & 40 & 4.8 & 49 & 275 & 240 & 261 & 159 \\
jewel & 16 & 30 & 4.39 & 60 & 166 & 74 & 166 & 74 \\
karn & 19 & 35 & 6.37 & 65 & 339 & 86 & 231 & 63 \\
library & 7 & 12 & 2.48 & 49 & 42 & 0 & 42 & 0 \\
loose & 8 & 14 & 3.0 & 39 & 56 & 0 & 56 & 0 \\
lostpig & 6 & 9 & 1.96 & 56 & 16 & 9 & 16 & 9 \\
ludicorp & 22 & 43 & 4.91 & 70 & 351 & 111 & 351 & 111 \\
lurking & 10 & 16 & 2.89 & 56 & 66 & 16 & 65 & 16 \\
moonlit & 4 & 6 & 1.67 & 45 & 9 & 3 & 9 & 3 \\
murdac & 30 & 52 & 6.34 & 70 & 537 & 195 & 528 & 183 \\
night & 20 & 41 & 6.93 & 70 & 633 & 59 & 380 & 0 \\
omniquest & 29 & 59 & 7.75 & 70 & 536 & 1198 & 290 & 298 \\
partyfoul & 4 & 6 & 1.67 & 24 & 12 & 0 & 11 & 1 \\
pentari & 18 & 30 & 3.72 & 48 & 208 & 4 & 208 & 4 \\
planetfall & 21 & 37 & 5.27 & 68 & 246 & 50 & 246 & 50 \\
plundered & 10 & 11 & 3.23 & 32 & 47 & 0 & 47 & 0 \\
reverb & 12 & 21 & 4.07 & 40 & 129 & 12 & 120 & 12 \\
seastalker & 10 & 15 & 2.7 & 53 & 50 & 3 & 50 & 3 \\
sherlock & 8 & 13 & 2.48 & 33 & 43 & 5 & 43 & 0 \\
snacktime & 4 & 6 & 1.5 & 33 & 12 & 0 & 12 & 0 \\
sorcerer & 15 & 26 & 3.53 & 54 & 120 & 17 & 120 & 17 \\
spellbrkr & 15 & 23 & 3.6 & 52 & 120 & 17 & 120 & 16 \\
spirit & 15 & 26 & 3.55 & 50 & 171 & 12 & 171 & 12 \\
temple & 10 & 15 & 3.11 & 46 & 48 & 18 & 48 & 18 \\
trinity & 10 & 10 & 3.61 & 45 & 45 & 1 & 45 & 1 \\
tryst205 & 9 & 14 & 1.93 & 65 & 57 & 0 & 57 & 0 \\
wishbringer & 18 & 30 & 5.2 & 45 & 174 & 48 & 175 & 47 \\
yomomma & 8 & 15 & 2.39 & 41 & 49 & 17 & 31 & 18 \\
zenon & 13 & 24 & 4.05 & 63 & 83 & 73 & 83 & 73 \\
zork1 & 19 & 34 & 7.16 & 70 & 351 & 46 & 279 & 45 \\
zork2 & 22 & 33 & 5.65 & 50 & 242 & 18 & 146 & 103 \\
zork3 & 23 & 44 & 6.84 & 61 & 600 & 167 & 394 & 68 \\
ztuu & 11 & 18 & 2.79 & 43 & 91 & 0 & 91 & 0 \\
\bottomrule
\end{tabular}
\end{center}
\end{subtable}%
\end{sc}
\caption{Map statistics for GPTs and Claudes. Here, \textsc{\# Locs} represents the number of locations in each walkthrough; \textsc{\# Edges} represents the number of edges; \textsc{AVG LEN PATH} denotes the average length of all paths; \textsc{\# STEPS} indicates the number of steps in each walkthrough; \textsc{EASY} and \textsc{HARD} of the \textsc{DF} and \textsc{RF} respectively represent the number of easy and hard skeletons of the DF and RF tasks.}
\label{tab:expdatastatgpt}
\end{table}
\begin{table}[htbp]
\small
\setlength{\tabcolsep}{4pt}
\begin{sc}
\begin{subtable}{1.0\linewidth}
\begin{center}
\begin{tabular}{lrrrrrrrr}
\toprule
\multirow{2}{*}{Maps} & \multirow{2}{*}{\# Locs} & \multirow{2}{*}{\# Edges} & \multirow{2}{*}{Avg Len Path} & \multirow{2}{*}{\# Steps} & \multicolumn{2}{c}{DF} & \multicolumn{2}{c}{RF} \\
\cmidrule(lr){6-7} \cmidrule(lr){8-9}
& & & & &Easy & Hard & Easy & Hard \\
\midrule
905 & 5 & 7 & 1.88 & 21 & 11 & 5 & 11 & 5 \\
advent & 31 & 57 & 7.79 & 70 & 692 & 100 & 532 & 100 \\
adventureland & 18 & 35 & 6.13 & 70 & 579 & 80 & 260 & 46 \\
afflicted & 11 & 20 & 2.95 & 70 & 100 & 10 & 100 & 10 \\
anchor & 25 & 46 & 5.99 & 70 & 327 & 153 & 302 & 132 \\
awaken & 15 & 28 & 5.02 & 56 & 365 & 45 & 171 & 25 \\
balances & 11 & 18 & 3.09 & 70 & 96 & 8 & 76 & 8 \\
ballyhoo & 17 & 35 & 4.83 & 70 & 302 & 188 & 213 & 59 \\
curses & 14 & 27 & 3.62 & 70 & 182 & 13 & 182 & 0 \\
cutthroat & 25 & 49 & 6.66 & 70 & 471 & 362 & 360 & 216 \\
deephome & 27 & 49 & 4.83 & 70 & 429 & 19 & 429 & 19 \\
detective & 32 & 40 & 8.79 & 50 & 505 & 4 & 505 & 4 \\
dragon & 21 & 44 & 7.11 & 70 & 533 & 990 & 272 & 148 \\
enchanter & 23 & 43 & 6.35 & 70 & 265 & 219 & 265 & 219 \\
enter & 14 & 26 & 3.36 & 70 & 117 & 65 & 117 & 65 \\
gold & 15 & 25 & 3.45 & 70 & 143 & 0 & 143 & 0 \\
hhgg & 9 & 11 & 2.85 & 70 & 38 & 1 & 38 & 1 \\
hollywood & 12 & 22 & 3.36 & 70 & 84 & 48 & 84 & 48 \\
huntdark & 12 & 11 & 4.33 & 66 & 66 & 0 & 66 & 0 \\
infidel & 24 & 48 & 7.53 & 70 & 312 & 446 & 264 & 288 \\
inhumane & 30 & 57 & 5.54 & 70 & 614 & 555 & 483 & 280 \\
jewel & 17 & 32 & 4.4 & 70 & 187 & 85 & 187 & 85 \\
karn & 19 & 35 & 6.37 & 70 & 339 & 86 & 231 & 63 \\
library & 7 & 12 & 2.48 & 51 & 42 & 0 & 42 & 0 \\
loose & 12 & 21 & 4.18 & 49 & 94 & 27 & 94 & 27 \\
lostpig & 7 & 11 & 2.28 & 70 & 22 & 14 & 22 & 14 \\
ludicorp & 22 & 43 & 4.91 & 70 & 351 & 111 & 351 & 111 \\
lurking & 16 & 29 & 4.29 & 70 & 144 & 97 & 143 & 97 \\
moonlit & 6 & 9 & 2.2 & 58 & 18 & 7 & 18 & 7 \\
murdac & 30 & 52 & 6.34 & 70 & 537 & 195 & 528 & 183 \\
night & 20 & 41 & 6.93 & 70 & 633 & 59 & 380 & 0 \\
omniquest & 29 & 59 & 7.75 & 70 & 536 & 1198 & 290 & 298 \\
partyfoul & 4 & 9 & 1.97 & 55 & 24 & 6 & 11 & 1 \\
pentari & 18 & 30 & 3.72 & 48 & 208 & 4 & 208 & 4 \\
planetfall & 22 & 39 & 5.48 & 70 & 267 & 63 & 267 & 63 \\
plundered & 22 & 37 & 6.02 & 70 & 450 & 52 & 289 & 26 \\
reverb & 17 & 31 & 5.26 & 70 & 321 & 20 & 253 & 19 \\
seastalker & 10 & 15 & 2.7 & 70 & 50 & 3 & 50 & 3 \\
sherlock & 18 & 28 & 4.36 & 70 & 175 & 5 & 175 & 0 \\
snacktime & 4 & 6 & 1.5 & 33 & 12 & 0 & 12 & 0 \\
sorcerer & 26 & 46 & 7.09 & 70 & 340 & 48 & 340 & 48 \\
spellbrkr & 20 & 31 & 4.84 & 70 & 295 & 23 & 276 & 21 \\
spirit & 22 & 41 & 4.09 & 70 & 354 & 87 & 354 & 87 \\
temple & 19 & 33 & 4.72 & 70 & 178 & 69 & 178 & 69 \\
trinity & 17 & 17 & 5.96 & 70 & 136 & 1 & 136 & 1 \\
tryst205 & 9 & 15 & 1.94 & 70 & 64 & 0 & 64 & 0 \\
wishbringer & 21 & 40 & 6.34 & 70 & 259 & 214 & 251 & 169 \\
yomomma & 9 & 20 & 2.74 & 70 & 82 & 59 & 43 & 21 \\
zenon & 14 & 26 & 4.27 & 70 & 96 & 86 & 96 & 86 \\
zork1 & 19 & 34 & 7.16 & 70 & 351 & 46 & 279 & 45 \\
zork2 & 22 & 45 & 7.01 & 70 & 536 & 754 & 239 & 130 \\
zork3 & 23 & 45 & 6.93 & 70 & 627 & 174 & 414 & 70 \\
ztuu & 15 & 26 & 3.15 & 70 & 183 & 0 & 183 & 0 \\
\bottomrule
\end{tabular}
\end{center}
\end{subtable}%
\end{sc}
\caption{Map statistics for RWKV, RWKV-S, Llama-1-S, Llama-2-S, Code-Llama-S, Code-Llama-Instruct-S. Here, \textsc{\# Locs} represents the number of locations in each walkthrough; \textsc{\# Edges} represents the number of edges; \textsc{AVG LEN PATH} denotes the average length of all paths; \textsc{\# STEPS} indicates the number of steps in each walkthrough; \textsc{EASY} and \textsc{HARD} of the \textsc{DF} and \textsc{RF} respectively represent the number of easy and hard skeletons of the DF and RF tasks.}
\label{tab:expdatastatother}
\end{table}
\begin{table}[htbp]
\small
\setlength{\tabcolsep}{4pt}
\begin{sc}
\begin{subtable}{1.0\linewidth}
\begin{center}
\begin{tabular}{lrrrrrrrr}
\toprule
\multirow{2}{*}{Maps} & \multirow{2}{*}{\# Locs} & \multirow{2}{*}{\# Edges} & \multirow{2}{*}{Avg Len Path} & \multirow{2}{*}{\# Steps} & \multicolumn{2}{c}{DF} & \multicolumn{2}{c}{RF} \\
\cmidrule(lr){6-7} \cmidrule(lr){8-9}
& & & & &Easy & Hard & Easy & Hard \\
\midrule
905 & 5 & 7 & 1.88 & 19 & 11 & 5 & 11 & 5 \\
advent & 12 & 21 & 3.73 & 18 & 67 & 30 & 67 & 30 \\
adventureland & 8 & 11 & 2.09 & 20 & 32 & 1 & 32 & 1 \\
afflicted & 3 & 3 & 1.25 & 8 & 4 & 0 & 4 & 0 \\
anchor & 5 & 5 & 1.91 & 5 & 10 & 1 & 7 & 4 \\
awaken & 6 & 8 & 2.32 & 10 & 24 & 4 & 21 & 4 \\
balances & 5 & 5 & 1.91 & 16 & 10 & 1 & 10 & 1 \\
ballyhoo & 5 & 4 & 2.0 & 9 & 10 & 0 & 10 & 0 \\
curses & 6 & 8 & 1.95 & 12 & 17 & 2 & 17 & 0 \\
cutthroat & 2 & 2 & 1.0 & 13 & 1 & 1 & 1 & 1 \\
deephome & 3 & 3 & 1.25 & 11 & 3 & 1 & 3 & 1 \\
detective & 7 & 8 & 2.35 & 16 & 23 & 0 & 23 & 0 \\
dragon & 4 & 4 & 1.57 & 4 & 6 & 1 & 6 & 1 \\
enchanter & 9 & 13 & 2.63 & 15 & 38 & 13 & 38 & 13 \\
enter & 0 & 0 & 0 & 3 & 0 & 0 & 0 & 0 \\
gold & 5 & 5 & 1.91 & 9 & 11 & 0 & 11 & 0 \\
hhgg & 4 & 4 & 1.57 & 15 & 6 & 1 & 6 & 1 \\
hollywood & 4 & 5 & 1.56 & 10 & 6 & 3 & 6 & 3 \\
huntdark & 3 & 2 & 1.33 & 10 & 3 & 0 & 3 & 0 \\
infidel & 6 & 8 & 1.9 & 9 & 16 & 5 & 17 & 4 \\
inhumane & 6 & 8 & 1.95 & 14 & 19 & 2 & 19 & 2 \\
jewel & 6 & 9 & 2.08 & 17 & 21 & 4 & 21 & 4 \\
karn & 3 & 4 & 1.33 & 20 & 6 & 0 & 6 & 0 \\
library & 2 & 1 & 1.0 & 12 & 1 & 0 & 1 & 0 \\
loose & 4 & 3 & 1.67 & 8 & 6 & 0 & 6 & 0 \\
lostpig & 3 & 3 & 1.25 & 17 & 3 & 1 & 3 & 1 \\
ludicorp & 8 & 11 & 2.62 & 24 & 32 & 0 & 32 & 0 \\
lurking & 3 & 3 & 1.25 & 21 & 4 & 0 & 4 & 0 \\
moonlit & 3 & 2 & 1.33 & 14 & 3 & 0 & 3 & 0 \\
murdac & 11 & 19 & 3.75 & 25 & 110 & 8 & 102 & 8 \\
night & 11 & 12 & 3.62 & 17 & 58 & 0 & 58 & 0 \\
omniquest & 10 & 18 & 2.73 & 22 & 54 & 36 & 54 & 36 \\
partyfoul & 0 & 0 & 0 & 4 & 0 & 0 & 0 & 0 \\
pentari & 8 & 10 & 2.78 & 7 & 28 & 4 & 28 & 4 \\
planetfall & 2 & 2 & 1.0 & 20 & 1 & 1 & 1 & 1 \\
plundered & 2 & 1 & 1.0 & 9 & 1 & 0 & 1 & 0 \\
reverb & 5 & 5 & 1.91 & 8 & 10 & 1 & 10 & 1 \\
seastalker & 2 & 2 & 1.0 & 12 & 1 & 1 & 1 & 1 \\
sherlock & 4 & 3 & 1.67 & 7 & 6 & 0 & 6 & 0 \\
snacktime & 2 & 2 & 1.0 & 15 & 2 & 0 & 2 & 0 \\
sorcerer & 4 & 7 & 1.56 & 16 & 7 & 2 & 7 & 2 \\
spellbrkr & 3 & 3 & 1.25 & 10 & 3 & 1 & 3 & 1 \\
spirit & 4 & 4 & 1.57 & 9 & 6 & 1 & 6 & 1 \\
temple & 3 & 3 & 1.25 & 11 & 3 & 1 & 3 & 1 \\
trinity & 4 & 4 & 1.57 & 14 & 6 & 1 & 6 & 1 \\
tryst205 & 3 & 2 & 1.33 & 9 & 3 & 0 & 3 & 0 \\
wishbringer & 7 & 8 & 2.35 & 11 & 23 & 0 & 23 & 0 \\
yomomma & 4 & 5 & 1.57 & 8 & 6 & 1 & 4 & 3 \\
zenon & 4 & 4 & 1.57 & 21 & 6 & 1 & 6 & 1 \\
zork1 & 8 & 12 & 2.27 & 22 & 34 & 3 & 34 & 3 \\
zork2 & 8 & 10 & 2.65 & 11 & 29 & 2 & 29 & 2 \\
zork3 & 7 & 14 & 2.59 & 17 & 40 & 24 & 30 & 12 \\
ztuu & 5 & 4 & 2.0 & 5 & 10 & 0 & 10 & 0 \\
\bottomrule
\end{tabular}
\end{center}
\end{subtable}%
\end{sc}
\caption{Map statistics for Llama-1.
Here, \textsc{\# Locs} represents the number of locations in each walkthrough; \textsc{\# Edges} represents the number of edges; \textsc{AVG LEN PATH} denotes the average length of all paths; \textsc{\# STEPS} indicates the number of steps in each walkthrough; \textsc{EASY} and \textsc{HARD} of the \textsc{DF} and \textsc{RF} respectively represent the number of easy and hard skeletons of the DF and RF tasks.}
\label{tab:expdatastatllama-1}
\end{table}
\begin{table}[htbp]
\small
\setlength{\tabcolsep}{4pt}
\begin{sc}
\begin{subtable}{1.0\linewidth}
\begin{center}
\begin{tabular}{lrrrrrrrr}
\toprule
\multirow{2}{*}{Maps} & \multirow{2}{*}{\# Locs} & \multirow{2}{*}{\# Edges} & \multirow{2}{*}{Avg Len Path} & \multirow{2}{*}{\# Steps} & \multicolumn{2}{c}{DF} & \multicolumn{2}{c}{RF} \\
\cmidrule(lr){6-7} \cmidrule(lr){8-9}
& & & & &Easy & Hard & Easy & Hard \\
\midrule
905 & 5 & 7 & 1.88 & 21 & 11 & 5 & 11 & 5 \\
advent & 25 & 45 & 7.52 & 54 & 473 & 76 & 355 & 76 \\
adventureland & 18 & 33 & 5.91 & 56 & 413 & 42 & 260 & 46 \\
afflicted & 9 & 14 & 2.43 & 32 & 46 & 0 & 46 & 0 \\
anchor & 9 & 15 & 3.0 & 19 & 50 & 16 & 41 & 13 \\
awaken & 13 & 18 & 3.98 & 34 & 114 & 4 & 90 & 4 \\
balances & 8 & 12 & 2.51 & 49 & 38 & 1 & 38 & 1 \\
ballyhoo & 12 & 21 & 3.98 & 43 & 111 & 35 & 101 & 11 \\
curses & 10 & 17 & 2.93 & 42 & 67 & 9 & 67 & 0 \\
cutthroat & 19 & 28 & 4.75 & 42 & 180 & 9 & 136 & 53 \\
deephome & 12 & 21 & 3.43 & 40 & 102 & 10 & 102 & 10 \\
detective & 25 & 32 & 6.99 & 40 & 307 & 4 & 307 & 4 \\
dragon & 11 & 20 & 3.16 & 22 & 68 & 42 & 70 & 40 \\
enchanter & 17 & 31 & 4.61 & 41 & 142 & 114 & 142 & 114 \\
enter & 0 & 0 & 0 & 15 & 0 & 0 & 0 & 0 \\
gold & 8 & 14 & 2.5 & 40 & 56 & 0 & 56 & 0 \\
hhgg & 7 & 8 & 2.35 & 39 & 22 & 1 & 22 & 1 \\
hollywood & 7 & 8 & 2.5 & 34 & 21 & 3 & 21 & 3 \\
huntdark & 7 & 6 & 2.67 & 41 & 21 & 0 & 21 & 0 \\
infidel & 12 & 24 & 3.34 & 42 & 102 & 116 & 78 & 54 \\
inhumane & 18 & 34 & 4.82 & 40 & 218 & 168 & 190 & 116 \\
jewel & 15 & 25 & 3.77 & 46 & 141 & 30 & 141 & 30 \\
karn & 15 & 27 & 4.11 & 49 & 136 & 38 & 144 & 30 \\
library & 7 & 11 & 2.22 & 40 & 32 & 0 & 32 & 0 \\
loose & 8 & 13 & 2.86 & 28 & 49 & 0 & 49 & 0 \\
lostpig & 5 & 7 & 1.69 & 47 & 11 & 5 & 11 & 5 \\
ludicorp & 19 & 37 & 4.14 & 65 & 291 & 51 & 291 & 51 \\
lurking & 6 & 9 & 2.42 & 40 & 30 & 1 & 29 & 1 \\
moonlit & 3 & 2 & 1.33 & 36 & 3 & 0 & 3 & 0 \\
murdac & 23 & 43 & 5.9 & 62 & 355 & 171 & 346 & 160 \\
night & 20 & 39 & 5.79 & 53 & 380 & 19 & 380 & 0 \\
omniquest & 26 & 49 & 6.28 & 58 & 348 & 180 & 222 & 192 \\
partyfoul & 4 & 6 & 1.67 & 19 & 12 & 0 & 11 & 1 \\
pentari & 17 & 29 & 3.67 & 42 & 191 & 4 & 191 & 4 \\
planetfall & 17 & 29 & 4.96 & 52 & 164 & 16 & 164 & 16 \\
plundered & 7 & 7 & 2.59 & 26 & 22 & 0 & 22 & 0 \\
reverb & 12 & 15 & 3.54 & 28 & 68 & 2 & 68 & 2 \\
seastalker & 10 & 15 & 2.7 & 43 & 50 & 3 & 50 & 3 \\
sherlock & 6 & 9 & 2.08 & 21 & 25 & 0 & 25 & 0 \\
snacktime & 4 & 6 & 1.5 & 33 & 12 & 0 & 12 & 0 \\
sorcerer & 10 & 18 & 2.3 & 44 & 60 & 13 & 60 & 13 \\
spellbrkr & 10 & 13 & 2.71 & 41 & 60 & 2 & 59 & 2 \\
spirit & 13 & 21 & 3.22 & 38 & 99 & 6 & 99 & 6 \\
temple & 9 & 14 & 2.84 & 37 & 39 & 18 & 39 & 18 \\
trinity & 8 & 8 & 2.93 & 37 & 28 & 1 & 28 & 1 \\
tryst205 & 8 & 13 & 1.92 & 49 & 49 & 0 & 49 & 0 \\
wishbringer & 18 & 30 & 5.2 & 38 & 174 & 48 & 175 & 47 \\
yomomma & 7 & 13 & 2.18 & 31 & 36 & 14 & 26 & 10 \\
zenon & 12 & 22 & 3.71 & 52 & 70 & 62 & 70 & 62 \\
zork1 & 19 & 32 & 7.25 & 56 & 332 & 22 & 279 & 45 \\
zork2 & 19 & 27 & 5.15 & 42 & 176 & 18 & 107 & 76 \\
zork3 & 18 & 35 & 4.78 & 48 & 282 & 100 & 199 & 45 \\
ztuu & 8 & 11 & 2.26 & 28 & 38 & 0 & 38 & 0 \\
\bottomrule
\end{tabular}
\end{center}
\end{subtable}%
\end{sc}
\caption{Map statistics for Llama-2, Code-Llama, Code-Llama-Instruct.
Here, \textsc{\# Locs} represents the number of locations in each walkthrough; \textsc{\# Edges} represents the number of edges; \textsc{AVG LEN PATH} denotes the average length of all paths; \textsc{\# STEPS} indicates the number of steps in each walkthrough; \textsc{EASY} and \textsc{HARD} of the \textsc{DF} and \textsc{RF} respectively represent the number of easy and hard skeletons of the DF and RF tasks.}
\label{tab:expdatastatllama-2}
\end{table}

\clearpage

\section{More Results}\label{app:results}

\subsection{Success Rate Weighted by Route Length}\label{app:sr_weighted}
\begin{figure*}[!htbp]
	\begin{center}
	    \begin{subfigure}[t]{0.48\linewidth}
                \includegraphics[width=0.49\linewidth]{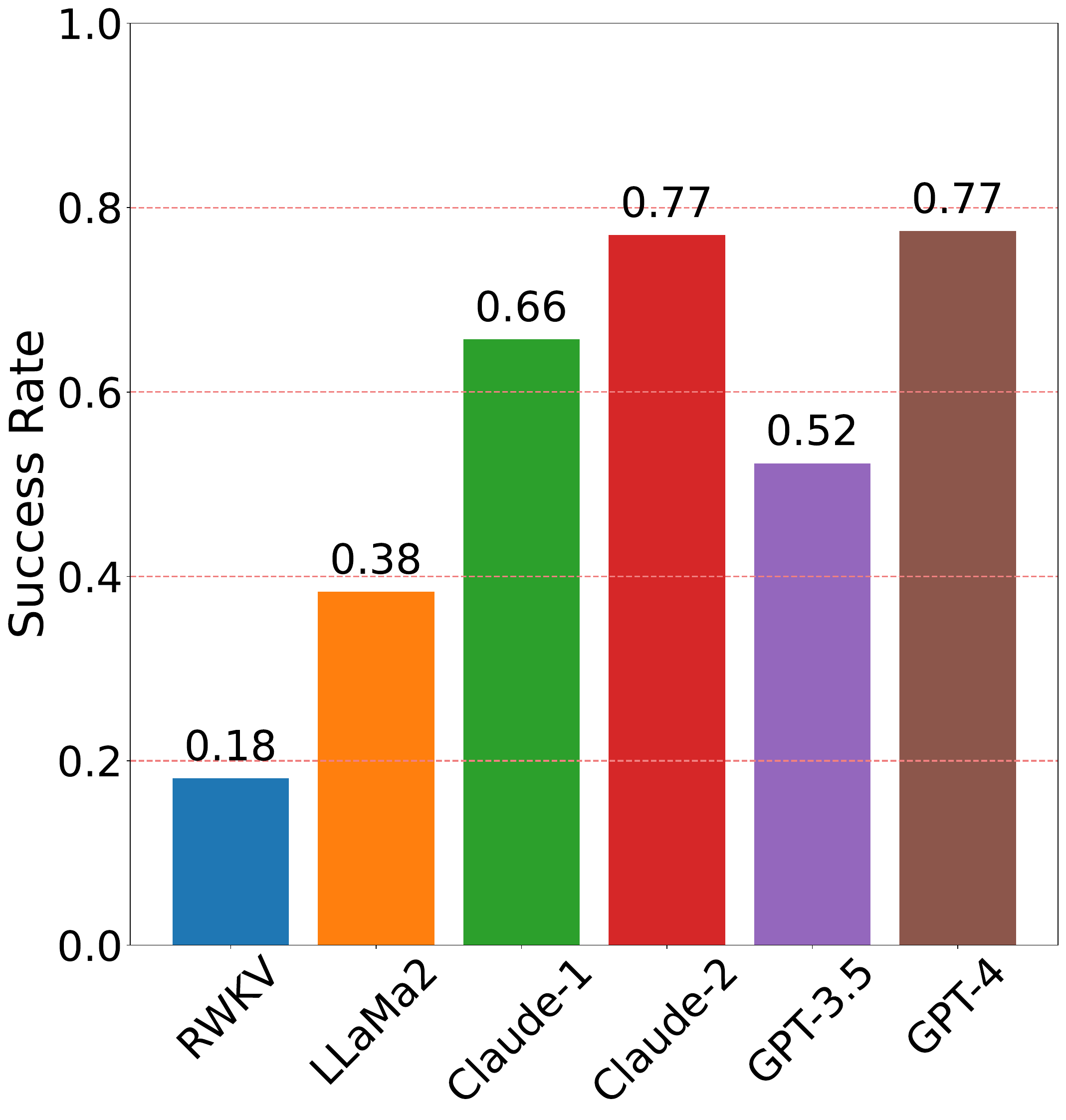}
                \hfill
                \includegraphics[width=0.49\linewidth]{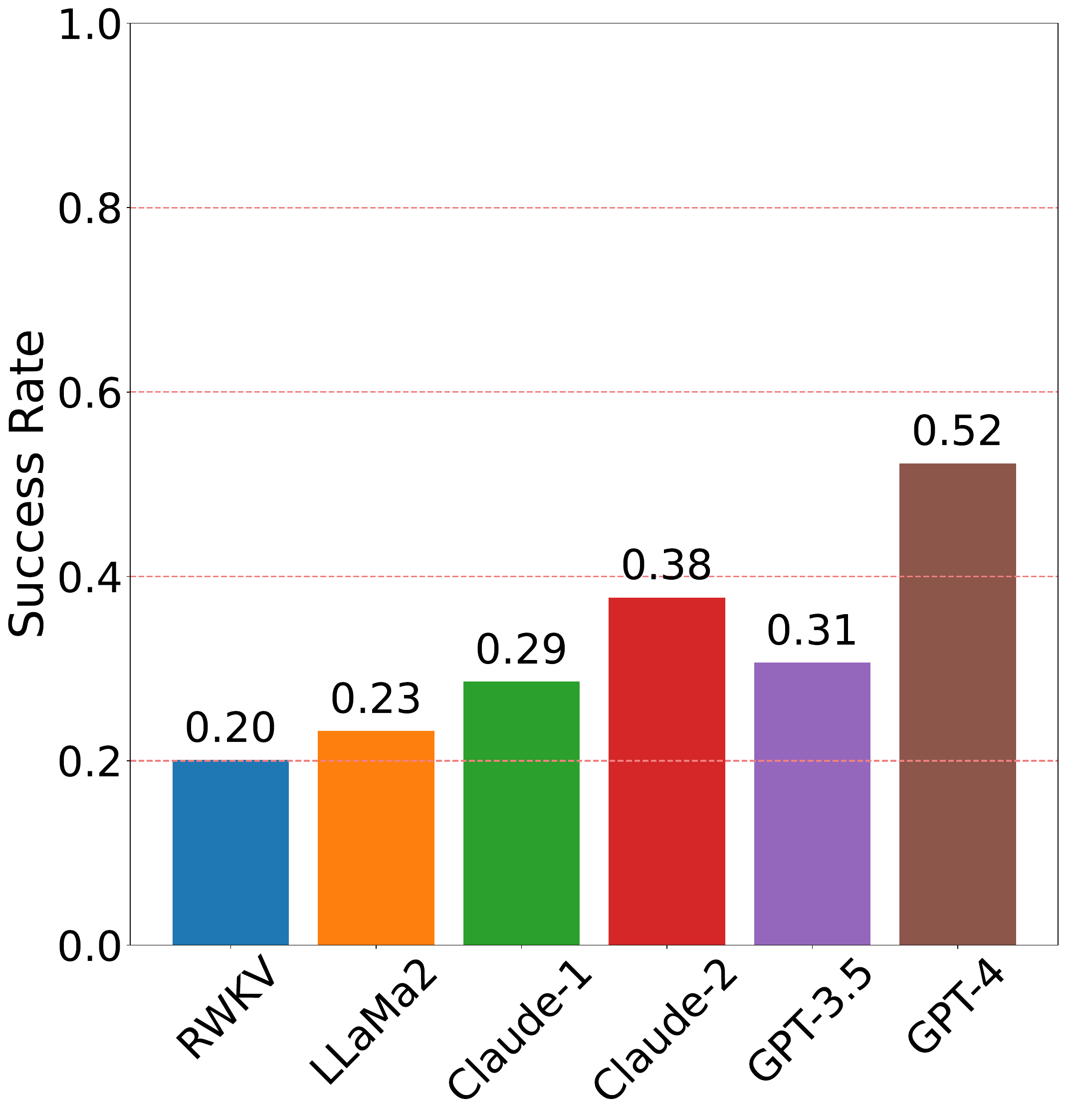}
		    \vspace{-16pt}
			\caption{On easy (left) and hard (right) DF questions.}\label{fig:length_df}
		\end{subfigure}
		\hfill
            \begin{subfigure}[t]{0.48\linewidth}
			\includegraphics[width=0.49\linewidth]{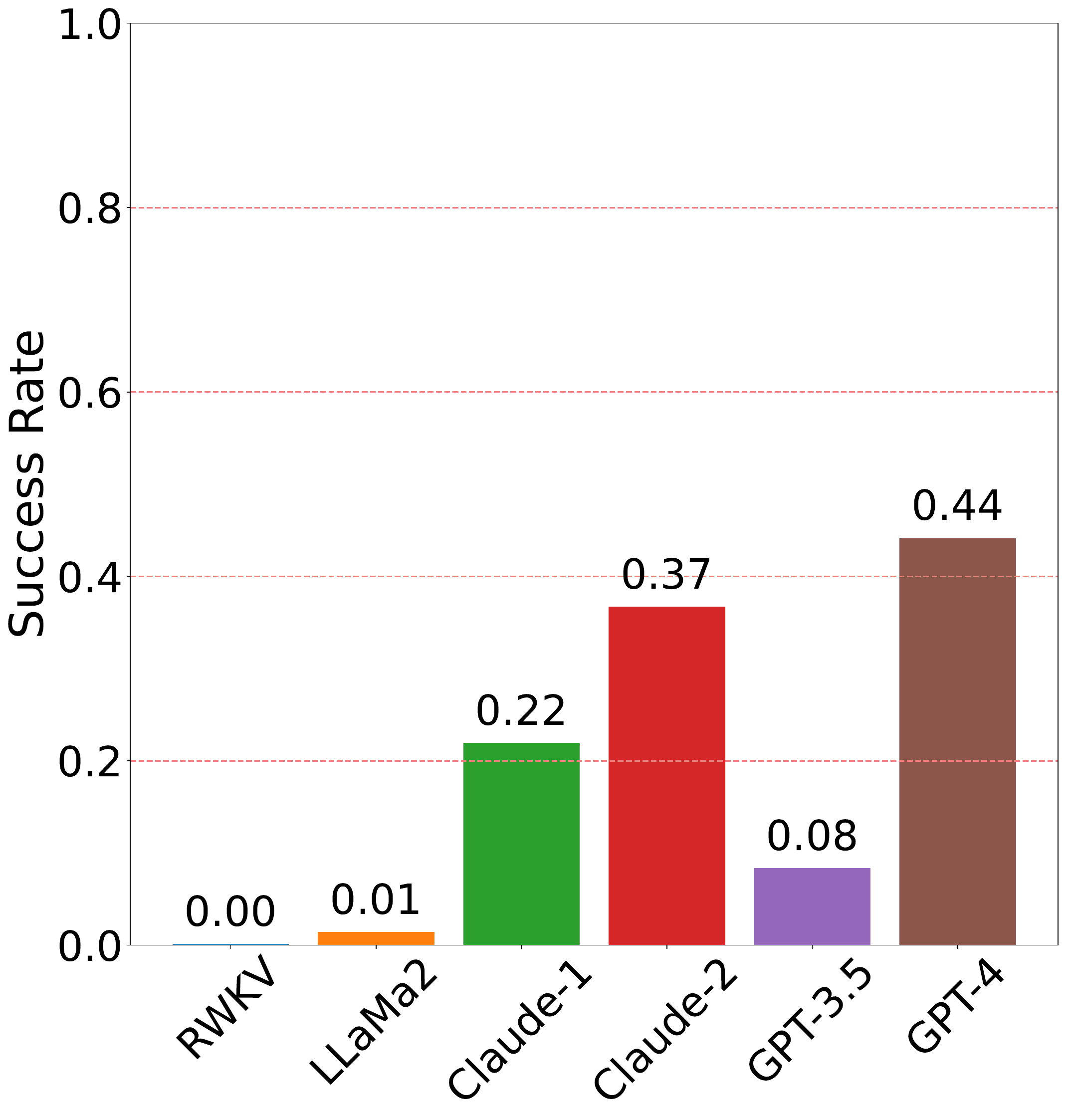}
                \hfill
                \includegraphics[width=0.49\linewidth]{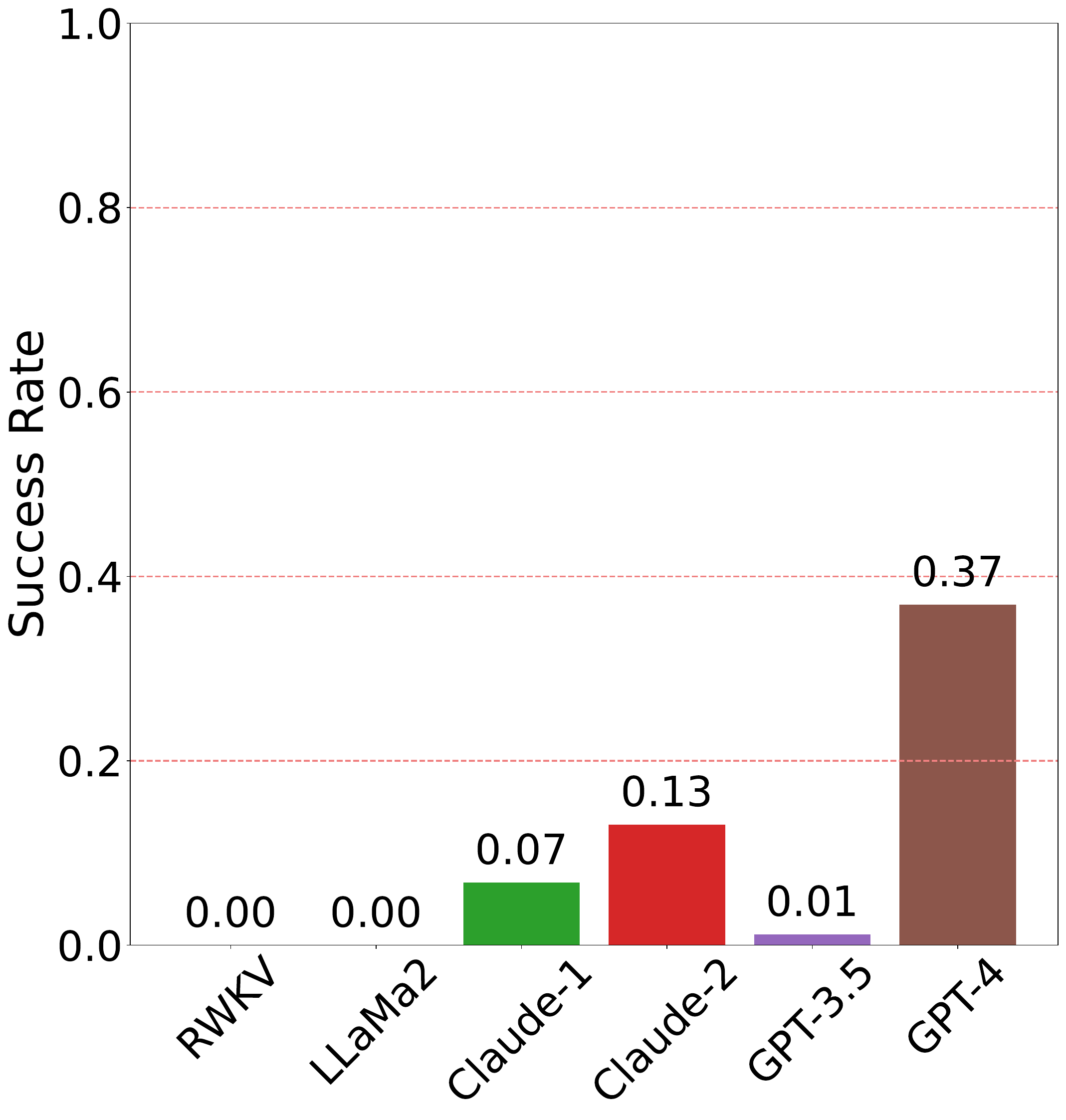}
			\vspace{-16pt}
			\caption{On easy (left) and hard (right) RF questions.}\label{fig:length_rf}
		\end{subfigure}
		\vspace{-8pt}
		\caption{Success rates weighted by route length on DF (\subref{fig:length_df}) and RF (\subref{fig:length_rf}) questions, averaged over all $53$ mazes.}\label{fig:length}
	\end{center}
	\vspace{-4pt}
    \label{fig:weightedsr}
\end{figure*}

Following the evaluation standard proposed by \citet{anderson2018evaluation}, we also computed the success rates weighted by route length, which are shown in \cref{fig:weightedsr}. 
As we can see, the ranking of the models stays the same as in \cref{fig:main}. 

\subsection{Reasoning Accuracy Results}\label{app:reasoningacc}
\begin{figure*}[!htbp]
	\begin{center}
	    \begin{subfigure}[t]{0.48\linewidth}
                \includegraphics[width=0.49\linewidth]{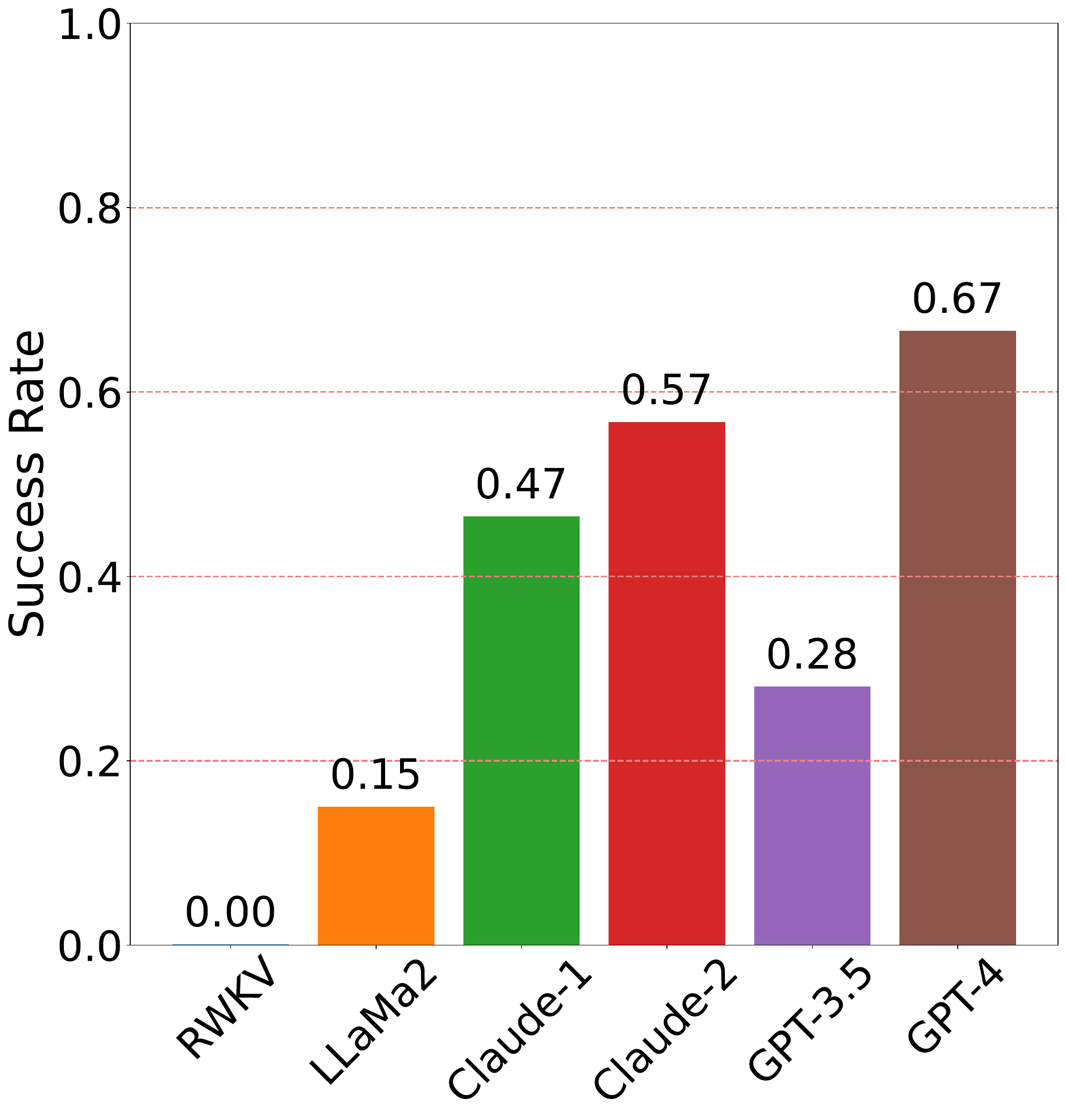}
                \hfill
                \includegraphics[width=0.49\linewidth]{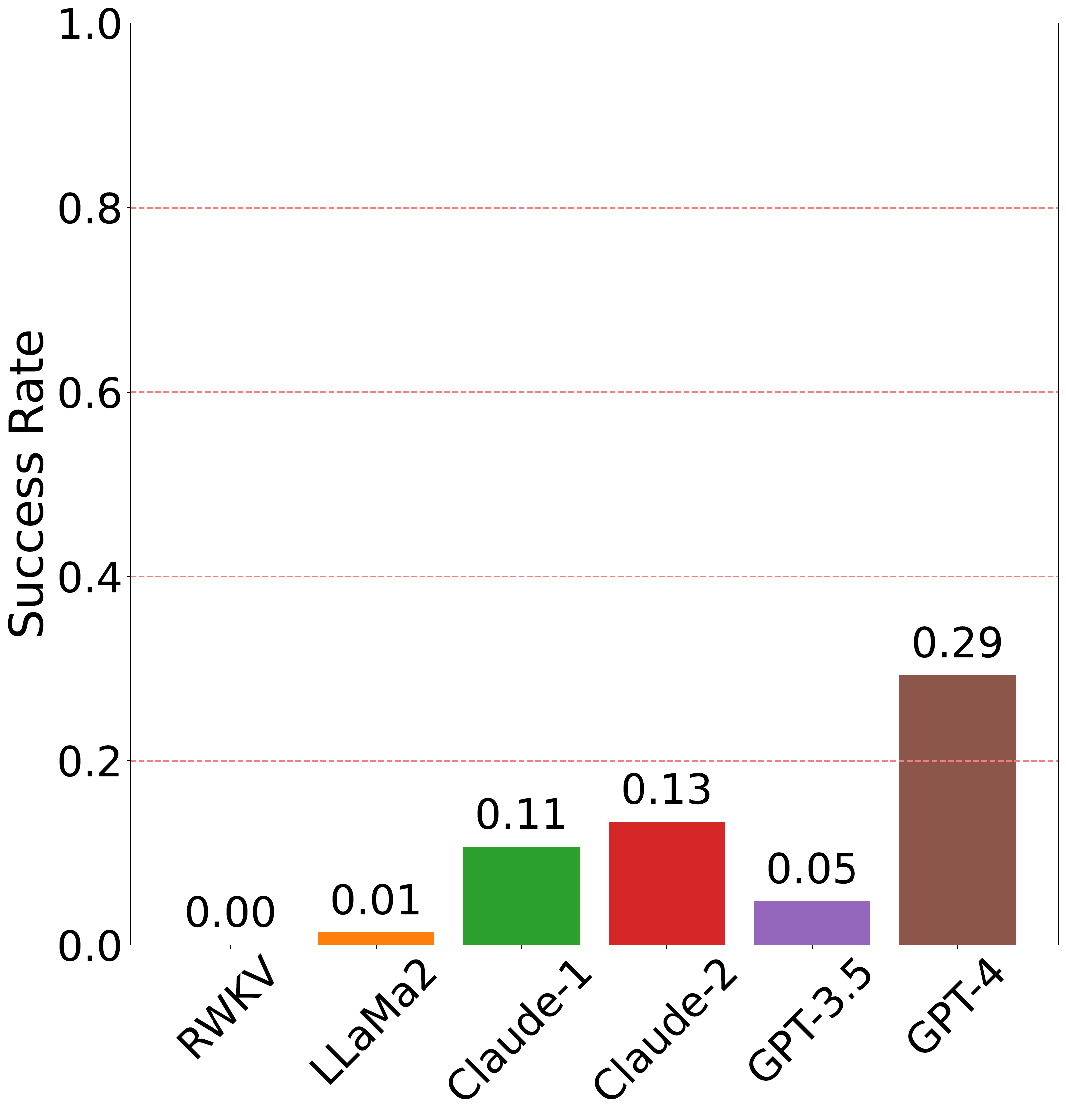}
			\caption{On easy (left) and hard (right) DF questions.}\label{fig:main_df_rea_acc}
		\end{subfigure}
		\hfill
            \begin{subfigure}[t]{0.48\linewidth}
			\includegraphics[width=0.49\linewidth]{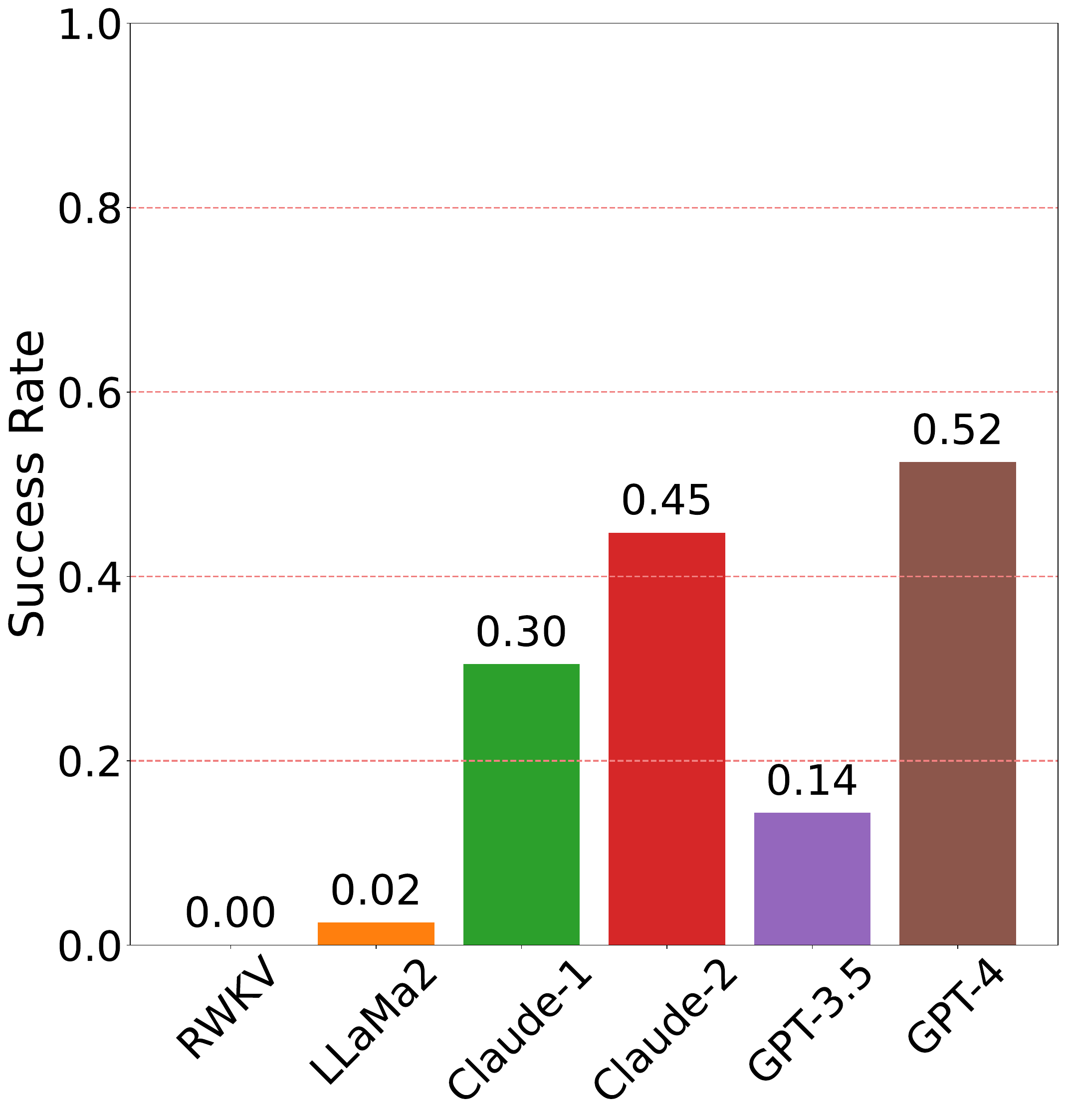}
                \hfill
                \includegraphics[width=0.49\linewidth]{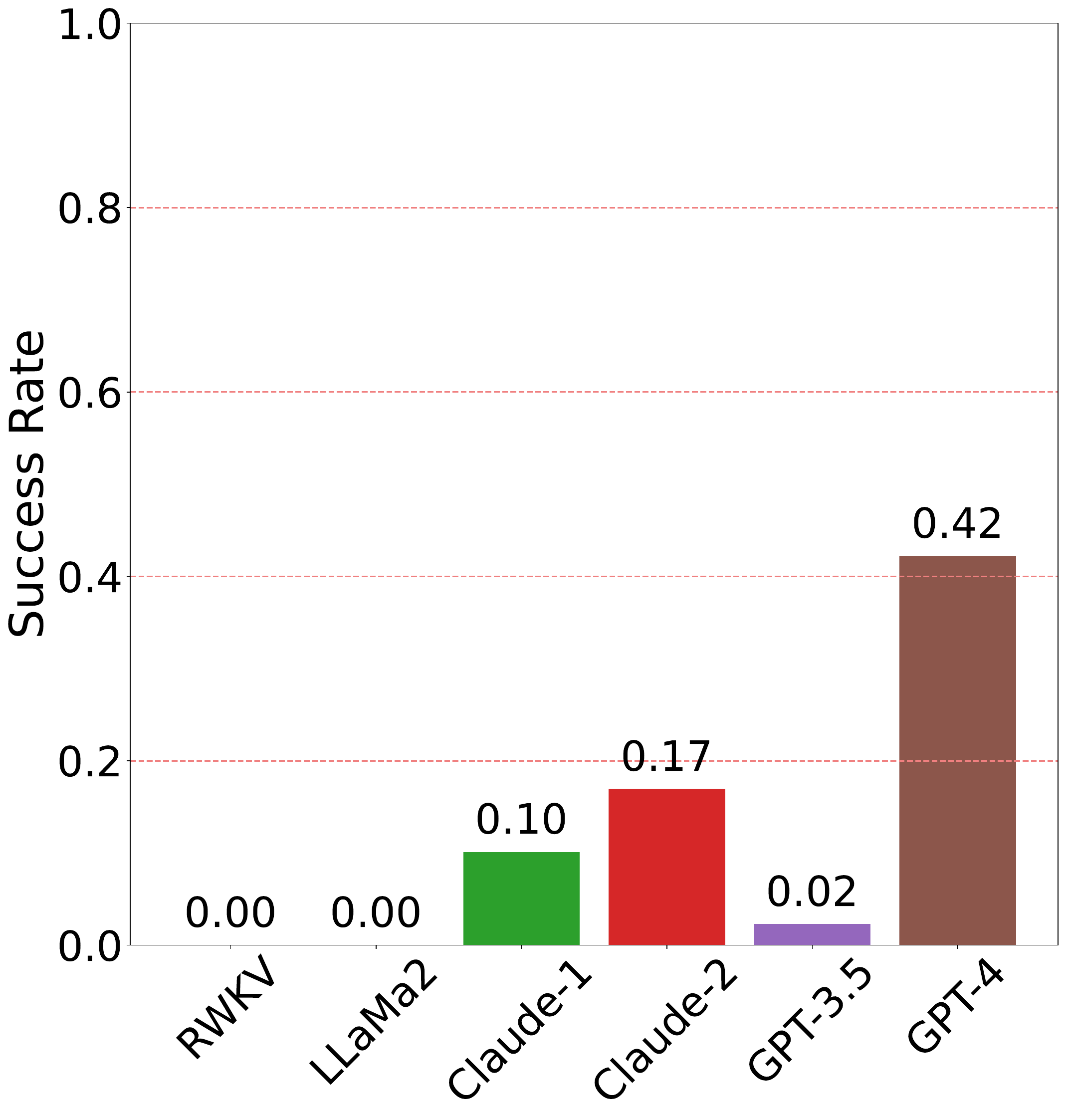}
			\caption{On easy (left) and hard (right) RF questions.}\label{fig:main_rf_rea_acc}
		\end{subfigure}
		\vspace{-4pt}
		\caption{Reasoning accuracy of each model on DF (\subref{fig:main_df_rea_acc}) and RF (\subref{fig:main_rf_rea_acc}) questions, averaged over all $53$ mazes. }\label{fig:main_rea_acc}
	\end{center}
\end{figure*}

\begin{figure*}[!htbp]
	\begin{center}
	    \begin{subfigure}[t]{0.49\linewidth}
                \begin{center}
                \includegraphics[width=0.49\linewidth]{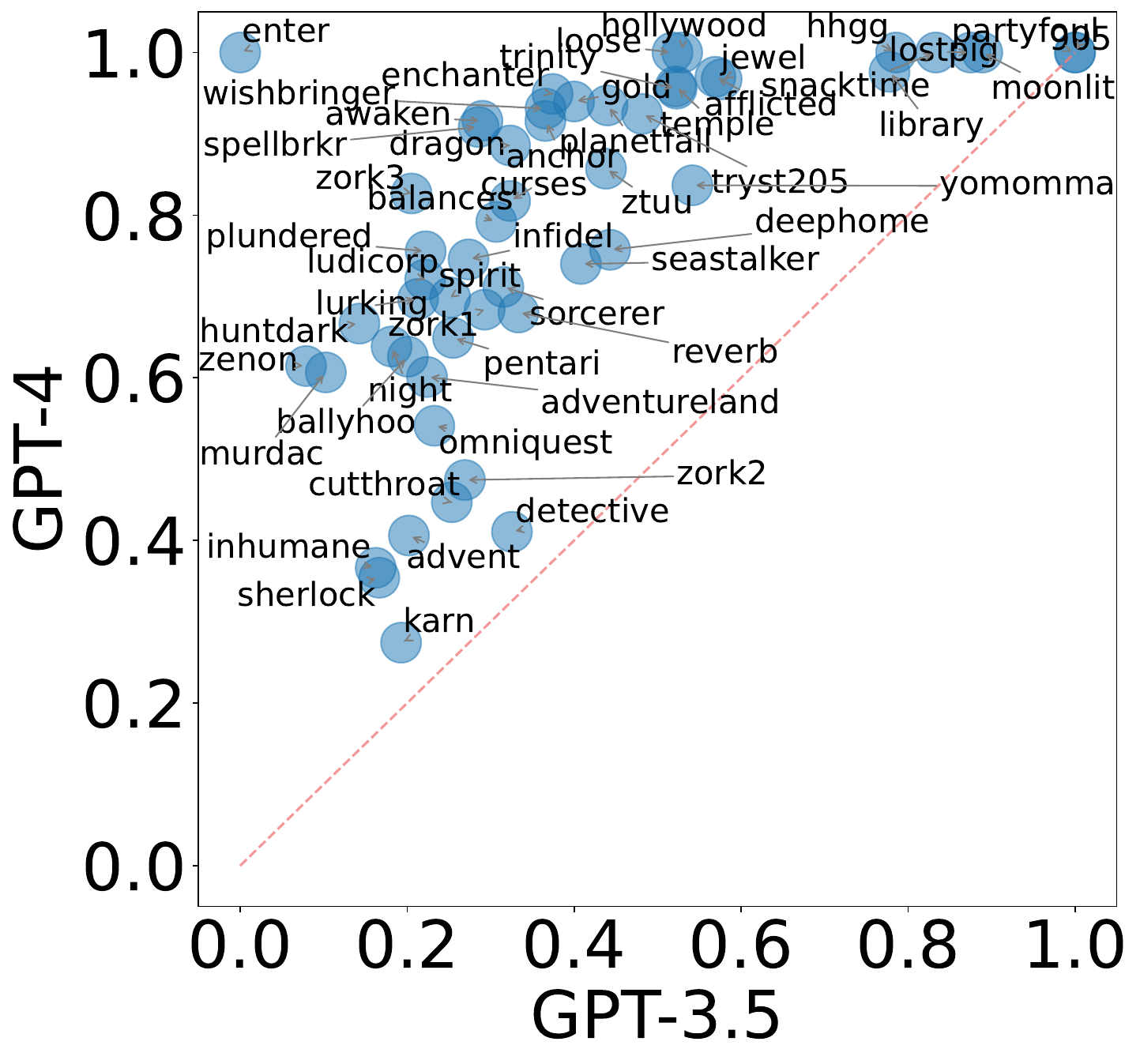}
                \hfill
                \includegraphics[width=0.49\linewidth]{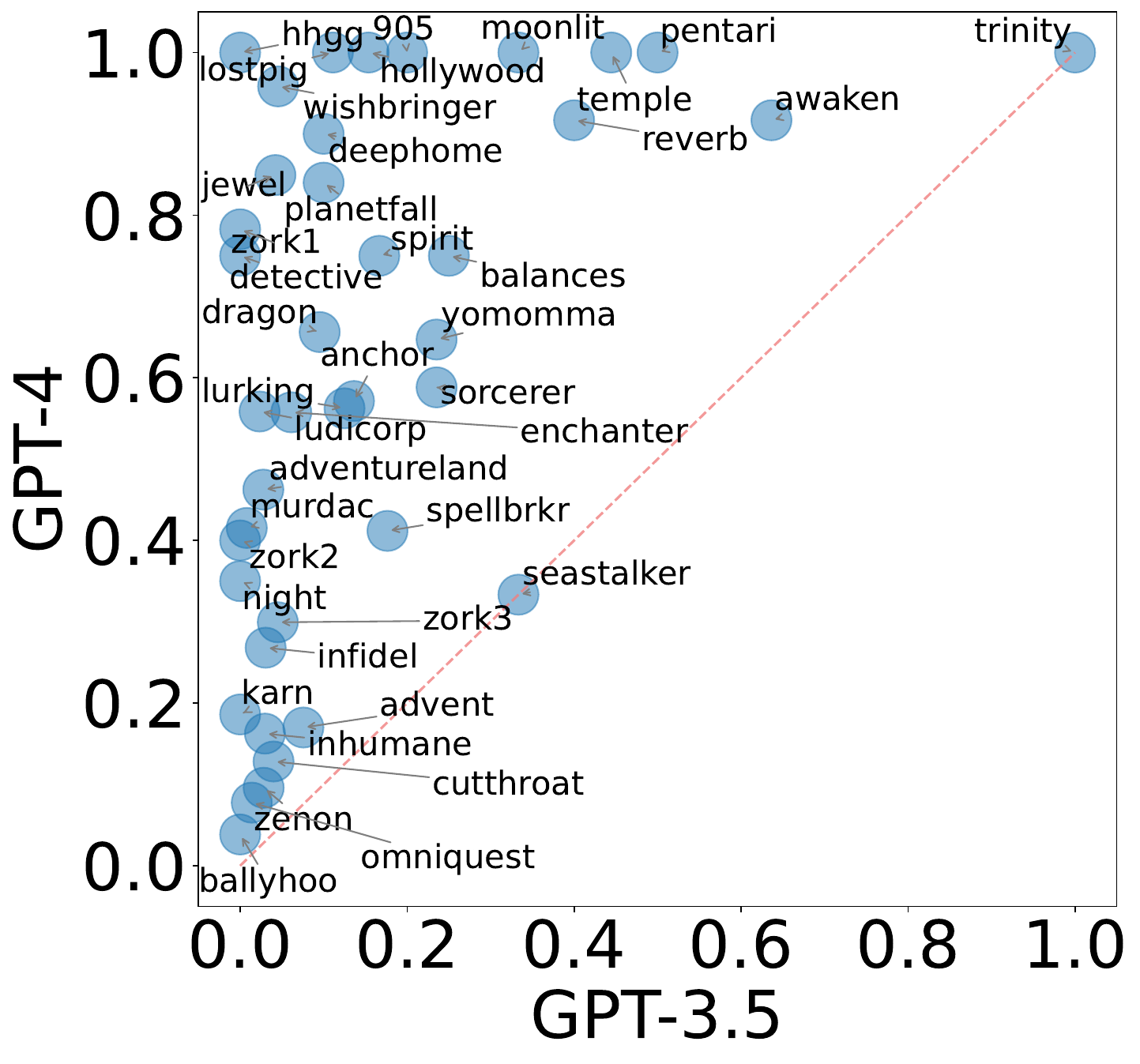}
                \end{center}
			\caption{GPT-3.5 vs.\@ GPT-4 on DF questions.}\label{fig:gpt3vs4_df_ra}
		\end{subfigure}
		\hfill
            \begin{subfigure}[t]{0.49\linewidth}
                \begin{center}
                \includegraphics[width=0.49\linewidth]{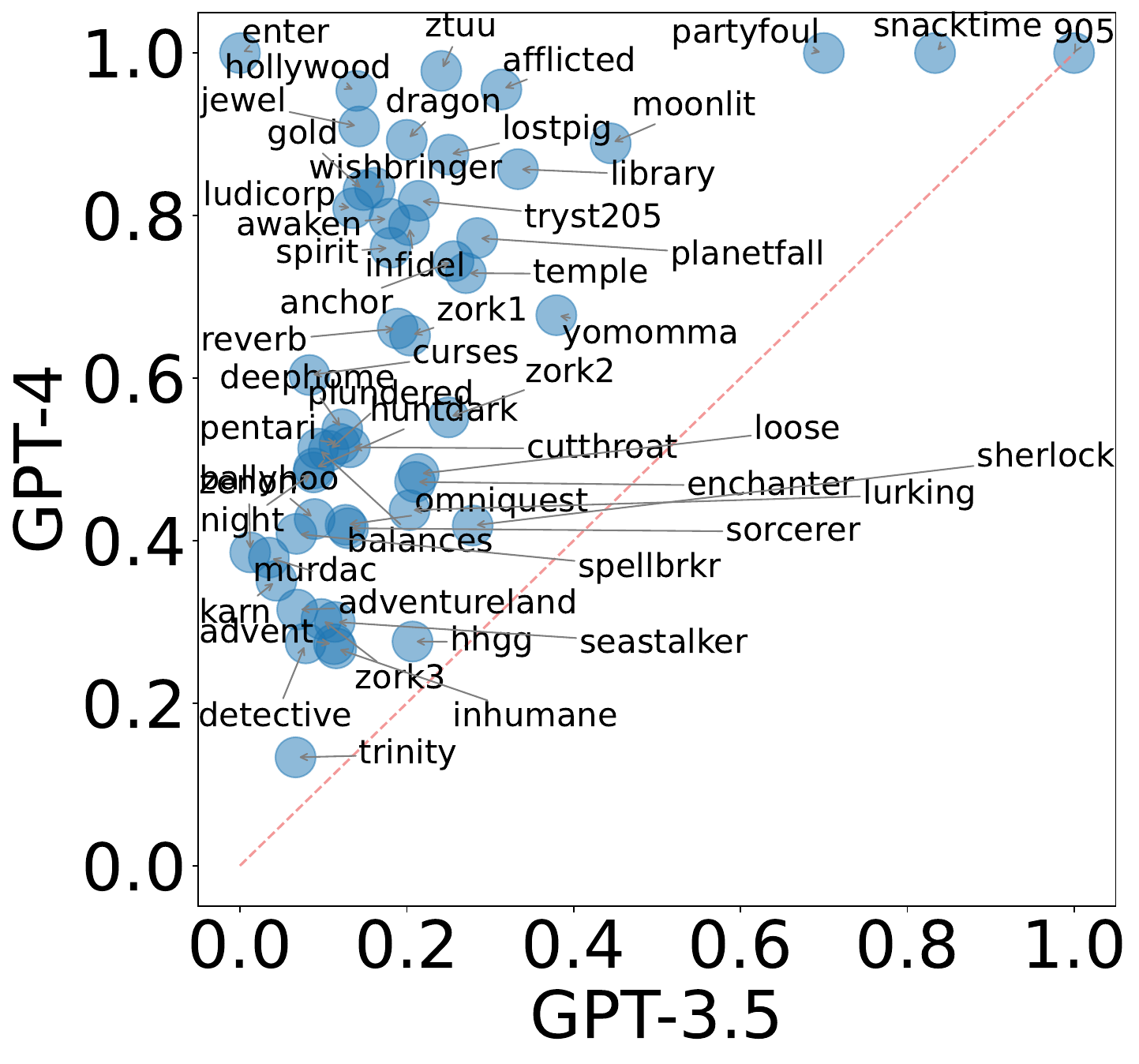}
                \hfill
                \includegraphics[width=0.49\linewidth]{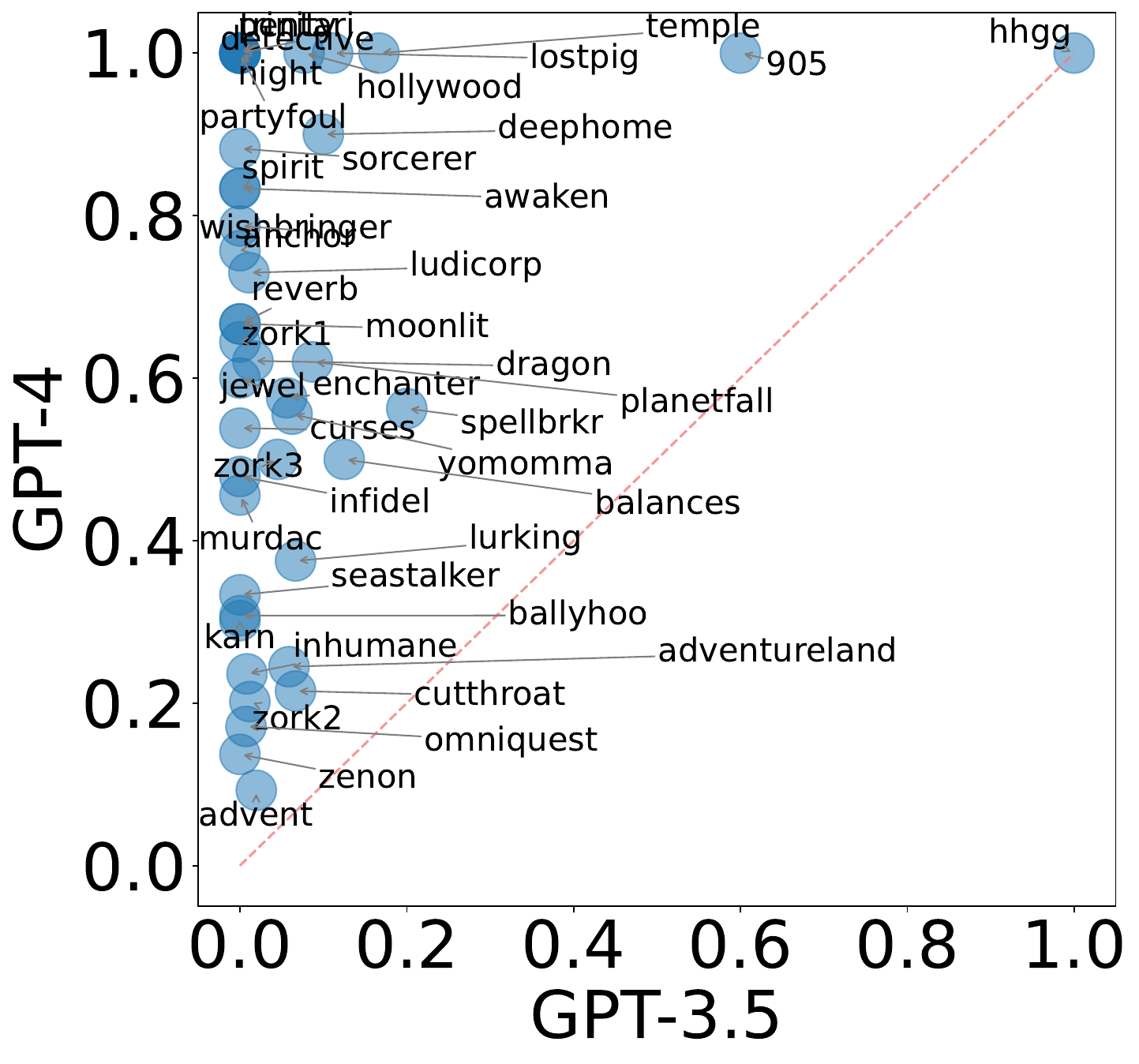}
                \end{center}
			\caption{GPT-3.5 vs.\@ GPT-4 on RF questions.}\label{fig:gpt3vs4_rf_ra}
		\end{subfigure}
		\vspace{-4pt}
		\caption{Reasoning accuracies of GPT-3.5 and GPT-4 broken down into individual games. Similar to \cref{fig:gpt3vs4}, in each subfigure, the left scatterplot is for easy questions while the right is for hard questions.}\label{fig:gpt3vs4_reasoning}
	\end{center}
\end{figure*}

In this section, we present the results of each LLM measured by reasoning accuracy, the metric introduced in \cref{app:eval}. 
Results are shown in \cref{fig:main_rea_acc}, with their pair-wise comparison shown in \cref{tab:reasoningacc}. 
As we can see, the trend measured by this metric is similar to what's shown in \cref{sec:exp}: GPT-4 is the best among all the evaluated models but still suffers a low accuracy. 
\cref{fig:gpt3vs4_reasoning} shows the reasoning accuracies of GPT-3.5 vs.\@ GPT-4 broken down into individual games, showing similar patterns with \cref{fig:gpt3vs4}. 
\begin{table}[htbp]
\small
\setlength{\tabcolsep}{4pt}
\begin{sc}
\begin{subtable}{1.0\linewidth}
\begin{center}
\begin{tabular}{lccccccc}
\toprule
Method & RWKV & Llama-2 & Claude-1 & Claude-2 & GPT-3.5 & GPT-4 & $\overline{{\rm HARD}}|$ \\
\midrule
RWKV & * & 0.00 | 0.02 & 0.00 | 0.15 & 0.00 | 0.18 & 0.00 | 0.05 & 0.00 | 0.41 & * \\
Llama-2 & 0.15 | 0.00 & * & 0.01 | 0.20 & 0.01 | 0.24 & 0.01 | 0.07 & 0.01 | 0.52 & * \\
Claude-1 & 0.51 | 0.00 & 0.62 | 0.15 & * & 0.11 | 0.13 & 0.13 | 0.05 & 0.11 | 0.29 & * \\
Claude-2 & 0.60 | 0.00 & 0.72 | 0.15 & 0.58 | 0.47 & * & 0.16 | 0.05 & 0.13 | 0.29 & * \\
GPT-3.5 & 0.32 | 0.00 & 0.38 | 0.15 & 0.28 | 0.52 & 0.28 | 0.63 & * & 0.05 | 0.35 & * \\
GPT-4 & 0.71 | 0.00 & 0.79 | 0.15 & 0.67 | 0.47 & 0.67 | 0.58 & 0.73 | 0.28 & * & * \\
|$\underline{{\rm EASY}}$ & * & * & * & * & * & * & *\\
\bottomrule
\end{tabular}
\caption{Pairwise comparison on easy (lower left) and hard (higher right) DF questions.}
\begin{tabular}{lccccccc}
\toprule
Method & RWKV & Llama-2 & Claude-1 & Claude-2 & GPT-3.5 & GPT-4 & $\overline{{\rm HARD}}|$ \\
\midrule
RWKV & * & 0.00 | 0.00 & 0.00 | 0.12 & 0.00 | 0.20 & 0.00 | 0.03 & 0.00 | 0.51 & * \\
Llama-2 & 0.02 | 0.00 & * & 0.00 | 0.14 & 0.00 | 0.19 & 0.00 | 0.04 & 0.00 | 0.45 & * \\
Claude-1 & 0.33 | 0.00 & 0.32 | 0.02 & * & 0.10 | 0.17 & 0.12 | 0.02 & 0.10 | 0.42 & * \\
Claude-2 & 0.46 | 0.00 & 0.45 | 0.03 & 0.45 | 0.31 & * & 0.18 | 0.02 & 0.17 | 0.43 & * \\
GPT-3.5 & 0.15 | 0.00 & 0.16 | 0.03 & 0.15 | 0.34 & 0.14 | 0.48 & * & 0.02 | 0.46 & * \\
GPT-4 & 0.56 | 0.00 & 0.55 | 0.02 & 0.53 | 0.31 & 0.53 | 0.45 & 0.56 | 0.15 & * & * \\
|$\underline{{\rm EASY}}$ & * & * & * & * & * & * & *\\
\bottomrule
\end{tabular}
\caption{Pairwise comparison on easy (lower left) and hard (higher right) RF questions.}
\end{center}
\end{subtable}%
\end{sc}
\caption{Reasoning accuracies on DF and RF questions broken down into pairwise comparison. }
\label{tab:reasoningacc}
\end{table}

\subsection{Analysis Details of GPTs}\label{app:analysis}

As discussed in \cref{sec:analysis}, we used regression analysis to understand the effects of the aforementioned variables on model performance. 
In particular, for each model (GPT-3.5 or GPT-4) on each type of question (DF or RF, easy or hard), we ran single-variable linear regression to understand how the success rate varies with each of the variables of interest. 
\cref{tab:permapregression} displays the regression results. 
\begin{table}
\begin{small}
\begin{sc}
\begin{subtable}{\linewidth}
\centering
\begin{tabular}{l|rr|rr|rr|rr}
\hline
\multirow{3}{*}{\centering Metric} & \multicolumn{4}{c|}{GPT-3.5} & \multicolumn{4}{c}{GPT-4} \\ 
\cline{2-9} 
 & \multicolumn{2}{c|}{\centering DF} & \multicolumn{2}{c|}{\centering RF} & \multicolumn{2}{c|}{\centering DF} & \multicolumn{2}{c}{\centering RF} \\ 
\cline{2-9} 
 & \multicolumn{1}{c|}{\centering $\beta$} & \multicolumn{1}{c|}{\centering $p$} & \multicolumn{1}{c|}{\centering $\beta$} & \multicolumn{1}{c|}{\centering $p$} & \multicolumn{1}{c|}{\centering $\beta$} & \multicolumn{1}{c|}{\centering $p$} & \multicolumn{1}{c|}{\centering $\beta$} & \multicolumn{1}{c}{\centering $p$} \\ 
\cline{2-9} 
\# locations & $-0.076$ & $0.001$ & $-0.084$ & $0.000$ & $-0.073$ & $0.000$ & $-0.120$ & $0.000$ \\ 
\# exp edges & $-0.068$ & $0.003$ & $-0.081$ & $0.001$ & $-0.067$ & $0.000$ & $-0.096$ & $0.003$ \\ 
\# conf locations & $-0.056$ & $0.017$ & $-0.067$ & $0.006$ & $-0.065$ & $0.000$ & $-0.088$ & $0.006$ \\ 
avg len easy & $-0.066$ & $0.005$ & $-0.081$ & $0.001$ & $-0.066$ & $0.000$ & $-0.131$ & $0.000$ \\ 
avg len scene & $-0.037$ & $0.119$ & $-0.002$ & $0.939$ & $0.031$ & $0.036$ & $0.069$ & $0.035$ \\ 
\hline
\end{tabular}
\vspace{-4pt}
\caption{Regression analysis results on easy questions.}
\vspace{4pt}
\centering
\begin{tabular}{l|rr|rr|rr|rr}
\hline
\multirow{3}{*}{\centering Metric} & \multicolumn{4}{c|}{GPT-3.5} & \multicolumn{4}{c}{GPT-4} \\ 
\cline{2-9} 
 & \multicolumn{2}{c|}{\centering DF} & \multicolumn{2}{c|}{\centering RF} & \multicolumn{2}{c|}{\centering DF} & \multicolumn{2}{c}{\centering RF} \\ 
\cline{2-9} 
 & \multicolumn{1}{c|}{\centering $\beta$} & \multicolumn{1}{c|}{\centering $p$} & \multicolumn{1}{c|}{\centering $\beta$} & \multicolumn{1}{c|}{\centering $p$} & \multicolumn{1}{c|}{\centering $\beta$} & \multicolumn{1}{c|}{\centering $p$} & \multicolumn{1}{c|}{\centering $\beta$} & \multicolumn{1}{c}{\centering $p$} \\ 
\cline{2-9} 
\# locations & $-0.049$ & $0.088$ & $-0.055$ & $0.043$ & $-0.083$ & $0.007$ & $-0.115$ & $0.006$ \\ 
\# exp edges & $-0.052$ & $0.070$ & $-0.064$ & $0.018$ & $-0.080$ & $0.010$ & $-0.107$ & $0.011$ \\ 
\# imp edges & $-0.059$ & $0.038$ & $-0.032$ & $0.252$ & $-0.118$ & $0.000$ & $-0.152$ & $0.000$ \\ 
\# conf locations & $-0.045$ & $0.125$ & $-0.055$ & $0.043$ & $-0.068$ & $0.030$ & $-0.081$ & $0.061$ \\ 
avg len hard & $-0.072$ & $0.011$ & $-0.055$ & $0.046$ & $-0.094$ & $0.002$ & $-0.122$ & $0.004$ \\ 
avg \# imp in hard & $-0.057$ & $0.046$ & $-0.022$ & $0.428$ & $-0.081$ & $0.009$ & $-0.096$ & $0.024$ \\ 
avg len scene & $0.032$ & $0.268$ & $0.050$ & $0.069$ & $0.084$ & $0.007$ & $0.105$ & $0.013$ \\ 
\hline
\end{tabular}
\vspace{-4pt}
\caption{Regression analysis results on hard questions.}
\end{subtable}
\end{sc}
\end{small}
\vspace{-8pt}
\caption{Regression analysis results, where $\beta$ is the regression coefficient and $p$ denotes the $p$-value. When $p < 0.001$, we write $0.000$ for presentation simplicity.}
\label{tab:permapregression}
\end{table}

In our pilot experiments, we ran a multivariate regression analysis that used the aforementioned variables jointly. However, the results of this regression are misleading: due to colinearity among the explanatory variables, the estimated coefficients are unreliable and the $p$-values are inflated. 
We also tried principled component regression, but the first principle component has nearly equal loadings across all the variables, making it inconvenient to interpret the results.

\subsection{Results of Search-Based Approaches}\label{app:search}
We also explored an approach that first maps a walkthrough to a symbolic graph and then uses a search algorithm to answer the given (DF or RF) question.
However, we found it to be very challenging to translate natural language walkthrough into searchable symbolic graphs in the first place, making this approach not promising. 

We tried two different settings. 
In the first setting, we did 5-shot prompting and each in-context example is a single \mvar{S}-\mvar{A}-\mvar{D} triplet; the examples are the first five steps in the walkthrough. 
In the second setting, we did 0-shot prompting and the prompt includes the entire walkthrough; the LLM has to generate a sequence of \mvar{S}-\mvar{A}-\mvar{D} triplets which then could be used to construct the graph. 
For each setting, we ran experiments on the following six mazes: Zork-I, Night, Partyfoul, Plundered, Spirit, and Temple.

In both settings, GPT-3.5 yields a low accuracy of the \mvar{S}-\mvar{A}-\mvar{D} triplet completion, indicating the difficulty of translating natural language into symbolic graphs. 
In the first setting, the average accuracy is 70.5 \%. 
In the second setting, the average success rate is 6.3 \%. 
The second setting is more challenging because any mistake would cause all subsequent steps to be incorrect. 
Some LLM mistakes and our error analysis can be found at \url{https://github.com/Oaklight/mango/tree/camera-ready/utils/supp_exp}.%

\subsection{More Llama Results}\label{app:llama}
\begin{figure*}[!htbp]
	\begin{center}
	    \begin{subfigure}[t]{0.48\linewidth}
                \includegraphics[width=0.49\linewidth]{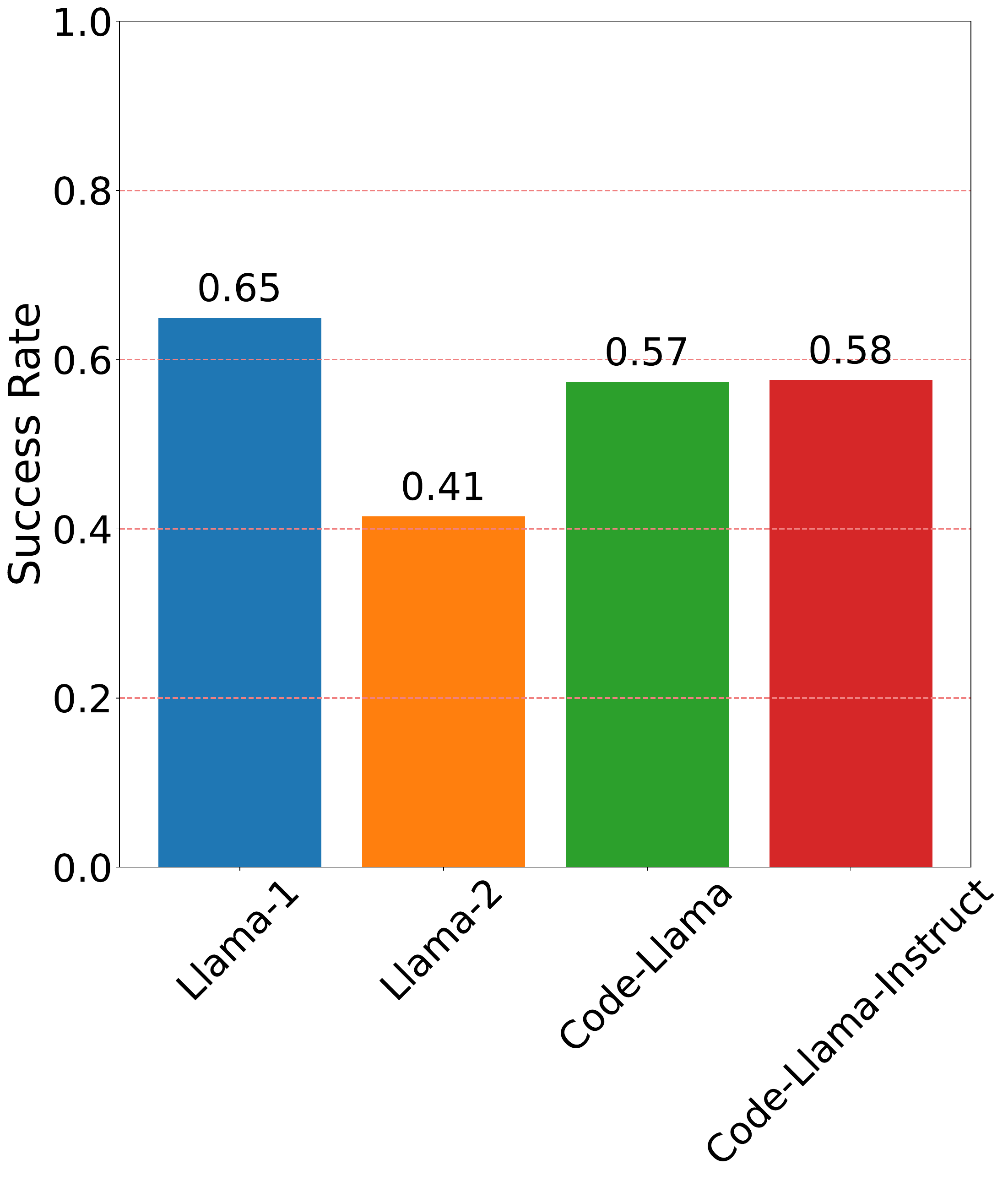}
                \hfill
                \includegraphics[width=0.49\linewidth]{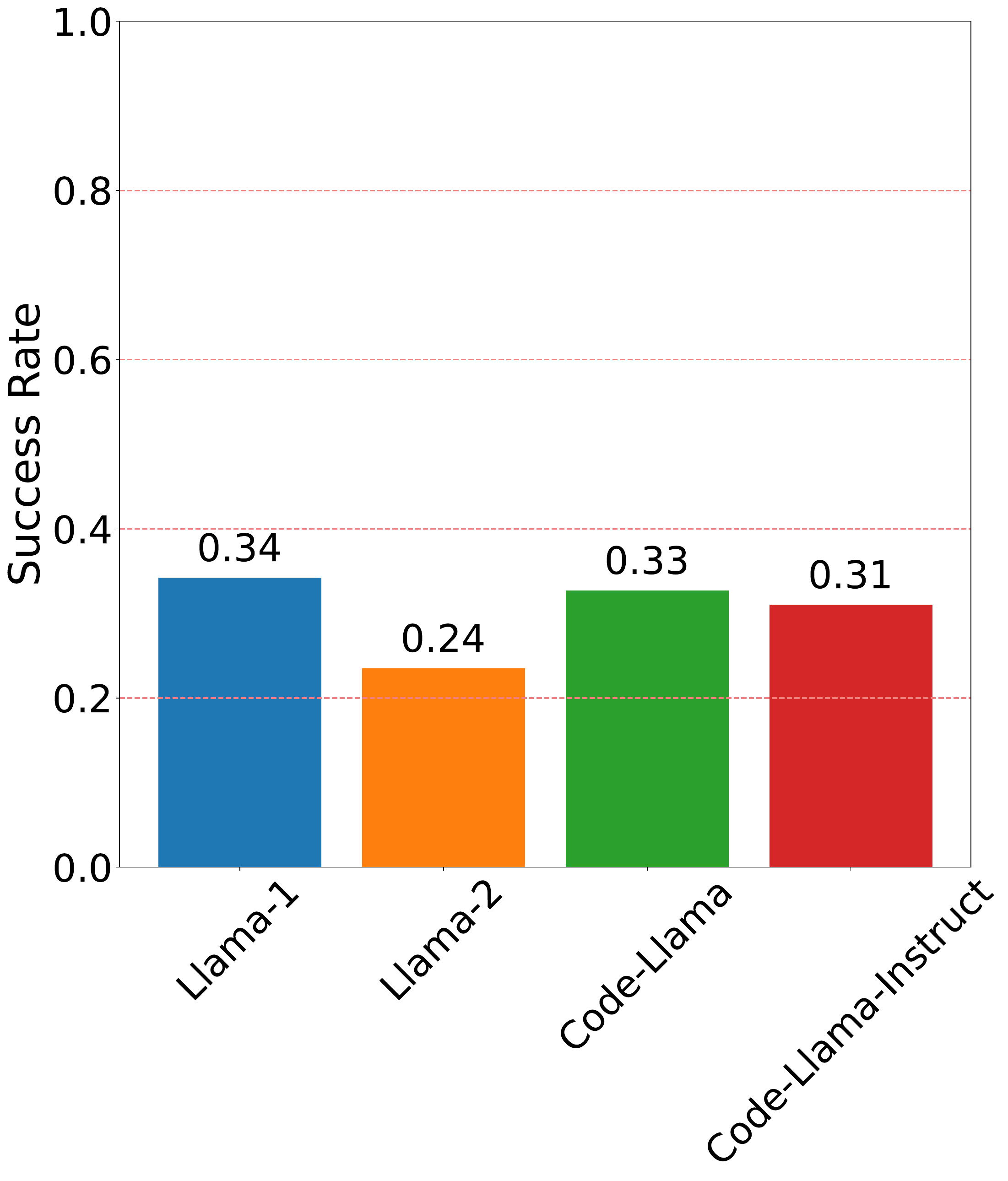}
			\caption{On easy (left) and hard (right) DF questions.}\label{fig:main_df_sr_llama}
		\end{subfigure}
		\hfill
            \begin{subfigure}[t]{0.48\linewidth}
			\includegraphics[width=0.49\linewidth]{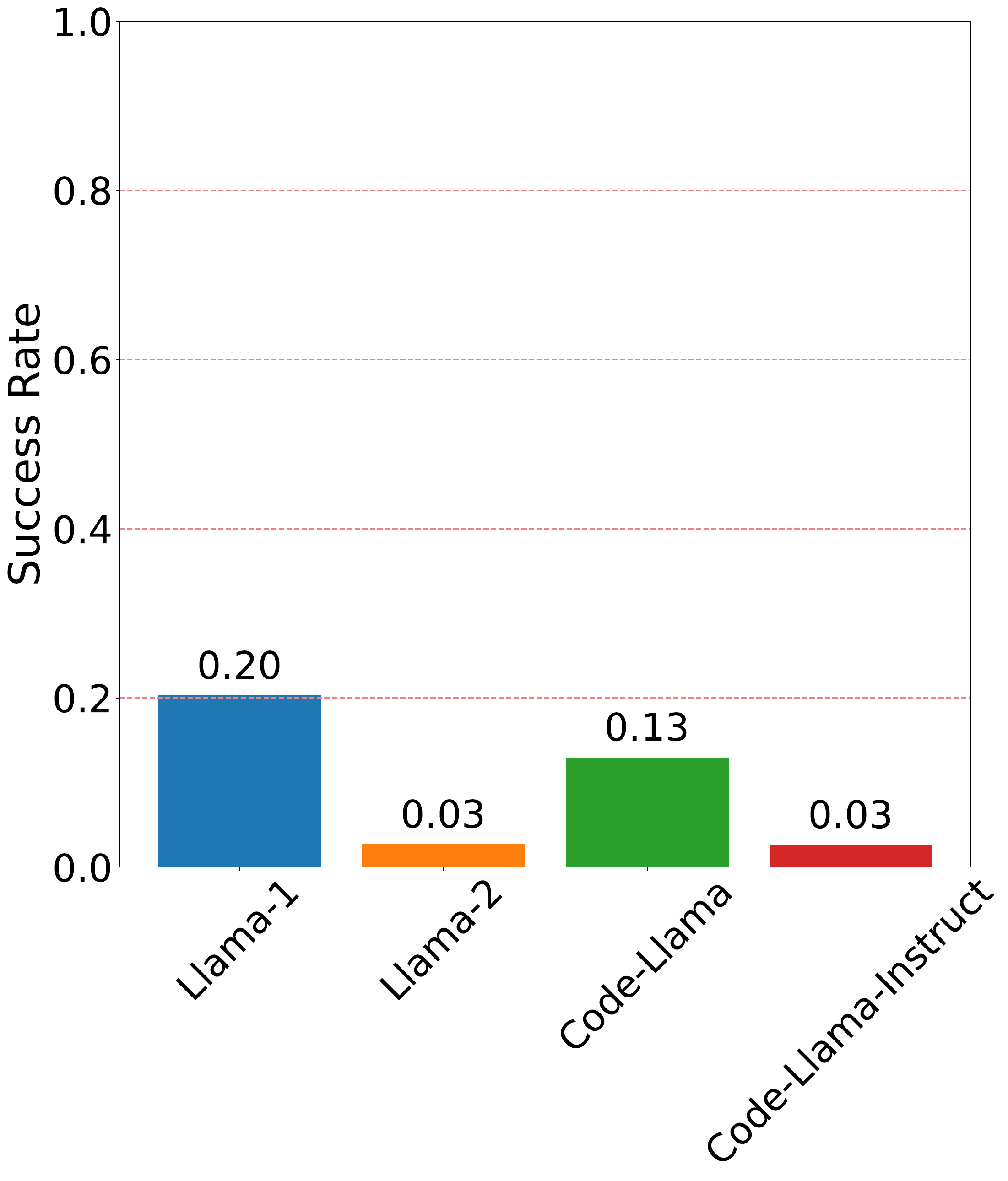}
                \hfill
                \includegraphics[width=0.49\linewidth]{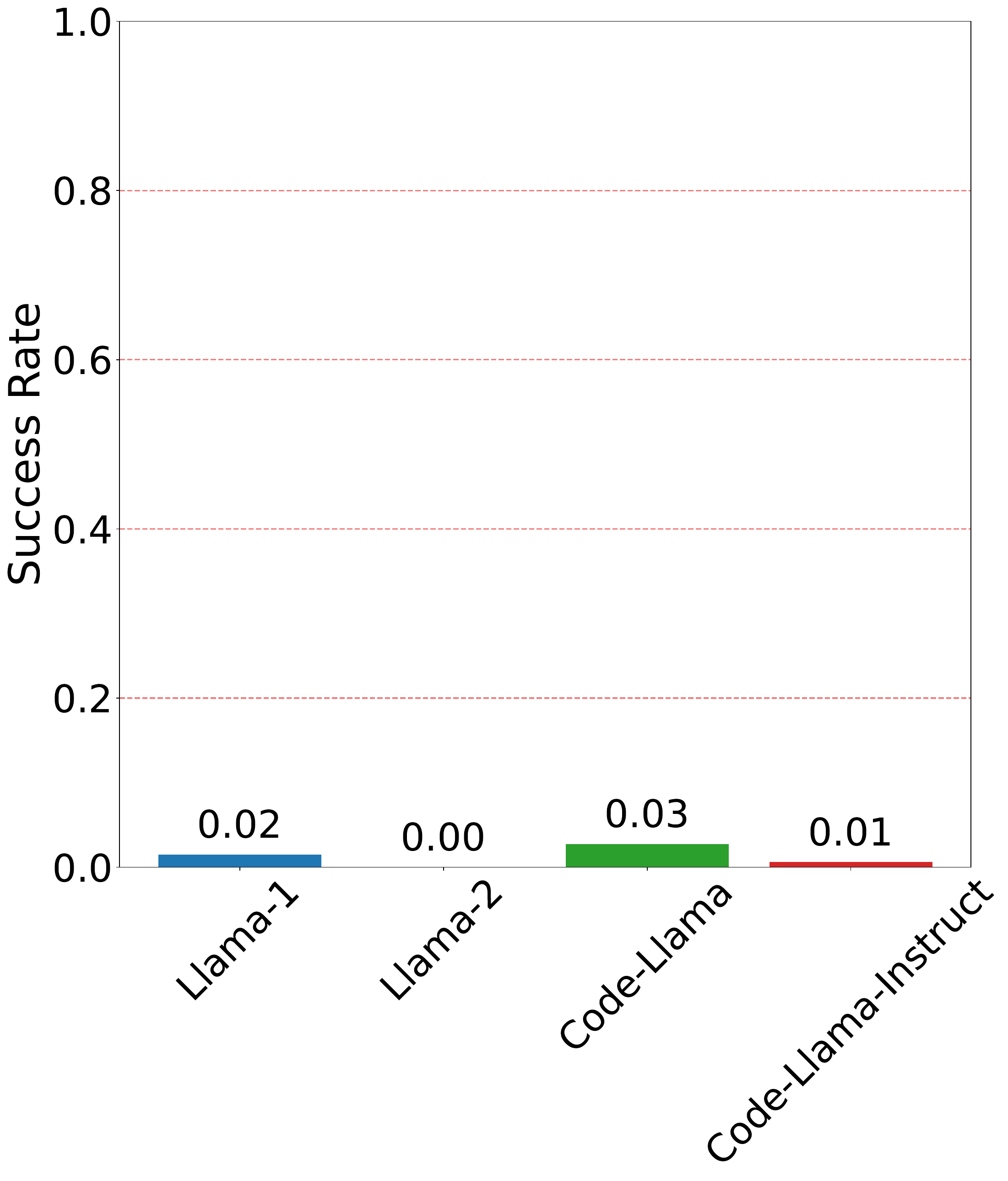}
			\caption{On easy (left) and hard (right) RF questions.}\label{fig:main_rf_sr_llama}
		\end{subfigure}
		\vspace{-4pt}
		\caption{Success rates of the Llama-1 and Llama-2 on DF (\subref{fig:main_df_sr_llama}) and RF (\subref{fig:main_rf_sr_llama}) questions, averaged over all $53$ mazes. }\label{fig:main_sr_llama}
	\end{center}
\end{figure*}

\begin{figure*}[!htbp]
	\begin{center}
	    \begin{subfigure}[t]{0.48\linewidth}
                \includegraphics[width=0.49\linewidth]{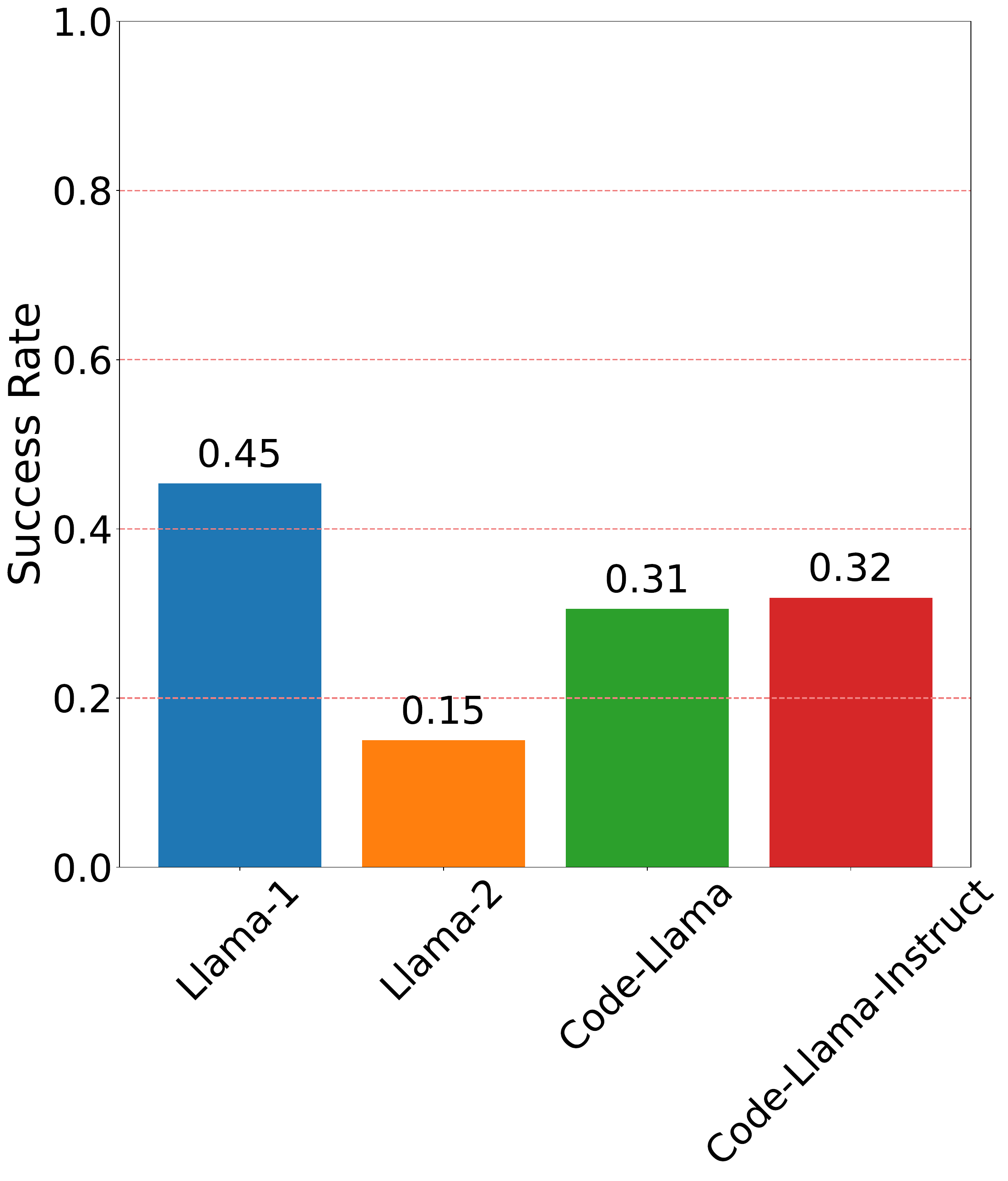}
                \hfill
                \includegraphics[width=0.49\linewidth]{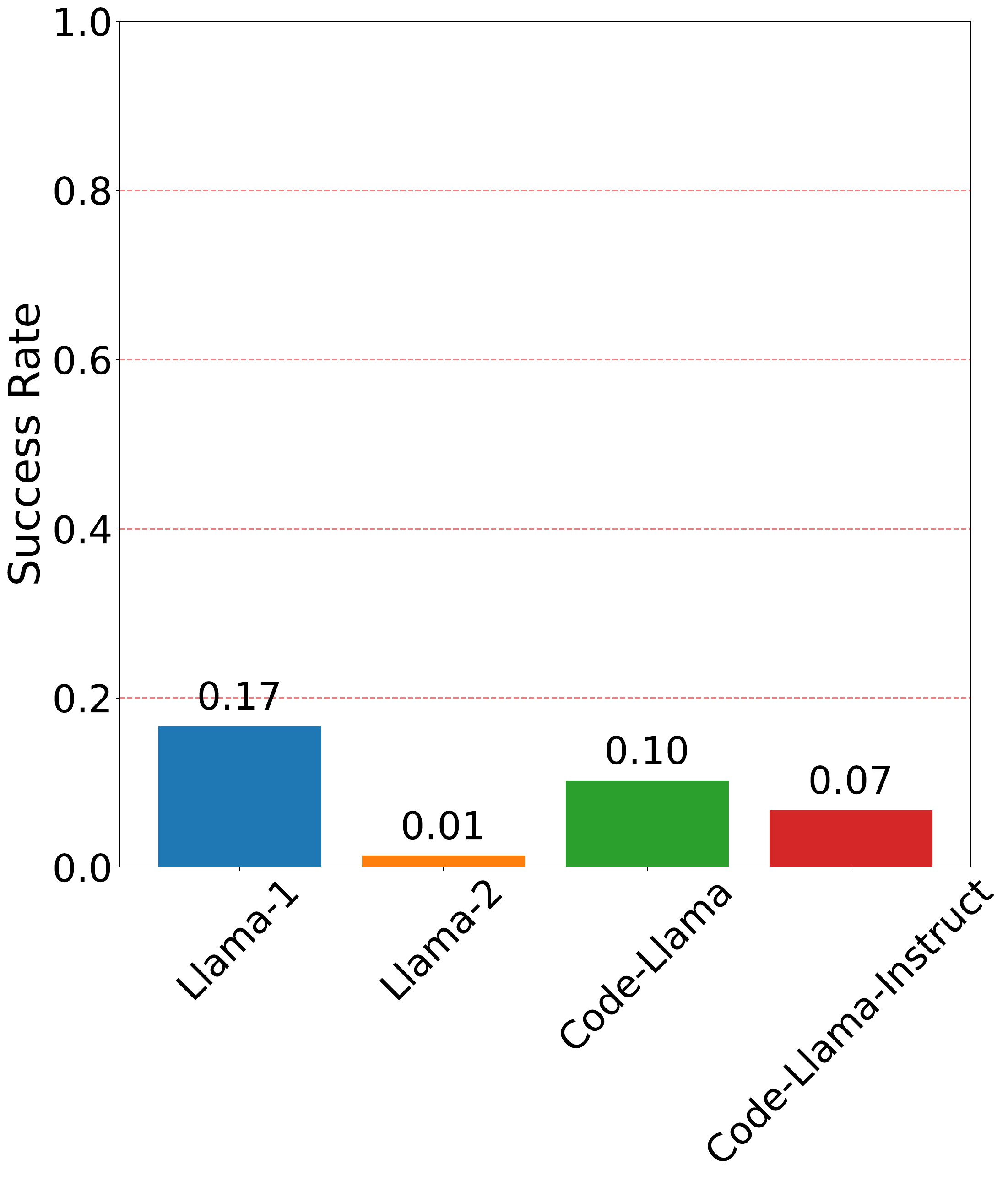}
			\caption{On easy (left) and hard (right) DF questions.}\label{fig:main_df_rea_acc_llama}
		\end{subfigure}
		\hfill
            \begin{subfigure}[t]{0.48\linewidth}
			\includegraphics[width=0.49\linewidth]{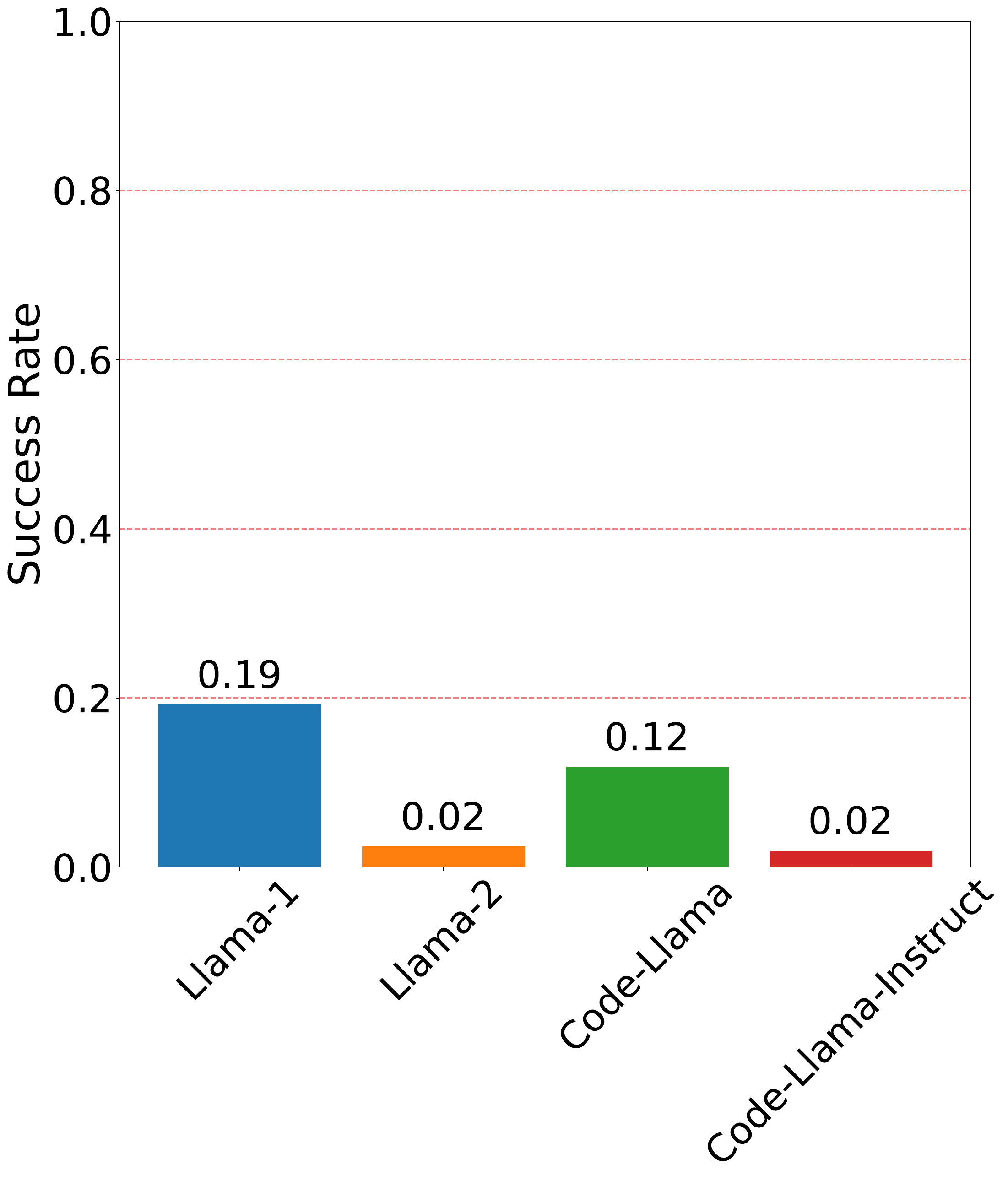}
                \hfill
                \includegraphics[width=0.49\linewidth]{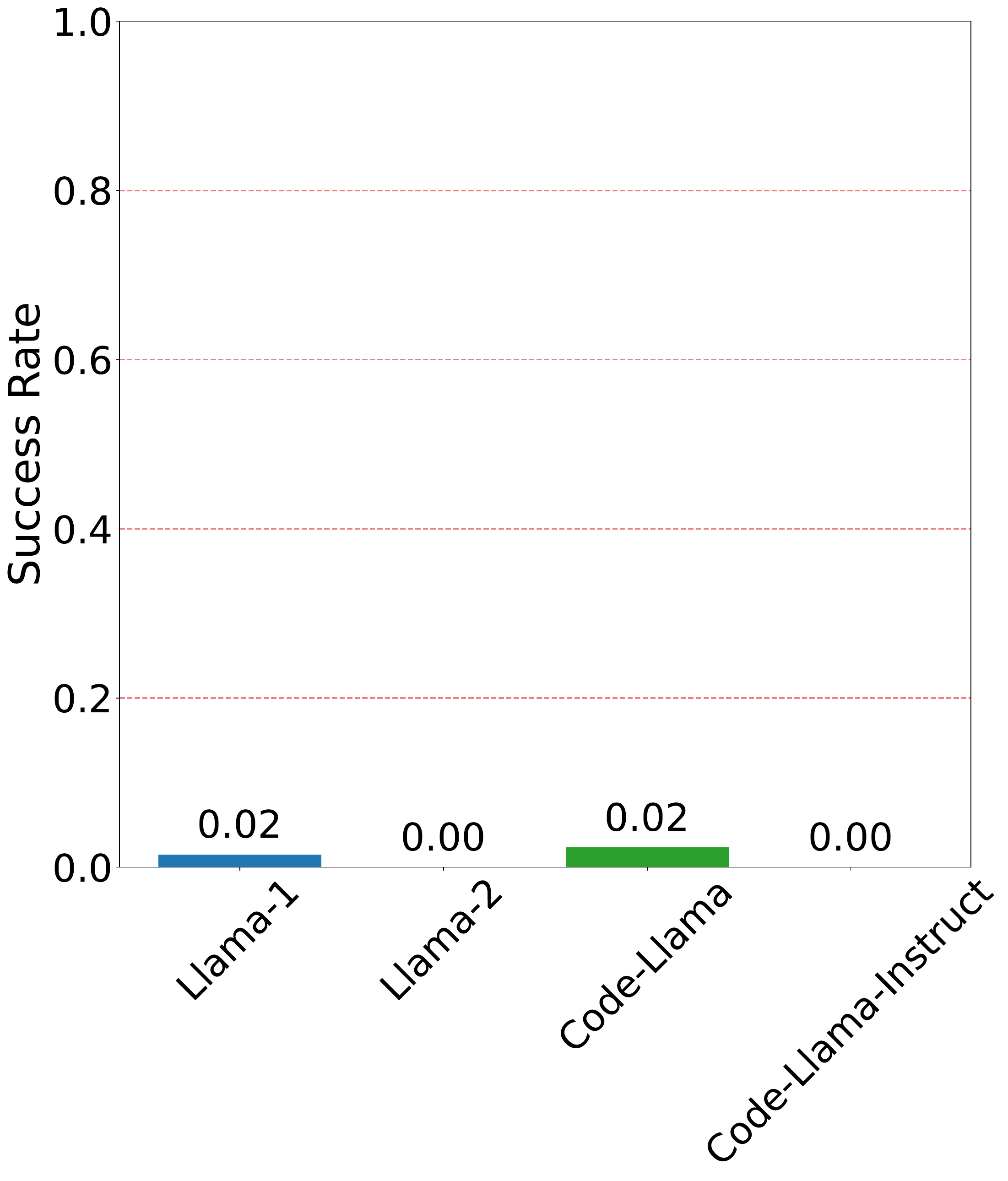}
			\caption{On easy (left) and hard (right) RF questions.}\label{fig:main_rf_rea_acc_llama}
		\end{subfigure}
		\vspace{-4pt}
		\caption{Reasoning accuracy of the Llama-1 and Llama-2 on DF (\subref{fig:main_df_rea_acc_llama}) and RF (\subref{fig:main_rf_rea_acc_llama}) questions, averaged over all $53$ mazes.}\label{fig:main_rea_acc_llama}
	\end{center}
\end{figure*}

The Llama-2 we used in the experiments of \cref{sec:exp} is the base model. 
We also experimented with the the 32.5B Llama-1 model released earlier~\citep{touvron2023llama}, 13b CodeLlama and 13b CodeLlama-Instruct~\citep{codellama}. 
Llama-1 has a significantly smaller context window, so it had to read shorter walkthrough prefixes and answer fewer questions. 
The results of comparing different Llamas are in \cref{fig:main_sr_llama}, \cref{fig:main_rea_acc_llama}, \cref{tab:successratesllama}, and \cref{tab:reasoningaccllama}. 
As we can see, both CodeLlama and CodeLlama-Instruct outperform Llama-2, and Llama-1 performs the best in the group. 
In the main paper, we present the results of Llama-2-base because it is commonly considered to be the standard choice of the Llama series. 
\begin{table}[!htbp]
\small
\setlength{\tabcolsep}{4pt}
\begin{sc}
\begin{subtable}{1.0\linewidth}
\begin{center}
\begin{tabular}{lccccc}
\toprule
Method & Llama-1 & Llama-2 & Code-Llama & Code-Llama-Instruct & $\overline{{\rm HARD}}|$ \\
\midrule
Llama-1 & * & 0.29 | 0.25 & 0.34 | 0.38 & 0.34 | 0.38 & * \\
Llama-2 & 0.47 | 0.65 & * & 0.24 | 0.30 & 0.24 | 0.30 & * \\
Code-Llama & 0.65 | 0.67 & 0.60 | 0.44 & * & 0.31 | 0.31 & * \\
Code-Llama-Instruct & 0.65 | 0.67 & 0.60 | 0.44 & 0.58 | 0.58 & * & * \\
|$\underline{{\rm EASY}}$ & * & * & * & * & * \\
\bottomrule
\end{tabular}
\caption{Pairwise comparison on easy (lower left) and hard (higher right) DF questions.}
\begin{tabular}{lccccc}
\toprule
Method & Llama-1 & Llama-2 & Code-Llama & Code-Llama-Instruct & $\overline{{\rm HARD}}|$ \\
\midrule
Llama-1 & * & 0.00 | 0.00 & 0.00 | 0.00 & 0.00 | 0.00 & * \\
Llama-2 & 0.05 | 0.18 & * & 0.00 | 0.01 & 0.00 | 0.01 & * \\
Code-Llama & 0.03 | 0.20 & 0.04 | 0.03 & * & 0.01 | 0.01 & * \\
Code-Llama-Instruct & 0.03 | 0.20 & 0.04 | 0.03 & 0.03 | 0.03 & * & * \\
|$\underline{{\rm EASY}}$ & * & * & * & * & * \\
\bottomrule
\end{tabular}
\caption{Pairwise comparison on easy (lower left) and hard (higher right) RF questions.}
\end{center}
\end{subtable}%
\end{sc}
\caption{Llamas' success rates on DF and RF questions broken down into pairwise comparison. }
\label{tab:successratesllama}
\end{table}
\begin{table}[!htbp]
\small
\setlength{\tabcolsep}{4pt}
\begin{sc}
\begin{subtable}{1.0\linewidth}
\begin{center}
\begin{tabular}{lccccc}
\toprule
Method & Llama-1 & Llama-2 & Code-Llama & Code-Llama-Instruct & $\overline{{\rm HARD}}|$ \\
\midrule
Llama-1 & * & 0.09 | 0.01 & 0.16 | 0.19 & 0.16 | 0.19 & * \\
Llama-2 & 0.25 | 0.40 & * & 0.02 | 0.06 & 0.02 | 0.06 & * \\
Code-Llama & 0.47 | 0.47 & 0.34 | 0.18 & * & 0.07 | 0.07 & * \\
Code-Llama-Instruct & 0.47 | 0.47 & 0.34 | 0.18 & 0.32 | 0.32 & * & * \\
|$\underline{{\rm EASY}}$ & * & * & * & * & * \\
\bottomrule
\end{tabular}
\caption{Pairwise comparison on easy (lower left) and hard (higher right) DF questions.}
\begin{tabular}{lccccc}
\toprule
Method & Llama-1 & Llama-2 & Code-Llama & Code-Llama-Instruct & $\overline{{\rm HARD}}|$ \\
\midrule
Llama-1 & * & 0.00 | 0.00 & 0.00 | 0.00 & 0.00 | 0.00 & * \\
Llama-2 & 0.05 | 0.18 & * & 0.00 | 0.00 & 0.00 | 0.00 & * \\
Code-Llama & 0.02 | 0.18 & 0.03 | 0.03 & * & 0.00 | 0.00 & * \\
Code-Llama-Instruct & 0.02 | 0.18 & 0.03 | 0.03 & 0.02 | 0.02 & * & * \\
|$\underline{{\rm EASY}}$ & * & * & * & * & * \\
\bottomrule
\end{tabular}
\caption{Pairwise comparison on easy (lower left) and hard (higher right) RF questions.}
\end{center}
\end{subtable}%
\end{sc}
\caption{Llamas' reasoning accuracies on DF and RF questions broken down into pairwise comparison.}
\label{tab:reasoningaccllama}
\end{table}

\clearpage

\subsection{Human Performance Details}\label{app:human}
To measure human agreement, we computed the mean square error (MSE) for each metric across the questions that were answered by more than one human rater. 
As shown in \cref{tab:inter_annotator}, the human agreement turns out to be high. 
\begin{table}[h]
\small
\begin{sc}
\begin{center}
\begin{tabular}{lcc|cc|cc}
\hline
& \multicolumn{2}{c|}{All} & \multicolumn{2}{c|}{Easy} & \multicolumn{2}{c}{Hard} \\
Task Type & SR & RA & SR  & RA & SR & RA \\
\hline
Route Finding & 0.1442 & 0.3333 & 0.0425 & 0.1429 & 0.5000 & 1.0000 \\
Destination Finding & 0.0000 & 0.6000 & 0.0000 & 0.5714 & 0.0000 & 0.6667 \\
\hline
\end{tabular}
\end{center}
\end{sc}
\caption{Human agreement measured by MSE. SR stands for success rate, while RA denotes reasoning accuracy.}
\label{tab:inter_annotator}
\end{table}

\subsection{More Results About Playing Minigames}\label{app:minigame}
In \cref{sec:downstream}, we have shown that a strong mapping and navigation ability can help an LLM achieve better performance in playing the minigames. 
Now we show the fine-grained results for that set of experiments. 
Precisely, \cref{tab:minigamedetails} shows the spelled-out success rates of GPT-3.5 and GPT-4 on each maze, where $M/N$ means that $M$ of $N$ minigames were successfully played by this model. 
Since GPT-4 by default has a larger context window size (8K) than GPT-3.5 (4K), in order to make a fair comparison, we restrict the context window size to 4K for both GPTs. %

\begin{table}[htbp]
\small
\setlength{\tabcolsep}{4pt}
\begin{sc}
\begin{center}
    \begin{tabular}{lrrrr}
    \toprule
\multirow{2}{*}{Games} & \multicolumn{2}{c}{GPT-3.5} & \multicolumn{2}{c}{GPT-4} \\
\cmidrule(lr){2-3} \cmidrule(lr){4-5}
&  w/o maps & w/ maps & w/o maps & w/ maps \\
        \midrule
        905 & 0/0 & 0/0 & 0/0 & 0/0 \\ 
        advent & 3/9 & 3/9 & 4/9 & 5/9 \\ 
        adventureland & 1/7 & 1/7 & 3/7 & 5/7  \\ 
        afflicted & 0/12 & 1/12 & 4/12 & 4/12 \\
        anchor & 0/2 & 0/2 & 2/2 & 2/2  \\ 
        awaken & 0/7 & 0/7 & 3/7 & 7/7 \\ 
        balances & 0/3 & 1/3 & 2/3 & 2/3 \\ 
        ballyhoo & 1/6 & 1/6 & 3/6 & 2/6 \\ 
        curses & 2/7 & 3/7 & 2/7 & 6/7 \\ 
        cutthroat & 4/5 & 5/5 & 4/5 & 4/5 \\ 
        deephome & 3/10 & 6/10 & 8/10 & 10/10 \\ 
        detective & 4/6 & 5/6 & 6/6 & 6/6 \\ 
        dragon & 0/6 & 1/6 & 0/6 & 2/6 \\ 
        enchanter & 0/1 & 1/1 & 1/1 & 1/1  \\ 
        enter & 0/0 & 0/0 & 0/0 & 0/0 \\ 
        gold & 1/7 & 2/7 & 3/7 & 4/7 \\ 
        hhgg & 0/0 & 0/0 & 0/0 & 0/0  \\ 
        hollywood & 2/6 & 2/6 & 4/6 & 5/6\\ 
        huntdark & 0/0 & 0/0 & 0/0 & 0/0  \\ 
        infidel & 1/4 & 2/4 & 2/4 & 2/4  \\ 
        inhumane & 1/2 & 2/2 & 2/2 & 2/2 \\ 
        jewel & 0/10 & 1/10 & 4/10 & 7/10 \\ 
        karn & 0/8 & 0/8 & 4/8 & 5/8 \\ 
        library & 1/5 & 2/5 & 5/5 & 5/5  \\ 
        loose & 3/4 & 3/4 & 3/4 & 3/4 \\ 
        lostpig & 0/1 & 0/1 & 0/1 & 1/1 \\ 
        ludicorp & 1/13 & 2/13 & 9/13 & 10/13 \\ 
        lurking & 1/4 & 0/4 & 0/4 & 2/4 \\ 
        moonlit & 0/2 & 2/2 & 2/2 & 2/2 \\
        murdac & 1/2 & 1/2 & 2/2 & 2/2 \\ 
        night & 5/16 & 8/16 & 9/16 & 13/16 \\ 
        omniquest & 1/11 & 1/11 & 1/11 & 7/11 \\ 
        partyfoul & 0/2 & 1/2 & 1/2 & 1/2 \\ 
        pentari & 4/13 & 3/13 & 4/13 & 8/13 \\ 
        planetfall & 3/6 & 4/6 & 3/6 & 5/6  \\ 
        plundered & 0/0 & 0/0 & 0/0 & 0/0 \\ 
        reverb & 0/2 & 0/2 & 1/2 & 1/2 \\ 
        seastalker & 0/2 & 0/2 & 0/2 & 2/2 \\ 
        sherlock & 3/3 & 3/3 & 3/3 & 3/3 \\ 
        snacktime & 0/3 & 0/3 & 3/3 & 3/3 \\ 
        sorcerer & 4/5 & 4/5 & 3/5 & 4/5 \\ 
        spellbrkr & 1/1 & 1/1 & 1/1 & 1/1 \\ 
        spirit & 4/14 & 5/14 & 8/14 & 10/14  \\ 
        temple & 0/2 & 1/2 & 1/2 & 2/2  \\ 
        trinity & 0/0 & 0/0 & 0/0 & 0/0  \\ 
        tryst205 & 0/3 & 1/3 & 0/3 & 1/3  \\ 
        wishbringer & 0/1 & 1/1 & 0/1 & 0/1 \\ 
        yomomma & 0/4 & 0/4 & 0/4 & 1/4  \\ 
        zenon & 1/5 & 3/5 & 5/5 & 5/5  \\ 
        zork1 & 2/18 & 3/18 & 10/18 & 13/18 \\ 
        zork2 & 4/9 & 5/9 & 7/9 & 8/9  \\ 
        zork3 & 0/6 & 1/6 & 0/6 & 4/6  \\ 
        ztuu & 0/9 & 0/9 & 7/9 & 7/9  \\ 
        \bottomrule
    \end{tabular}
    \end{center}
    \end{sc}
    \caption{The experiment results (\# successful / \# answerable minigames) of each model.}
    \label{tab:minigamedetails}
\end{table}

\section{Future Directions}\label{app:future}

Our MANGO benchmark sets up a leaderboard, measuring the mapping and navigation abilities of LLMs. 
An interesting future direction is to investigate how the internal representations of each LLM have---if any---captured the structures of the maps. 
For example, one may probe the LLM representations and examine whether these representations are predictive of certain characteristics of the maze, similar to the experiments of \citet{li2022emergent} in board games. 
This can potentially reveal critical insights into the underlying mechanism of an LLM to represent spatial relations, benefiting their application into real-world scenarios. 

The second interesting direction is to investigate how low-cost-adaptation can improve the performance of an LLM. 
For example, can an LLM, trained on a limited set of mazes, generalize its knowledge and reasoning capability to unseen mazes? 
Findings in this direction could enhance their usefulness in dynamic environments. 

Another direction is to upgrade the MANGO benchmark by enriching its spatial and structural configurations on top of the current maps. These configurations include: 
\begin{itemize}[leftmargin=*,nosep,topsep=0pt]
\item spatial notations (e.g., distance in meters, area in square meters) such that one would need complex movements to achieve a target (e.g., not "north" but "north 3 meters"); 
\item notions of facing directions and rotations such that one would need to turn to switch facing directions. 
\end{itemize}
Such upgrades will enhance its practical application in more complex scenarios. 
They will also enable investigation in interesting topics, including 
\begin{itemize}[leftmargin=*,nosep,topsep=0pt]
    \item LLMs' generalization to significantly out-of-distribution navigation scenarios (e.g., different structures, styles, and configurations); 
    \item LLMs' robustness to distracting context. 
\end{itemize}

\end{document}